\documentclass[final,3p,times]{elsarticle}

\usepackage{amssymb}

\usepackage[ansinew]{inputenc}  
\usepackage{multicol}
\usepackage{amsmath}
\usepackage{graphicx}
\usepackage{booktabs}
\usepackage{xcolor}
\usepackage{svg}
\usepackage{nth}
\usepackage{calc}
\usepackage{cancel}
\usepackage{mathtools}
\usepackage{placeins}
\usepackage{listings}
\usepackage{xspace}
\usepackage{footnote}
\usepackage[bottom]{footmisc}
\usepackage{subfig}
\usepackage{layouts}
\usepackage{physics}
\usepackage{romannum}
\usepackage{bm}
\usepackage{scalerel}
\usepackage{stackengine}
\usepackage{mleftright}
\usepackage{courier}
\usepackage{multirow}
\usepackage{xfrac}
\usepackage{tablefootnote}
\usepackage{bbding}
\usepackage{stmaryrd}
\usepackage{wasysym}
\usepackage{nicefrac}

\usepackage{algorithm} 
\usepackage{algpseudocode} 
\usepackage{transparent}

\usepackage{bbm}

\usepackage[draft]{changes}
\definechangesauthor[name=fs,color=red]{1}

\algnewcommand{\algorithmicor}{\textbf{ or }}
\algnewcommand{\OR}{\algorithmicor}
\algnewcommand{\AND}{\textbf{ and }}

\usepackage{framed} 
\usepackage{multicol} 

\usepackage{setspace}
\journal{}

\svgsetup{inkscapearea=page}

\newcommand{\diag}{\mathop{\mathrm{diag}}}
\newcommand{\conj}[1]{\overline{#1}}

\definecolor{TUMBlau}{RGB}{0,101,189}
\definecolor{TUMBlauDunkel}{RGB}{0,82,147}
\definecolor{TUMBlauHell}{RGB}{152,198,234}
\definecolor{TUMBlauMittel}{RGB}{100,160,200}
\definecolor{TUMElfenbein}{RGB}{218,215,203}
\definecolor{TUMGruen}{RGB}{162,173,0}
\definecolor{TUMOrange}{RGB}{227,114,34}
\definecolor{TUMDunkelGrau}{gray}{0.8}
\definecolor{TUMGrau}{gray}{0.5}
\definecolor{TUMHellGrau}{gray}{0.2}
\definecolor{TUMLila}{RGB}{105,008,090}
\definecolor{TUMRed}{RGB}{156,013,022}
\definecolor{TUMDarkGreen}{RGB}{000,124,048}

\newcommand{\linefullblue}{{\color{TUMBlau}---}}

\newcommand{\linefullgreen}{{\color{TUMGruen}---}}

\newcommand{\linedasheddarkgreen}{{\color{TUMDarkGreen}-- --}}
\newcommand{\linedasheddarkgrey}{{\color{TUMDunkelGrau}-- --}}
\newcommand{\linedashedblue}{{\color{TUMBlau}-- --}}

\newcommand{\linedashdottedorange}{{\color{TUMOrange}-- $\cdot$ --}}

\newcommand{\linedottedblue}{{\color{TUMBlau}$\cdots$}}
\newcommand{\linedottedgreen}{{\color{TUMGruen}$\cdots$}}

\newcommand{\AsteriskGreen}{{\color{TUMGruen}$\ast$}}
\newcommand{\AsteriskRed}{{\color{TUMRed}$\ast$}}

\usepackage{tikz}		
\usepackage{pgfplots}
\pgfplotsset{width=7cm,compat=newest}

\usepackage[framemethod=default]{mdframed}		
\usepackage{paralist}													
\makeatletter 																
\let\@captype\@undefined
\def\newcaption{%
	\begingroup%
	\def\@captype{figure}%
	\refstepcounter\@captype\@dblarg{\@newcaption\@captype}%
	\endgroup%
}
\makeatother

\newcommand{\BackgroundColorBoxxx}{white}
\mdfdefinestyle{boxxxstyle}{
	topline=true,
	frametitleaboveskip=-1pt,
	linewidth=2pt,
	linecolor={black!50!white},
	backgroundcolor={\BackgroundColorBoxxx},
	fontcolor=black,
	font={},
	innerleftmargin=3ex,
	innerrightmargin=3ex,
	innertopmargin=3ex,
	innerbottommargin=3ex,
	leftmargin=0ex,
	rightmargin=0,
	skipabove=3ex,
	skipbelow=3ex
}

\NewDocumentEnvironment{boxxx}{O{} O{5cm}}{
	\Needspace{#2}
	\ifstrempty{#1}{}{
		\mdfsetup{
			frametitle={\hspace{-3ex}
				\tikz[baseline=(current bounding box.east),outer sep=0pt]
				\node[anchor=east,rectangle,fill=black!50!white]
				{\strut \textcolor{white}{\textsf{\textbf{#1}}}};}
		}
	}
	\mdfsetup{style=boxxxstyle}
	\begin{mdframed}[]\relax
	}
	{
	\end{mdframed}
}

\newcommand{\Grad}{\ifmmode \mGrad \else  $\hspace{-0.3em}^\circ$\,  \fi}
\newcommand{\mGrad}{\ensuremath{^\circ}} 

\newcommand{\lived}[2]{($\ast$#1\ifthenelse{\equal{#2}{}}{}{, $\dagger$#2})}

\newcommand{\pdvtwo}[2]{\frac{}{}}

\renewcommand{\Re}{\operatorname{Re}}
\renewcommand{\Im}{\operatorname{Im}}

\newcommand{\pdf}{f}
\newcommand{\obj}{h}

\newcommand{\RNum}[1]{\uppercase\expandafter{\romannumeral #1\relax}}

\newcommand{\matlab}{MATLAB\textsuperscript{{\tiny \textregistered}}\xspace}
\newcommand{\ansys}{ANSYS\textsuperscript{{\tiny \textregistered}}\xspace}

\newcommand{\softplus}{\operatorname{softplus}}

\newcommand{\E}{\operatorname{\mathbb{E}}}

\newcommand{\Cov}{\operatorname{Cov}}

\makeatletter
\newcommand{\specialcell}[1]{\ifmeasuring@#1\else\omit$\displaystyle#1$\ignorespaces\fi}
\makeatother

 \graphicspath{{figures/}{svg-inkscape/}} 

\DeclareMathOperator*{\argmax}{arg\,max}

\newcommand{\ntrain}{n_{\mathrm{tr}}}

\usepackage{pgfplots}
\usepackage{tikz,pgfplotstable}                    
\usetikzlibrary{arrows}
\usepgfplotslibrary{statistics}      
\usepgfplotslibrary{fillbetween}     

\vfuzz\hfuzz
\pgfplotsset{compat=newest,  
	legend style={font=\color{black}\sffamily\footnotesize, draw=none,legend cell align=left},
	every mark/.append style={scale=0.7},
	every axis/.append style={ 
		execute at begin axis={
			scale only axis,
			minor y tick num=1, 
			font=\color{black}\sffamily\footnotesize,
			text=black,
			major grid style = solid,
			minor grid style = dotted,
		},},
	every axis label/.append style={font=\sffamily\footnotesize,text=black},
	tick label style={execute at begin axis={
			font=\sffamily\footnotesize,text=black},
	},
	unbounded coords=jump,
	,
}

\newlength\fheight  
\newlength\fwidth   

\pgfplotsset{
	every axis plot/.append style={line width=0.8pt}
}

\begin{document}

\begin{frontmatter}

	\pagenumbering{arabic} 
		
	\title{Maximum a Posteriori Estimation for Linear Structural Dynamics Models Using Bayesian Optimization with Rational Polynomial Chaos Expansions}
	
	\author[label2]{Felix Schneider}
	\author[label1]{Iason Papaioannou}
	\author[label3]{Bruno Sudret}
	\author[label2]{Gerhard M\"{u}ller}
	
	\address[label1]{Engineering Risk Analysis Group, Technical University of Munich, Arcisstr. 21, 80333 Munich, Germany}
	\address[label2]{Chair of Structural Mechanics, Technical University of Munich, Arcisstr. 21, 80333 Munich, Germany}
	\address[label3]{Chair of Risk, Safety and Uncertainty Quantification, ETH Zurich, Stefano-Franscini-Platz 5, 8093 Zurich, Switzerland}
	
	\begin{abstract}
		Bayesian analysis enables combining prior knowledge with measurement data to learn model parameters.
		Commonly, one resorts to computing the maximum a posteriori (MAP) estimate, when only a point estimate of the parameters is of interest. 
		We apply MAP estimation in the context of structural dynamic models, where the system response can be described by the frequency response function.
		To alleviate high computational demands from repeated expensive model calls, we utilize a rational polynomial chaos expansion (RPCE) surrogate model that expresses the system frequency response as a rational of two polynomials with complex coefficients. 
		We propose an extension to an existing sparse Bayesian learning approach for RPCE based on Laplace's approximation for the posterior distribution of the denominator coefficients.
		Furthermore, we introduce a Bayesian optimization approach, which allows to adaptively enrich the experimental design throughout the optimization process of MAP estimation. 
		Thereby, we utilize the expected improvement acquisition function as a means to identify sample points in the input space that are possibly associated with large objective function values.
		The acquisition function is estimated through Monte Carlo sampling based on the posterior distribution of the expansion coefficients identified in the sparse Bayesian learning process.
		By combining the sparsity-inducing learning procedure with the sequential experimental design, we effectively reduce the number of model evaluations in the MAP estimation problem.
		We demonstrate the applicability of the presented methods on the parameter updating problem of an algebraic two-degree-of-freedom system and the finite element model of a cross-laminated timber plate. 
	\end{abstract}
	

\end{frontmatter}


\section{Introduction}
\label{sec:intro}
A fundamental task in almost all engineering and applied sciences fields is the assessment of the reliability, serviceability or comfortability of technical systems.
This assessment can be done through various approaches, which we broadly categorize into model- and data-driven approaches.
Within the field of data-driven approaches, data of a system's performance is gathered and processed in order to make predictions of a system's performance in the sense outlined above. 
Data-driven approaches are also commonly referred to as non-parametric.
On the other hand, in model-based approaches one tries to develop a fundamental understanding of the underlying natural laws that govern the behavior of the system under consideration.
Based on this understanding, a mathematical model can be formulated that allows one to predict the system's performance.
Typically, the model is defined by a set of differential equations, whose parameters define the characteristics of the system. 
In many cases, both of the above, i.e., data and a model, are available. 
In these cases it is of great interest to combine the two in order to improve the assessment of the system.

This task is often referred to as parametric system identification or model updating \cite{Astrom.1971,Soederstroem.1989,Mottershead.1993,Pintelon.2012}.
The field of parametric system identification can further be split into probabilistic and non-probabilistic approaches. 
Non-probabilistic (deterministic) approaches to model updating aim at finding an optimal set of model parameters that minimize the misfit between model prediction and measurement data \cite{friswell2013finite}.
Probabilistic approaches to parametric model updating treat the unknown parameters in the model as random variables and express the discrepancy between the measurement data and the model as uncertain as well. 
Often, this discrepancy is expressed through uncertain measurement and modeling errors, whereby the formulation and assumptions regarding the involved errors are highly problem-specific.
A common approach to solving the probabilistic parametric model updating problem is through the application of the Bayesian approach.
Within Bayesian updating, the goal is to compute the probabilistic description of the system parameters conditional on the observed measurement data. 

Within the Bayesian approach, one states the updating problem through the application of Bayes' rule.
The goal is to find the probability distribution of the model parameters conditional on the observed data \cite{gelman2013bayesian}.
This so-called posterior distribution is expressed as the normalized product of the likelihood function and the prior distribution.
The likelihood function is commonly derived based on an assumption on the misfit between the model and the observations, while the prior distribution encodes belief about the model parameters before the observations become available.
The inclusion of the prior distribution effectively tackles problems of ill-posedness or non-identifiability that are common to non-Bayesian paradigms \cite{Beck.2010}.
A common issue in Bayesian inference is the intractability of the normalizing integral, which requires performing integration over high-dimensional parameter spaces. 
To circumvent this problem, approximate numerical or semi-analytical approaches have been proposed, including Laplace approximation \cite{mackay2003information}, Markov Chain Monte Carlo (MCMC) sampling \cite{gilks1995markov, cheung2017new, muto.2008,Beck.2010, Zhang.2010,Zhang.2011,Zhang.2013} or Bayesian Updating with Structural Reliability Methods \cite{Straub.2015, diazdelao2017bayesian, Betz.2018}.

Whenever one is only interested in a point-estimate of the posterior distribution, however, the maximum-a-posteriori (MAP) estimate is often considered.
The MAP estimate is defined as the mode of the posterior probability density function (PDF). 
The problem of finding the MAP estimate can thus be stated as an optimization problem.
Nevertheless, finding the optimum with numerical schemes requires evaluating the likelihood, and thus the model function, multiple times.
For computationally complex models this renders the solution of the optimization problem an expensive task. 
In order to lower the computational burden, one can resort to surrogate modeling, whereby the original model is replaced by sufficiently accurate approximations that are fast to evaluate.
Popular choices include polynomial chaos expansions (PCE) \cite{ghanem1991stochastic,xiu2002wiener}, Neumann series expansions \cite{yamazaki1988neumann} and machine learning techniques such as neural networks \cite{papadrakakis1996structural} or Gaussian process models \citep{Rasmussen.2008}.
The surrogate modeling approach commonly involves an offline and an online phase.
During the offline phase, the model is trained by means of a suitable training technique. 
Here, we consider non-intrusive approaches that rely on discrete evaluations of the model outcome, i.e., the training data, at a set of realizations of the model input parameters, i.e., the experimental design.
After the successful training of the surrogate model to the training data, the surrogate model is utilized in the application at hand.

A number of methods have been proposed for the generation of the experimental design.
Popular choices include Latin hypercube sampling (LHS) \cite{mckay1979comparison} or quasi-random sampling \cite{owen2003quasi}. 
In their standard settings, these strategies generate a fixed specified number of samples that can be subsequently used in the training procedure. 
Using a fixed and static experimental design that follows the prior joint distribution of the input parameters can become sub-optimal for inverse problems. 
This is due to the fact that for inverse problems, the bulk of the posterior probability mass may lie in a region of the input space that is only sparsely represented by a set of samples following the prior distribution. 
Thereby, the accuracy of the surrogate models in this relevant part of the input space might be poor. 
To circumvent this problem, active learning or sequential experimental design strategies have been proposed, in which the experimental design is enriched throughout the training procedure.

Subsequently, we consider sequential design strategies in the context of solving optimization problems.
Within this field, Bayesian optimization has emerged as a powerful tool for solving global optimization problems for black-box functions.
The methodology is based on replacing the objective function in the problem through a random process surrogate model over the space of input parameters.
Subsequently, based on an assessment of the predictive uncertainty of the random process model, suitable experimental design points are chosen for which an increase in the objective function value is expected.
More specifically, the sample locations are commonly chosen based on maximizing a so called acquisition function.
Popular acquisition functions are the probability of improvement (PI), upper confidence bound (UCB), or the expected improvement (EI).
These acquisition functions have in common that they are \textit{myopic} in the sense that only the improvement gained through a single next observation is considered, which is considered a one-step lookahead approach \cite{garnett_bayesoptbook_2023}.
For an extensive review and further literature we refer the reader to \cite{Frazier.2018,Shahriari.2016,garnett_bayesoptbook_2023,Greenhill.2020,Wang.2023}.
Bayesian optimization has found widespread use since the publication of \cite{Jones.1998}.
Therein, the authors propose to use a Gaussian process model, specifically, ordinary Kriging, to approximate the expensive-to-evaluate objective function. 
Their framework, termed Efficient Global Optimization, sequentially chooses sample locations through maximizing the expected improvement acquisition function. 
In recent years, various variants and modifications, including constrained and multi-objective versions, have been proposed. 
We refer the reader to \cite{Shahriari.2016,garnett_bayesoptbook_2023} and the references therein.
A sequential surrogate modeling approach for finite element model updating has been proposed in \cite{Jin.2016}.
Therein, a Gaussian process surrogate model is used to approximate the objective function and the expected improvement acquisition function is employed.
The use of non-stationary Gaussian processes in conjunction with the expected information gain criterion has been proposed in the context of Bayesian experimental design \cite{Pandita.2021} for efficient sample allocation.
An active learning approach for sequential surrogate modeling in the context of parametric model order reduction is proposed in \cite{Kapadia.2024}.

Within the scope of this paper, we are particularly interested in Bayesian updating problems that utilize dynamic data and formulate the likelihood in terms of the misfit of the data in the frequency domain as proposed in, e.g., \cite{Schneider.2022, Zhang.2010, Zhang.2011, Zhang.2013, Mares.2006}. 
In such problems, due to the formulation of the likelihood function based on frequency domain data, the log-likelihood function will show a rational dependency on the model input parameters.
In initial investigations, it was found that standard Bayesian optimization techniques utilizing Gaussian process regression with commonly used kernel functions do not lead to satisfactory MAP estimates.
Recently, a kernel-based interpolation for complex-valued functions in the context of approximating frequency response functions has been proposed in \cite{Bect.25.07.2023}.
While this approach enables efficient Gaussian process regression by utilizing specifically suitable kernels for frequency responses that could be utilized in the Bayesian optimization context, it is currently restricted to one-dimensional input spaces.
The extension to multidimensional inputs still remains an open research question.
Furthermore, the authors in \cite{wagner2021bayesian} propose a multi-element PCE approach for Bayesian inversion to tackle locally complex dependencies of the likelihood function on the input parameters. 
Therein, locally refined spectral expansions are used to surrogate the likelihood function.

This paper proposes a sequential design-based MAP estimation procedure for parameter identification that extends the methodologies in \cite{Schneider.2022} and \cite{Schneider.2023}.
The likelihood function is derived on the basis of an assumption on the logarithm of the complex-valued model error between the measured and simulated frequency response of the system as proposed in \cite{Schneider.2022}.
The model misfit is considered for a set of spatial and frequency domain observation points. 
To improve computational efficiency, we replace the original models that enter the likelihood function by a set of rational polynomial chaos expansion (RPCE) models that are specifically suitable for approximating frequency response functions.
The RPCE expresses the model output as a ratio of two PCEs with complex-valued coefficients and is chosen due to its improved accuracy over standard polynomial chaos expansion (PCE) models in the context of approximating frequency responses \cite{Jacquelin.2015,Jacquelin.2015b,Jacquelin.2016,JacquelinE.2016}.
In \cite{Schneider.2022}, based on \cite{Schneider.2019}, a least-squares regression approach and static experimental design have been applied to find the coefficients in the RPCE for subsequent use in the Bayesian model updating procedure.
However, this still requires a significant number of model evaluations to compute the surrogate model coefficients to avoid poor approximation due to overfitting. 
Recently, building upon the sparse Bayesian regression methodology in \cite{Tipping.2001}, a Bayesian regression approach to learning the coefficients of the RPCE has been proposed in \cite{Schneider.2023}.
In this paper, we introduce an alternative Bayesian regression approach to the one in \cite{Schneider.2023} that is based on a linearized error model formulation.
Through the linearized formulation, we have access to a closed-form expression of the Hessian of the log-posterior of the denominator coefficients.
This allows us to approximate the posterior distribution of the denominator coefficients through a proper complex normal distribution, i.e., a complex variant of Laplace's approximation. 
The proposed method is able to accurately represent the model output with a higher degree of sparsity compared to the approach in \cite{Schneider.2023}.
In addition, the Bayesian paradigm allows us to obtain a measure of uncertainty of the surrogate model prediction.
Based on the developed Bayesian RPCE models, we tackle the MAP estimation problem in a Bayesian optimization framework. 
Under the above derivation, the objective function is expressed as a random process over the space of input parameters that is further parameterized by the random surrogate model coefficients as well as the surrogate model errors. 
We then sequentially add sample points to the experimental design based on maximizing the expected improvement acquisition function.
Since we do not have direct access to the distribution of the random process model, the expected improvement is not available in closed-form and is therefore approximated through Monte Carlo sampling.
Two numerical examples are presented that apply the proposed MAP estimation methodology to the parameter updating problem of a two-degree-of-freedom vibratory system and a cross-laminated timber (CLT) plate model.

The outline of the paper is as follows.
First, the methodology is introduced in Section~\ref{sec:method}.
In Section~\ref{subsec:bu_par_upd}, the Bayesian updating problem, as presented in \cite{Schneider.2022} is comprehensively summarized.
Subsequently, Section~\ref{subsec:rpce} presents the novel sparse Bayesian learning strategy for RPCE surrogate models.
Finally, based on the presentation in Sections~\ref{subsec:bu_par_upd}~and~\ref{subsec:rpce}, a novel Bayesian optimization strategy for MAP estimation utilizing RPCE models is introduced in Section~\ref{subsec:bayes_opt}.
The proposed method is then applied to the parameter updating problem for an algebraic two-degree-of-freedom model as well as a finite element model in \ref{sec:num_examples}.
In Section~\ref{subsec:disc}, open questions and critical issues are discussed. 
The paper closes with the conclusions in Section~\ref{subsec:concl}.

\section{Methodology}
\label{sec:method}
In this section, we introduce the novel Bayesian model parameter updating methodology.
The considered frequency domain dynamic model is comprehensively introduced in Section~\ref{sec:sys}.
Subsequently, the Bayesian approach to model updating is outlined in Section~\ref{subsec:bu_subsec_freq}.
An modified version of a recently introduced sparse Bayesian learning algorithm for RPCE is introduced in Section~\ref{subsec:rpce}. 
This will be the basis for the Bayesian optimization approach introduced in Section~\ref{subsec:bayes_opt}.
\subsection{Bayesian model parameter updating}
\label{subsec:bu_par_upd}
\subsubsection{Linear dynamic model with parameter uncertainty}
\label{sec:sys}
Within the proposed framework, Bayesian updating makes use of measurement data to update the parameters of an engineering model of a given physical system. 
We assume that the physical system is modeled by a space-discretized, linear dynamic system with $n_{\mathrm{dof}}$ degrees of freedom (DOF). 
Let $\mathbf{X} $ be a random vector with outcome space $\mathbb{R}^d$ and joint PDF $\pdf_{\mathbf{X}}$. 
$\mathbf{X} $ models a set of uncertain parameters that influence the state of the dynamic system. 
The matrices $\mathbf{K} \left(\mathbf{X}\right)$, $\mathbf{C} \left(\mathbf{X}\right)$ and $\mathbf{M} \left(\mathbf{X}\right)$ denote the $n_{\mathrm{dof}} \times n_{\mathrm{dof}}$ sized stiffness, damping and mass matrix with parametric uncertainty. 
The equation of motion describing the system state in the frequency domain is given as
\begin{equation}
	\mathbf{K} \left(\mathbf{X}\right) \Tilde{\mathbf{u}} \left( \omega , \mathbf{X}\right) + \mathrm{i}\omega \mathbf{C} \left(\mathbf{X}\right) \Tilde{\mathbf{u}} \left(\omega, \mathbf{X}\right) - \omega^2 \mathbf{M} \left(\mathbf{X}\right) \Tilde{\mathbf{u}} \left(\omega, \mathbf{X}\right) = \Tilde{\mathbf{f}} (\omega) \, .
	\label{eq:eq_of_mot_fe2}
\end{equation}
Here, $\Tilde{\mathbf{f}}  \left(\omega\right)$ and $\Tilde{\mathbf{u}} \left(\omega, \mathbf{X}\right)$ are the deterministic force and the uncertain displacement vector in the frequency domain and $\mathrm{i}=\sqrt{-1}$ denotes the imaginary number. 

From the above, it is evident that the outcome space of the solution $\Tilde{\mathbf{u}}$ is the $N$-dimensional complex set $\mathbb{C}^{n_{\mathrm{dof}}}$. 
The frequency response function (FRF) $\Tilde{h}_{ij} :\mathbb{R} \times \mathbb{R}^d \to \mathbb{C}$, defining the acceleration at DOF $i$ due to a force $\Tilde{f}_j$ at DOF $j$ in terms of the circular frequency $\omega$ is then found by the ratio 
\begin{equation}
	\Tilde{h}_{ij} \left(\omega , \mathbf{X}\right) = \mathbf{e}_i^\mathrm{T} \left( \mathbf{K} \left(\mathbf{X}\right) + \mathrm{i}\omega \mathbf{C} \left(\mathbf{X}\right) - \omega^2 \mathbf{M} \left(\mathbf{X}\right) \right)^\mathrm{-1} \mathbf{e}_j \, , \label{eq:frf_model}
\end{equation}
where $\mathbf{e}_i$ is the standard unit vector with unit entry at the $i$-th position and zero otherwise, and $(\cdot)^\mathrm{T}$ denotes the transpose of a vector.
\subsubsection{Bayesian model updating utilizing frequency domain data}
\label{subsec:bu_subsec_freq}
In this section, we introduce the Bayesian formulation of the model updating problem.
We follow the previously proposed approach in \cite{Schneider.2022} and assume that the considered system is equipped with a set of $n_s$ sensors, which measure the system's acceleration. 
Under controlled force excitation, the measured system excitation and response can be processed to compute the system's frequency response functions. 
Commonly this includes a discrete Fourier transform, which yields frequency response information on a fine frequency grid. 
The available data is further reduced to a set of frequencies of interest $\{ \omega_{\mathcal{O},j} | j = 1, \ldots, n_\omega \}$. 
We thus obtain a set of frequency response function measurements, denoted by $\mathcal{D}_\mathcal{O} = \{y_{\mathcal{O},i} | i = 1, \ldots, n_\mathcal{O} \}$, where $n_\mathcal{O} = n_\omega n_s$ denotes the overall number of data points.
This is illustrated on the left hand side of Fig.~\ref{fig:measurement_model_bayes}.

Similarly, the system response at the $n_s$ spatial points is predicted at the $n_\omega$ frequency points through the model in Eq.~\eqref{eq:frf_model}.
Motivated by the proposed surrogate modeling approach in Section~\ref{subsec:rpce}, which allows the approximation of scalar-valued model responses, we denote by $\{ \mathcal{M}_{i} \left(\mathbf{X}\right) | i = 1, \ldots, n_\mathcal{O} \}$ a set of models that predict the response at the $n_\mathcal{O}$ spatial and frequency points.
Furthermore, $\mathbf{Y}_\mathcal{M} = \left[ Y_{\mathcal{M},1} ; \ldots; Y_{\mathcal{M},n_\mathcal{O}} \right]$, where $Y_{\mathcal{M},i} = \mathcal{M}_{i} \left(\mathbf{X}\right)$, denotes the corresponding output random vector. 
Each of the models $\mathcal{M}_{i}$ depends on the uncertain input parameters $\mathbf{X}$.
This is illustrated on the right hand side of Fig.~\ref{fig:measurement_model_bayes}.

%
\begin{center}
\begingroup%
  \makeatletter%
  \providecommand\color[2][]{%
    \errmessage{(Inkscape) Color is used for the text in Inkscape, but the package 'color.sty' is not loaded}%
    \renewcommand\color[2][]{}%
  }%
  \providecommand\transparent[1]{%
    \errmessage{(Inkscape) Transparency is used (non-zero) for the text in Inkscape, but the package 'transparent.sty' is not loaded}%
    \renewcommand\transparent[1]{}%
  }%
  \providecommand\rotatebox[2]{#2}%
  \newcommand*\fsize{\dimexpr\f@size pt\relax}%
  \newcommand*\lineheight[1]{\fontsize{\fsize}{#1\fsize}\selectfont}%
  \ifx\svgwidth\undefined%
    \setlength{\unitlength}{399.99998474bp}%
    \ifx\svgscale\undefined%
      \relax%
    \else%
      \setlength{\unitlength}{\unitlength * \real{\svgscale}}%
    \fi%
  \else%
    \setlength{\unitlength}{\svgwidth}%
  \fi%
  \global\let\svgwidth\undefined%
  \global\let\svgscale\undefined%
  \makeatother%
  \begin{picture}(1,0.28125001)%
    \lineheight{1}%
    \setlength\tabcolsep{0pt}%
    \put(0,0){\includegraphics[width=\unitlength,page=1]{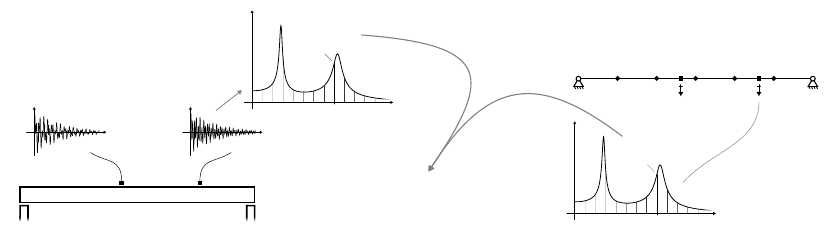}}%
    \put(0.47407106,0.16704064){\color[rgb]{0,0,0}\makebox(0,0)[lt]{\lineheight{1.25}\smash{\begin{tabular}[t]{l}$\scriptstyle \omega$\end{tabular}}}}%
    \put(0.31548227,0.12548919){\color[rgb]{0,0,0}\makebox(0,0)[lt]{\lineheight{1.25}\smash{\begin{tabular}[t]{l}$\scriptstyle t$\end{tabular}}}}%
    \put(0.12754537,0.12548919){\color[rgb]{0,0,0}\makebox(0,0)[lt]{\lineheight{1.25}\smash{\begin{tabular}[t]{l}$\scriptstyle t$\end{tabular}}}}%
    \put(0.02106294,0.22215546){\color[rgb]{0,0,0}\makebox(0,0)[lt]{\lineheight{1.25}\smash{\begin{tabular}[t]{l}\underline{Measurement}\end{tabular}}}}%
    \put(0.44479028,0.03869696){\color[rgb]{0,0,0}\makebox(0,0)[lt]{\lineheight{1.25}\smash{\begin{tabular}[t]{l}\underline{Bayes' rule}\end{tabular}}}}%
    \put(0.25722305,0.17049456){\color[rgb]{0,0,0}\makebox(0,0)[lt]{\lineheight{1.25}\smash{\begin{tabular}[t]{l}$\mathcal{F}$\end{tabular}}}}%
    \put(0.38385638,0.13681059){\color[rgb]{0,0,0}\makebox(0,0)[lt]{\lineheight{1.25}\smash{\begin{tabular}[t]{l}$\omega_{\mathcal{O},j}$\end{tabular}}}}%
    \put(0.37423756,0.23322448){\color[rgb]{0,0,0}\makebox(0,0)[lt]{\lineheight{1.25}\smash{\begin{tabular}[t]{l}$y_{\mathcal{O},i}$\end{tabular}}}}%
    \put(0.86124167,0.03376638){\color[rgb]{0,0,0}\makebox(0,0)[lt]{\lineheight{1.25}\smash{\begin{tabular}[t]{l}$\scriptstyle \omega$\end{tabular}}}}%
    \put(0.77102709,0.00353633){\color[rgb]{0,0,0}\makebox(0,0)[lt]{\lineheight{1.25}\smash{\begin{tabular}[t]{l}$\omega_{\mathcal{O},j}$\end{tabular}}}}%
    \put(0.7501581,0.09620023){\color[rgb]{0,0,0}\makebox(0,0)[lt]{\lineheight{1.25}\smash{\begin{tabular}[t]{l}$\mathcal{M}_i (\mathbf{x})$\end{tabular}}}}%
    \put(0.75794477,0.22215546){\color[rgb]{0,0,0}\makebox(0,0)[lt]{\lineheight{1.25}\smash{\begin{tabular}[t]{l}\underline{Simulation}\end{tabular}}}}%
    \put(0.93848702,0.14343167){\color[rgb]{0,0,0}\makebox(0,0)[lt]{\lineheight{1.25}\smash{\begin{tabular}[t]{l}$\mathcal{M} (\mathbf{x})$\end{tabular}}}}%
  \end{picture}%
\endgroup%

	
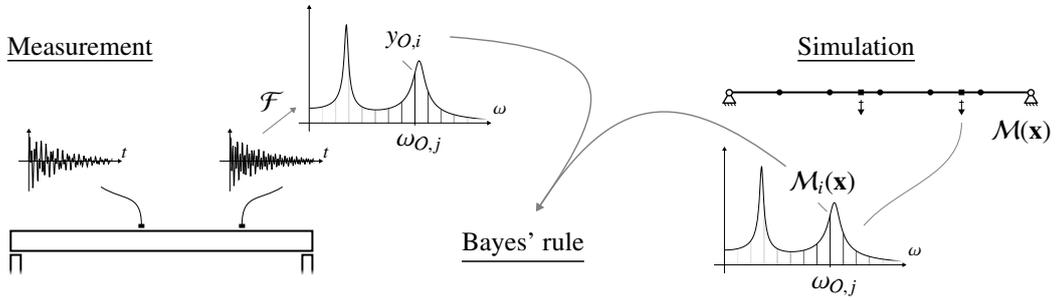
\captionof{figure}{
		Illustration of data and model fusion through Bayes' rule. 
		Data is gathered on a structure at specific locations. 
		We consider this data in the frequency domain through application of the Fourier transform (denoted by $\mathcal{F}$) to the time domain data.
		The system response is also predicted through a simulation model. 
		The information from observations as well as the models can then be combined through Bayes' rule to update the model parameters in $\mathbf{X}$.
	}
	\label{fig:measurement_model_bayes}
\end{center}
We are now interested in updating our belief about the system parameters.
To this end we apply Bayes' rule and obtain the posterior distribution of the system parameters as
\begin{equation}
	\pdf ( \mathbf{x} | \mathcal{D}_{\mathcal{O}}) = c_E^{-1} L ( \mathbf{x} | \mathcal{D}_{\mathcal{O}}) \pdf_{\mathbf{X}} ( \mathbf{x} ) \, , \label{eq:bayes} 
\end{equation}
where $L ( \mathbf{x} | \mathcal{D}_{\mathcal{O}}) \propto \pdf ( \mathcal{D}_{\mathcal{O}} | \mathbf{x})$ denotes the likelihood function and $\pdf_{\mathbf{X}} ( \mathbf{x} )$ denotes the prior distribution of the parameters.
The normalization constant $c_E$, commonly referred to as marginal likelihood or evidence, is defined as $c_E = \int L ( \mathbf{x} | \mathcal{D}_{\mathcal{O}}) \pdf ( \mathbf{x} ) \dd{\mathbf{x}}$ and ensures that the posterior distribution is properly normalized.
The likelihood is derived based on the assumption of the following error model,
\begin{equation}
	y_{\mathcal{O},i} + \varepsilon_{\mathcal{O},i} = \mathcal{M}_i \left( \mathbf{x} \right) \cdot \varepsilon_{\mathcal{M},i} \, , %
	\label{eq:error} 
\end{equation}
where $\varepsilon_{\mathcal{O},i}$ and $\varepsilon_{\mathcal{M},i}$ denote the observation noise and the multiplicative model error that jointly explain the misfit between the model output $\mathcal{M}_i \left( \mathbf{x} \right)$ and the observation $y_{\mathcal{O},i}$.
%
Subsequently, we assume that $\varepsilon_{\mathcal{O},i} \ll \varepsilon_{\mathcal{M},i}$ and hence can be disregarded and denote $\varepsilon_{i} = \varepsilon_{\mathcal{M},i}$ for notational convenience.
We denote by 
$\bm{\varepsilon} = \left[ \varepsilon_{1}; \ldots; \varepsilon_{n_\mathcal{O}} \right]$ 
the vector collecting all errors; $ \bm{\varepsilon}$ can be expressed as
$\bm{\varepsilon} = \abs{\bm{\varepsilon}} \cdot \mathrm{e}^{\mathrm{i} \bm{\varphi}_{\varepsilon}}$ with
$\abs{\bm{\varepsilon}}$ and $\bm{\varphi}_{\varepsilon} = \angle \bm{\varepsilon}$ denoting the element-wise absolute value and phase of the complex-valued vector $\bm{\varepsilon}$.
Furthermore, we choose to model the distribution of $\abs{\bm{\varepsilon}}$ using a multivariate lognormal distribution and the distribution of $\bm{\varphi}_{\varepsilon}$ through a multivariate normal distribution and consider them to be independent.
Then the element-wise logarithm of the error vector
\begin{equation}
	\ln \bm{\varepsilon} = \underbrace{\ln \abs{\bm{\varepsilon}}}_{=:\mathbf{w}_{\varepsilon}} + \mathrm{i} \bm{\varphi}_{\varepsilon} \, ,  
	\label{eq:error_dist} 
\end{equation}
follows a complex normal distribution. 
Taking the logarithm of Eq.~\eqref{eq:error}, and by introducing $\mathbf{y}_\mathcal{O} = \left[ y_{\mathcal{O},1} ; \ldots; y_{\mathcal{O},{n_{\mathcal{O}}}} \right]$ and $\mathbf{y}_\mathcal{M} (\mathbf{x}) = \left[ \mathcal{M}_1 (\mathbf{x}); \ldots; \mathcal{M}_{n_{\mathcal{O}}} (\mathbf{x}) \right]$ we obtain
\begin{equation}
	\ln \mathbf{y}_\mathcal{O} = \ln \mathbf{y}_\mathcal{M} (\mathbf{x}) + \ln \bm{\varepsilon} \, .
	\label{eq:error_log} 
\end{equation}
Based on Eq.~\eqref{eq:error_log}, we can write the likelihood as
\begin{equation}
	L \left( \mathbf{x} | \mathcal{D}_\mathcal{O} \right) = \pdf_{\ln \bm{\varepsilon}} \left( \ln \mathbf{y}_\mathcal{O} - \ln \mathbf{y}_\mathcal{M} (\mathbf{x}) \right) \, .
	\label{eq:likeliii}
\end{equation}
We proceed by defining the mean vector and covariance matrix of the logarithm of the error in order to fully define the likelihood function. 
Firstly, we assume that the logarithm of the absolute value and the phase of the error have zero mean, i.e., $\mathbb{E} \left[\mathbf{w}_{\varepsilon}\right] = \mathbf{0}_{n_{\mathcal{O}}}$ and $\mathbb{E} \left[\bm{\varphi}_{\varepsilon}\right] = \mathbf{0}_{n_{\mathcal{O}}}$.
%
%
This will render the median of $\abs{\bm{\varepsilon}}$ to be $\abs{\bm{\varepsilon}}_{0.5} = \mathbf{1}_{n_{\mathcal{O}}}$. 
Here, $\mathbf{0}_{n}$ and  $\mathbf{1}_{n}$ denote the all zero and all ones vector with $n$ entries.
Under the above assumptions the real composite vector $\mathbf{r} = \left[ \mathbf{w}_{\varepsilon}; \bm{\varphi}_{\varepsilon} \right]$ will follow a zero-mean normal distribution with covariance matrix
\begin{equation}
	\bm{\Sigma}_{\mathbf{r}\mathbf{r}} = \mathbb{E} \left[ \mathbf{r} \mathbf{r}^\mathrm{T} \right] =  
	\begin{bmatrix}
		\bm{\Sigma}_{\mathbf{w}_{\varepsilon}\mathbf{w}_{\varepsilon}} & \mathbf{0}_{n_{\mathcal{O}} \times n_{\mathcal{O}}} \\
		\mathbf{0}_{n_{\mathcal{O}} \times n_{\mathcal{O}}} & \bm{\Sigma}_{\bm{\varphi}_{\varepsilon}\bm{\varphi}_{\varepsilon}}
	\end{bmatrix} \, ,
\end{equation}
where $\bm{\Sigma}_{\mathbf{w}_{\varepsilon}\mathbf{w}_{\varepsilon}} = \mathbb{E} \left[ \mathbf{w}_{\varepsilon} \mathbf{w}_{\varepsilon}^\mathrm{T} \right]$, $\bm{\Sigma}_{\bm{\varphi}_{\varepsilon}\bm{\varphi}_{\varepsilon}} = \mathbb{E} \left[ \bm{\varphi}_{\varepsilon} \bm{\varphi}_{\varepsilon}^\mathrm{T} \right]$ and $\mathbf{0}_{n \times n}$ denotes the $n\times n$ matrix of zeros.
The choice of a specific model for the above covariance matrices is problem-dependent and will be further discussed in the specific application cases in Section~\ref{sec:num_examples}.
The same holds for the definition of the prior distribution.
Having fully specified the likelihood function and joint prior distribution, we proceed to the computation of the posterior distribution.
Except for few cases, e.g., in the case of conjugate priors, a closed-form solution for the posterior distribution is not available.
Full inference on the model parameters is then often done through sampling-based approaches.
If the engineering analysis does not require computation of the full posterior distribution, one can resort to computing posterior point descriptors of the model parameters.
A commonly chosen descriptor is the mode of the posterior PDF, the maximum a posteriori value.
\paragraph{Maximum a posteriori estimation}
\label{subsec:map}
%
%
The maximum a posteriori (MAP) value $\mathbf{x}^\ast$ can be found as the solution of the following optimization problem
\begin{equation}
	\mathbf{x}^\ast \in \argmax_{\mathbf{x} \in \mathbb{R}^{d}} \left[ \ln L \left( \mathbf{x} | \mathcal{D}_{\mathcal{O}} \right) + \ln \pdf \left( \mathbf{x} \right) \right] \, .
	\label{eq:def_map}
\end{equation}
Under consideration of the derivation in the preceding section, the objective function $\obj (\mathbf{x}) \propto \ln \pdf \left( \mathbf{x} | \mathcal{D}_{\mathcal{O}} \right) + \ln \pdf \left( \mathbf{x} \right)$ can be written as
\begin{equation}
	\obj (\mathbf{x}) = \frac{1}{2} \left[
	- \ln \det \bm{\Sigma}_{\mathbf{w}_{\varepsilon}\mathbf{w}_{\varepsilon}} 
	- \ln \det \bm{\Sigma}_{\bm{\varphi}_{\varepsilon}\bm{\varphi}_{\varepsilon}} 
	- \norm{\vphantom{\bm{\varphi}}\mathbf{w}_{\mathcal{O}} - \mathbf{w}_{\mathcal{M}} \left(\mathbf{x}\right)}_{\bm{\Sigma}_{\mathbf{w}_{\varepsilon}\mathbf{w}_{\varepsilon}}^{-1}}^2
	- \norm{\bm{\varphi}_{\mathcal{O}} - \bm{\varphi}_{\mathcal{M}} \left(\mathbf{x}\right)}_{\bm{\Sigma}_{\bm{\varphi}_{\varepsilon}\bm{\varphi}_{\varepsilon}}^{-1}}^2
	\right]
	+ \ln \pdf \left( \mathbf{x} \right) \, ,
	\label{eq:map_obj}
\end{equation}
where 
$\norm{\mathbf{x}}_{\mathbf{A}}^2 = \mathbf{x}^\mathrm{H} \mathbf{A} \mathbf{x}$,
$\mathbf{w}_\mathcal{O} = \left[ \ln |y_{\mathcal{O},1}| ; \ldots; \ln |y_{\mathcal{O},{n_{\mathcal{O}}}}| \right]$, 
$\mathbf{w}_\mathcal{M} (\mathbf{x}) = \left[ \ln |\mathcal{M}_1 (\mathbf{x})| , \ldots, \ln |\mathcal{M}_{n_{\mathcal{O}}} (\mathbf{x})| \right]$, 
$\bm{\varphi}_{\mathcal{O}} = \left[ \angle{y_{\mathcal{O},1}} , \ldots, \angle {y_{\mathcal{O},{n_{\mathcal{O}}}}} \right]$ and 
$\bm{\varphi}_\mathcal{M} (\mathbf{x}) = \left[ \angle {\mathcal{M}_1 (\mathbf{x})} , \ldots, \angle {\mathcal{M}_{n_{\mathcal{O}}} (\mathbf{x})} \right]$.
For i.i.d. proper complex errors with $\bm{\Sigma}_{\mathbf{w}_{\varepsilon}\mathbf{w}_{\varepsilon}} = \bm{\Sigma}_{\bm{\varphi}_{\varepsilon}\bm{\varphi}_{\varepsilon}} = \frac{\beta_{\mathcal{O}}^{-1}}{2} \mathbf{I}_{n_\mathcal{O}}$, we obtain
\begin{equation}
	\obj (\mathbf{x}) = 
	n_\mathcal{O} \ln \beta_\mathcal{O}
	-  \beta_\mathcal{O} \sum_{i=1}^{n_\mathcal{O}} \abs{\ln \mathbf{y}_{\mathcal{O},i} - \ln \mathbf{y}_{\mathcal{M},i}}^2 + \ln \pdf \left( \mathbf{x} \right) \, .
	\label{eq:map_obj_iid}
\end{equation}
We observe that the evaluation of the objective function $\obj (\mathbf{x})$ of Eqs.~\eqref{eq:map_obj} or \eqref{eq:map_obj_iid} requires computing the model prediction of the $n_\mathcal{O}$ models.
Whenever this requires significant computing resources, solving the optimization problem in Eq.~\eqref{eq:def_map} using standard techniques becomes challenging.
Therefore, we subsequently propose a Bayesian optimization strategy that utilizes RPCE surrogate models.
To this end, we introduce the RPCE model and propose a novel Bayesian learning method for estimating its coefficients in the next section.
\subsection{Rational polynomial chaos expansion}
\label{subsec:rpce}
The proposed updating strategy aims to utilize surrogate models for the individual model responses that enter the likelihood function of Eq.~\eqref{eq:likeliii}.
This likelihood depends on a numerical model of the systems' frequency response, which exhibit a rational type of dependency on the input parameters $\mathbf{X}$ \cite{Geradin.1997}.
It was shown in \cite{Jacquelin.2015, Schneider.2019} that standard PCE surrogate models converge slowly and suffer from poor generalization when applied to frequency response function data.
A natural modification to standard PCE is to consider rational PCE models, which are constructed through taking the ratio of two PCEs.
To this end, we approximate the set of models $\{ \mathcal{M}_i \left( \mathbf{X} \right) | i = 1, \ldots, n_{\mathcal{O}} \}$ through a set of surrogate models $\{ \mathcal{S}_i \left( \mathbf{X} \right) | i = 1, \ldots, n_{\mathcal{O}} \}$, for which, ideally, $\mathcal{S}_i \left( \mathbf{X} \right) \approx \mathcal{M}_i \left( \mathbf{X} \right)$.
\subsubsection{Definition of the Rational Polynomial Chaos Expansion}
As discussed above, assume that $\mathbf{X}$ is a random vector with outcome space $\mathbb{R}^d$ and given joint probability density function that models the uncertain input parameters of the numerical model $Y_{\mathcal{M},i} (\mathbf{X})$. 
Without loss of generality, we assume that the random vector $\mathbf{X}$ follows the independent standard Gaussian distribution. 
If $\mathbf{X}$ follows a non-Gaussian distribution, it is possible to express $Y_{\mathcal{M},i}$ as a function of an underlying independent standard Gaussian vector through an isoprobabilistic transformation \cite{rosenblatt1952remarks}. 
For the sake of simplicity, we drop the dependency on the index $i$ and treat scalar valued models in the following.
Vector-valued models, as introduced above, are treated by surrogating each entry in the output random vector individually.

Let $P \left( \mathbf{X} \right)$ and $Q \left( \mathbf{X} \right)$ be truncated polynomial chaos representations, such that
\begin{equation}
	P \left( \mathbf{X}; \mathbf{p} \right) = \sum_{i=0}^{n_{\scriptscriptstyle p}-1} p_i \Psi_{p,i} \left( \mathbf{X} \right) \, , \qquad Q \left( \mathbf{X}; \mathbf{q} \right) = \sum_{i=0}^{n_{\scriptscriptstyle q}-1} q_i \Psi_{q,i} \left( \mathbf{X} \right) .
	\label{eq:Hermite_pade}
\end{equation}
Here $\{p_i \in \mathbb{C},i=0,\ldots,n_p-1\}$ and $\{q_i \in \mathbb{C}, i=0, \ldots, n_q-1\}$ are complex coefficients and $\Psi_{p,i}$ and $\Psi_{q,i}$ are the multivariate orthonormal (probabilist) Hermite polynomials. 
The sets $\{\Psi_{p,i}, i = 0,\ldots, n_p \}$ and $\{ \Psi_{q,i}, i = 0,\ldots, n_q \}$ are constructed through the $d$-fold tensorization of the univariate normalized Hermite polynomials, i.e.,
\begin{equation}
	\Psi_{\mathbf{a}_p} = \prod_{i=1}^{d} \psi_{a_{p,i}} (X_i) \, , \qquad \Psi_{\mathbf{a}_q} = \prod_{i=1}^{d} \psi_{a_{q,i}} (X_i) \, .
\end{equation}
In here, $\mathbf{a}_p \in \mathbb{N}^d$ and $\mathbf{a}_q \in \mathbb{N}^d$ denote the index sets of the corresponding multivariate polynomials.
Different strategies can be applied to obtain the index sets, out of which the total degree and the hyperbolic truncation scheme \cite{Blatman.2011} are popular choices.
In the total degree (TD) truncation, we retain all polynomials with a total polynomial degree less than or equal to $m_p$ or $m_q$, i.e., 
\begin{equation}
	\sum_{i=1}^{d} a_{p,i} \leq m_p \, , \qquad \sum_{i=1}^{d} a_{q,i} \leq m_q \, .
	\label{eq:total_degree_trunc}
\end{equation}
The resulting number of polynomial terms in the total degree truncation are $n_p = \binom{d+m_p}{m_p}$ and $n_q = \binom{d+m_q}{m_q}$.
The truncated set of multivariate polynomials is finally sorted in the lexicographic order \cite{cox2007ideals}.
The truncation rules are separately applied to both, numerator and denominator polynomial, with maximum polynomial degrees $m_p$ and $m_q$.

We define the RPCE $R (\mathbf{X})$ obtained by taking the ratio of the two PCE representations of Eq.~\eqref{eq:Hermite_pade}:
\begin{equation}
	R (\mathbf{X}; \mathbf{p}, \mathbf{q}) := \frac{P \left( \mathbf{X}; \mathbf{p} \right)}{Q \left( \mathbf{X}; \mathbf{q} \right)} = \frac{\sum_{i=0}^{n_{\scriptscriptstyle p}-1} p_i \Psi_{p,i} \left( \mathbf{X} \right)}{ \sum_{i=0}^{n_{\scriptscriptstyle q}-1} q_i \Psi_{q,i} \left( \mathbf{X} \right)} .
	\label{eq:ratio_rep}
\end{equation}
Stochastic collocation \cite{Chantrasmi.2009,Schneider.2019} and Galerkin \cite{JacquelinE.2016} methods to determine the coefficients $\mathbf{p}$ and $\mathbf{q}$ in the expansions in Eq.~\eqref{eq:ratio_rep} have been presented in the literature.
Recently, a sparse Bayesian learning approach was presented in \cite{Schneider.2023} that allows to learn a sparse and probabilistic representation of the coefficient vectors.
Therein, a two-stage strategy is proposed that utilizes the fact that the RPCE is linear in terms of the numerator coefficients. 
This allows to express the posterior distribution of the numerator coefficients, conditional on the denominator coefficients, in closed-form.
The posterior distribution of the denominator coefficients is then approximated through a Dirac at its MAP value.
The method efficiently tackles the overfitting problem and significantly reduces the number of model evaluations compared to the least-squares regression approach in \cite{Schneider.2019}.
In \cite{Schneider.2023}, the likelihood of the expansion coefficients given a set of saples from $\mathbf{X}$ and corresponding model evaluations is derived based on an assumption on the model misfit $\varepsilon_{\mathcal{S}}$, defined as
\begin{equation}
	\varepsilon_{\mathcal{S}} = \mathcal{M} (\mathbf{X}) - R (\mathbf{X}; \mathbf{p} , \mathbf{q}) \, .
	\label{eq:residual_standard}
\end{equation}
It can be observed that the error in Eq.~\eqref{eq:residual_standard} is non-linear with respect to the denominator coefficients in $\mathbf{q}$. 
In the context of least-squares regression, minimizing the error requires an iterative solution procedure.
We can, however, adjust the error $\varepsilon_{\mathcal{S}}$, by multiplying Eq.~\eqref{eq:residual_standard} with the denominator polynomial, which gives the augmented error
\begin{equation}
	\Tilde{\varepsilon}_{\mathcal{S}} = Q (\mathbf{X}) \varepsilon_{\mathcal{S}} = Q \left( \mathbf{X}; \mathbf{q} \right)  \mathcal{M} (\mathbf{X}) - P \left( \mathbf{X}; \mathbf{p} \right) \, .
	\label{eq:residual_lin}
\end{equation}
The augmented error $\Tilde{\varepsilon}_{\mathcal{S}}$ is linear with respect to the denominator coefficients. 
In the following we will develop a Bayesian regression approach for learning the coefficients in the RPCE based on assuming a normal distribution for $\varepsilon_{\mathcal{S}}$.
\subsubsection{Training the RPCE through sparse Bayesian learning}
\label{sec:training}
For training the model, we assume that a data set $\mathcal{D}_\mathcal{M} = \{ (\mathbf{x}^{(k)},  m^{(k)} ) | k = 1, \ldots, \ntrain \}$ of input samples $\mathbf{x}^{(k)}$ and corresponding model evaluations $m^{(k)} = \mathcal{M} \left( \mathbf{x}^{(k)} \right)$ is available. 
We apply a Bayesian approach to learn the coefficients $\mathbf{p}$ and $\mathbf{q}$, i.e., we treat them as random variables and estimate their posterior distribution, expressed through Bayes' rule as
\begin{equation}
	\pdf (\mathbf{p},\mathbf{q}|\mathbf{m}) = c_{\mathcal{S}}^{-1} \pdf (\mathbf{m}|\mathbf{p},\mathbf{q}) \pdf (\mathbf{p},\mathbf{q}) \, ,
\end{equation}

where $\pdf (\mathbf{m}|\mathbf{p},\mathbf{q})$ is the likelihood function and $\pdf (\mathbf{p},\mathbf{q})$ is the prior distribution. 
The likelihood function is derived based on a Gaussian assumption on the augmented error of Eq.~\eqref{eq:residual_lin}.
We denote by $\mathbf{m} = \left[ m^{(1)}; \ldots; m^{(\ntrain)} \right] \in \mathbb{C}^{\ntrain \times 1}$ the vector of model responses.
Then, $\Tilde{\varepsilon}_{\mathcal{S},k}$ denotes the augmented error at the $k$-th sample location, i.e., $\Tilde{\varepsilon}_{\mathcal{S},k} =  Q \left(\mathbf{x}^{(k)}; \mathbf{q} \right) m^{(k)} - P \left(\mathbf{x}^{(k)}; \mathbf{p}\right)$ and 
$\Tilde{\bm{\varepsilon}}_{\mathcal{S}} = \left[ \Tilde{\varepsilon}_{\mathcal{S},1}; \ldots; \Tilde{\varepsilon}_{\mathcal{S},\ntrain} \right]$ is a random vector with outcome space $\mathbb{C}^{\ntrain \times 1}$ that collects all errors:
\begin{equation}
	\Tilde{\bm{\varepsilon}}_{\mathcal{S}} =  \diag \left( \bm{\Psi}_q \mathbf{q} \right) \mathbf{m} - \bm{\Psi}_p \mathbf{p} \, ,
\end{equation}
where $\mathbf{\Psi}_p \in \mathbb{R}^{\ntrain\times n_p}$ and $\mathbf{\Psi}_q \in \mathbb{R}^{\ntrain\times n_q}$ have as $(k,j)$-elements $\Psi_{p,j} (\mathbf{x}^{(k)})$ and $\Psi_{q,j} (\mathbf{x}^{(k)})$.
We assume that the augmented errors $\Tilde{\bm{\varepsilon}}_{\mathcal{S}}$ are jointly complex Gaussian with zero mean and covariance matrix $\bm{\Sigma}_{\Tilde{\bm{\varepsilon}}_{\mathcal{S}} \Tilde{\bm{\varepsilon}}_{\mathcal{S}}} = \beta_{\mathcal{S}}^{-1} \mathbf{I}_{\ntrain}$, where $\mathbf{I}_{n}$ denotes the $n \times n$ identity matrix. 
Note that, since $\varepsilon_{\mathcal{S},k} = \nicefrac{\Tilde{\varepsilon}_{\mathcal{S},k}}{Q (\mathbf{x}^{(k)}; \mathbf{q})}$, the resulting covariance matrix for the vector of standard errors is
\begin{equation}
	\bm{\Sigma}_{\bm{\varepsilon}_{\mathcal{S}} \bm{\varepsilon}_{\mathcal{S}}} = \beta_{\mathcal{S}}^{-1} \mathbf{Q}^{-1} \mathbf{Q}^{-H} \, ,
	\label{eq:cov_eps_standard}
\end{equation}
where $\mathbf{Q} := \diag(\bm{\Psi}_q \mathbf{q})$.
The conditional expectation and covariance matrix for the data vector $\mathbf{m}$ follow as
\begin{gather}
	\E \left[ \mathbf{m} | \mathbf{p}, \mathbf{q} \right] = \mathbf{Q}^{-1} \bm{\Psi}_p \mathbf{p} \, \label{eq:mom1_y_aug} \, ,\\
	\Cov \left[ \mathbf{m} | \mathbf{p}, \mathbf{q} \right] = \bm{\Sigma}_{\bm{\varepsilon} \bm{\varepsilon}} = \beta_{\mathcal{S}}^{-1} \mathbf{Q}^{-1} \mathbf{Q}^{-H} = \beta_{\mathcal{S}}^{-1} \diag \left( \abs{Q \left(\mathbf{x}^{(1)}; \mathbf{q} \right)}^{-2}, \ldots, \abs{Q \left(\mathbf{x}^{(\ntrain)}; \mathbf{q}\right)}^{-2} \right) \, . \label{eq:mom2_y_aug}
\end{gather}
We note that the mean of the data is the same as the one of the standard residual formulation used in the sparse Bayesian regression approach of \cite{Schneider.2023}, whereas the variance of the data is weighted with the inverse of the square-magnitude of the denominator polynomial at the sample locations. 
\begin{figure}[thb]
	\centering
	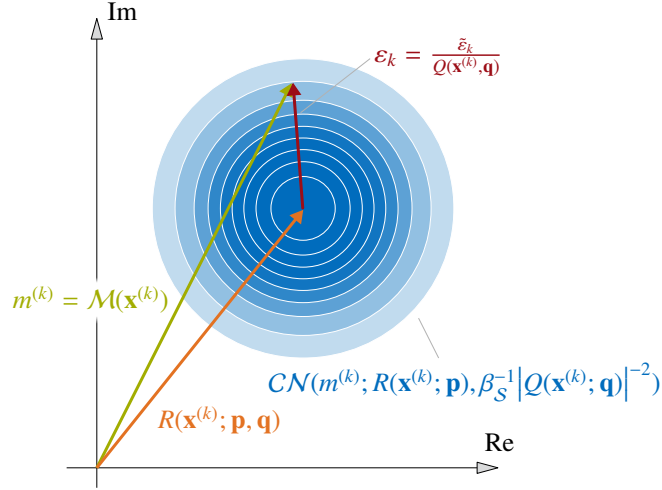
	\caption{
		Illustration of the model error in the complex plane that is defined to derive the likelihood function.
		We assume an additive error $\Tilde{\varepsilon}_k$ between the product of original model response and denominator polynomial $Q(\mathbf{x}^{(k)}) \mathcal{M} (\mathbf{x}^{(k)})$  and the numerator polynomial $P(\mathbf{x}^{(k)})$. 
	}
	\label{fig:illust_error}
\end{figure}
Since $\mathbf{m}$ depends linearly on $\Tilde{\bm{\varepsilon}}_{\mathcal{S}}$, the likelihood will be complex Gaussian with moments as in Eqs.~\eqref{eq:mom1_y_aug} and \eqref{eq:mom2_y_aug}.
Then,
\begin{alignat}{2}
	\pdf (\mathbf{m} | \mathbf{p}, \mathbf{q}) & = \left( \frac{\beta_{\mathcal{S}}}{\pi} \right)^{\ntrain} \det \left( \mathbf{Q}^{\mathrm{H}} \mathbf{Q} \right) \exp{- \beta_{\mathcal{S}} \left( \mathbf{m} - \mathbf{Q}^{-1} \bm{\Psi}_p \mathbf{p} \right)^{\mathrm{H}} \mathbf{Q}^{\mathrm{H}} \mathbf{Q} \left( \mathbf{m} - \mathbf{Q}^{-1} \bm{\Psi}_p \mathbf{p} \right)} 
	\label{eq:likeli_RAT_LINRES} \\
	& = \left( \frac{\beta_{\mathcal{S}}}{\pi} \right)^{\ntrain} \det \left( \mathbf{Q}^{\mathrm{H}} \mathbf{Q} \right) \exp{- \beta_{\mathcal{S}} \left( \mathbf{Q} \mathbf{m} - \bm{\Psi}_p \mathbf{p} \right)^{\mathrm{H}} \left( \mathbf{Q} \mathbf{m} - \bm{\Psi}_p \mathbf{p} \right)} \, . 
	\label{eq:likeli_RAT_LINRES2}
\end{alignat}
The error model and the involved relationship between the model and the surrogate are illustrated in Fig.~\ref{fig:illust_error}.
We employ independent priors for the numerator and denominator coefficients. 
Following the approach in \cite{Tipping.2001,Schneider.2023}, the prior distribution for the numerator coefficients is modeled as a zero mean complex proper Gaussian distribution, i.e.,
\begin{gather}
	\pdf (\mathbf{p} | \bm{\alpha}_p) = \mathcal{CN} ( \mathbf{p} | \mathbf{0} , \bm{\Lambda}_{\mathbf{p}\mathbf{p}}^{-1}, \mathbf{0}) \, , \label{eq:prior_p} 
\end{gather}
where $\bm{\Lambda}_{\mathbf{p}\mathbf{p}} = \diag{ \bm{\alpha}_p }$ constitutes the numerator coefficients precision matrix and $\bm{\alpha}_p= \left[ \alpha_{p,1}; \ldots ; \alpha_{p,n_p} \right]$ is a vector containing the $n_p$ hyperparameters (precisions).
Since $\mathbf{p}$ and $\mathbf{m}$ given $\mathbf{q}$ are jointly proper complex  Gaussian, the conditional distribution of the numerator coefficients $\mathbf{p}$ given $\mathbf{m}$ and $\mathbf{q}$ will also be proper complex Gaussian.
Through combining Eqs.~\eqref{eq:likeli_RAT_LINRES2} and \eqref{eq:prior_p}, after a few transformations and application of the Woodbury identity, we obtain the posterior mean and covariance matrix of the numerator coefficients as
\begin{gather}
	\bm{\mu}_{\mathbf{p}|\mathbf{m}, \mathbf{q}} = \beta_{\mathcal{S}} \bm{\Sigma}_{\mathbf{p}\mathbf{p}|\mathbf{m}} \bm{\Psi}_p^{\mathrm{H}} \underbrace{\mathbf{Q} \mathbf{m}}_{:= \Tilde{\mathbf{m}}} \, ,
	\label{eq:mean_p_lin} \\
	\bm{\Sigma}_{\mathbf{p}\mathbf{p}|\mathbf{m}} = \left( \bm{\Lambda}_{\mathbf{p}\mathbf{p}} + \beta_{\mathcal{S}} \bm{\Psi}_p^{\mathrm{H}} \bm{\Psi}_p \right)^{-1} \, , \label{eq:cov_p_lin} 
\end{gather}
where by $\Tilde{\mathbf{m}} := \mathbf{Q} \mathbf{m} = \left[ Q \left( \mathbf{x}^{(1)} \right) y^{(1)} ; \ldots; Q \left( \mathbf{x}^{(n_{\mathcal{O}})} \right) y^{(n_{\mathcal{O}})} \right]$ we denote the augmented data vector. 
In contrast to the formulation in \cite{Schneider.2023}, the posterior conditional covariance matrix of the numerator coefficients given $\mathbf{q}$ does not depend on $\mathbf{q}$.
Thus, the posterior distribution of the numerator coefficients, conditional on the denominator coefficients, is proper complex Gaussian with
\begin{equation}
	\pdf (\mathbf{p} | \mathbf{m}, \mathbf{q}, \bm{\alpha}_p, \beta_{\mathcal{S}}) = \frac{1}{\pi^{n_P} \det \bm{\Sigma}_{\mathbf{p}\mathbf{p}|\mathbf{m}}} \exp \left( - \left( \mathbf{p} - \bm{\mu}_{\mathbf{p}|\mathbf{m}, \mathbf{q}} \right)^\mathrm{H} \bm{\Sigma}_{\mathbf{p}\mathbf{p}|\mathbf{m}}^{-1} \left( \mathbf{p} - \bm{\mu}_{\mathbf{p}|\mathbf{m}, \mathbf{q}} \right) \right) \, .
	\label{eq:post_p_cond_q}
\end{equation}
The marginal likelihood of the data given $\mathbf{q}$ is then obtained as
\begin{gather}
	\pdf (\mathbf{m} | \mathbf{q}, \bm{\alpha}_p, \beta_{\mathcal{S}}) = \frac{\det \left( \mathbf{Q} \mathbf{Q}^\mathrm{H} \right)}{\pi^N \det \left( \mathbf{C} \right)} \exp{- \mathbf{m}^{\mathrm{H}} \mathbf{Q}^\mathrm{H} \mathbf{C}^{-1} \mathbf{Q} \mathbf{m}} \, , \label{eq:marg_likelihood_y_cond_q}
\end{gather}
where
\begin{equation}
	\mathbf{C} = \beta_{\mathcal{S}}^{-1} \mathbf{I}_{\ntrain} + \bm{\Psi}_p \bm{\Lambda}_{\mathbf{p}\mathbf{p}}^{-1} \bm{\Psi}_p^{\mathrm{T}} \, .
	\label{eq:evi_C_matrix}
\end{equation}
In analogy to Eq.~\eqref{eq:prior_p}, the prior distribution of the denominator coefficients is assumed to be a zero mean proper complex Gaussian distribution, i.e., 
\begin{gather}
	\pdf (\mathbf{q} | \bm{\alpha}_q) = \mathcal{CN} ( \mathbf{q} | \mathbf{0} , \bm{\Lambda}_{\mathbf{q}\mathbf{q}}^{-1}, \mathbf{0}) \, . \label{eq:prior_q}
\end{gather}
where $\bm{\Lambda}_{\mathbf{q}\mathbf{q}} = \diag{ \bm{\alpha}_q }$ constitutes the denominator coefficients precision matrix and $\bm{\alpha}_q= \left[ \alpha_{q,1}; \ldots ; \alpha_{q,n_q} \right]$ is a vector containing the $n_q$ hyperparameters (precisions).

The outlined formulation does not permit a closed-form solution for the posterior distribution of $\mathbf{q}$ since the computation of the model evidence 
\begin{equation}
	\pdf (\mathbf{m} | \bm{\alpha}_p, \bm{\alpha}_q, \beta_{\mathcal{S}}) = \int_{\mathbb{C}^{n_q}} f \left(\mathbf{m} | \mathbf{q}, \bm{\alpha}_p, \beta_{\mathcal{S}} \right) f \left( \mathbf{q} | \bm{\alpha}_q)\right) \dd{\mathbf{q}} \, ,
	\label{eq:laplace_evidence}
\end{equation}
cannot be performed in closed-form. 
We therefore choose to approximate the posterior distribution of $\mathbf{q}$ through a proper complex Gaussian distribution (cf. Appendix B.1 in \cite{Schneider.2023} and references therein) centered at the maximum a posteriori (MAP) estimate of $\mathbf{q}$.
In the analysis of real-valued random variables, this Gaussian approximation is commonly referred to as \textit{Laplace}'s approximation.
Note that choosing a proper Gaussian distribution as an approximation neglects the off-diagonal elements in the full Hessian matrix \cite{Schreier.2010}.
This will restrain the posterior distribution of the denominator coefficients such that the real and imaginary part of a coefficient $q_i$ will share the same variance.
Subsequently, we find a set of hyperparameters that maximizes the approximate model evidence, which is known as type-II-maximum likelihood estimation.
This strategy leads to a sequential two-stage approach for the approximation of the posterior distribution of the denominator coefficients as well as the estimation of the hyperparameters. 

In a first step, the MAP estimate of the denominator coefficients $\mathbf{q}^\ast$ is found through solving the following optimization problem
\begin{equation}
	\mathbf{q}^\ast = \argmax_{\mathbf{q} \in \mathbb{C}^{n_q}} \pdf (\mathbf{m} | \mathbf{q}, \bm{\alpha}_p, \beta_{\mathcal{S}} ) \pdf ( \mathbf{q} | \bm{\alpha}_q) \, .
	\label{eq:q_MAP}
\end{equation}
Then, the posterior distribution of the denominator coefficients is approximated through
\begin{equation}
	\pdf (\mathbf{q} | \mathbf{m}, \bm{\alpha}_p, \bm{\alpha}_q, \beta_{\mathcal{S}} ) \approx \frac{1}{\pi^{n_q} \det \left(- \left(  \mathbf{H}_{\mathbf{q}\mathbf{q}} \right)^{-1} \right)} \exp{- \left( \mathbf{q} - \mathbf{q}^\ast \right)^{\mathrm{H}} \left( - \mathbf{H}_{\mathbf{q}\mathbf{q}} \right) \left( \mathbf{q} - \mathbf{q}^\ast \right)} \, ,
	\label{eq:laplace_approx}
\end{equation}
where $\mathbf{H}_{\mathbf{q}\mathbf{q}}$ denotes the $\mathbb{C}$-complex Hessian \cite{Schreier.2010} of the log-posterior function of the denominator coefficients. 
We note that $\mathbf{H}_{\mathbf{q}\mathbf{q}}$ does not depend on the denominator coefficients and, hence, we do not explicitly state the evaluation point $\mathbf{q} = \mathbf{q}^\ast$.
Utilizing the approximation in Eq.~\eqref{eq:laplace_approx}, the model evidence can be approximated through
\begin{equation}
	\pdf (\mathbf{m} | \bm{\alpha}_p, \bm{\alpha}_q, \beta_{\mathcal{S}}) \approx f \left( \mathbf{m} | \mathbf{q}^\ast, \bm{\alpha}_p, \beta_{\mathcal{S}} \right) f \left( \mathbf{q}^\ast | \bm{\alpha}_q \right) \pi^{n_q} \det \left(- \left(  \mathbf{H}_{\mathbf{q}\mathbf{q}} \right)^{-1} \right) \, .
	\label{eq:laplace_evidence_approx2}
\end{equation}
An optimal set of hyperparameters is then found through type-II-maximum likelihood estimation.
To this end, we maximize the approximate model evidence in Eq.~\eqref{eq:laplace_evidence_approx2} over all hyperparameters, such that
\begin{equation}
	[ \bm{\alpha}_p^\ast, \bm{\alpha}_q^\ast, \beta_{\mathcal{S}}^\ast ] = \argmax_{\underset{\in \mathbb{R}^{n_p \times n_q \times 1}}{[ \bm{\alpha}_p, \bm{\alpha}_q, \beta_{\mathcal{S}} ]}} \pdf ( \mathbf{m} | \mathbf{q}^\ast, \bm{\alpha}_p, \beta_{\mathcal{S}}) \pdf(\mathbf{q}^\ast | \bm{\alpha}_q ) \det \left(- \left(  \mathbf{H}_{\mathbf{q}\mathbf{q}} \right)^{-1} \right) \, .
	\label{eq:hyper_MAP}
\end{equation}
In the following, we summarize the resulting expressions in each of the above steps.
Further detailed derivations are given in \ref{app:grad_q} to \ref{app:grad_beta}.

In order to find the MAP estimate of $\mathbf{q}$, we maximize the log of the objective function in Eq.~\eqref{eq:q_MAP}, which inserting Eqs.~\eqref{eq:marg_likelihood_y_cond_q} and \eqref{eq:prior_q} into Eq.~\eqref{eq:q_MAP} and taking the logarithm results in
\begin{equation}
	\mathbf{q}^\ast = \argmax_{\mathbf{q} \in \mathbb{C}^{n_q}} \left[ \ln \det \mathbf{Q} \mathbf{Q}^{\mathrm{H}} - \mathbf{q}^{\mathrm{H}} \left( \bm{\Upsilon}^{\mathrm{H}} \mathbf{C}^{-1} \bm{\Upsilon} + \bm{\Lambda}_{\mathbf{q}\mathbf{q}} \right) \mathbf{q} \right] \, , \label{eq:q_MAP2}
\end{equation}
where $\bm{\Upsilon} = \diag \left( \mathbf{m} \right) \bm{\Psi}_q \in \mathbb{C}^{\ntrain \times n_q}$.
Eq.~\eqref{eq:q_MAP2} is a nonlinear optimization problem in complex variables.
In order to solve Eq.~\eqref{eq:q_MAP2}, we resort to a gradient based maximization technique. 
Since $f( \mathbf{m} | \mathbf{q}, \bm{\alpha}_p, \beta_{\mathcal{S}}) f(\mathbf{q} | \bm{\alpha}_q )$ is the product of two PDFs, it is real-valued and thus a necessary condition for the objective function in Eq.~\eqref{eq:q_MAP2} to take a maximum is given by
\begin{equation}
	\pdv{\conj{\mathbf{q}}} \left[ 
	\ln \det \mathbf{Q} \mathbf{Q}^{\mathrm{H}} - \mathbf{q}^{\mathrm{H}} \left( \bm{\Upsilon}^{\mathrm{H}} \mathbf{C}^{-1} \bm{\Upsilon} + \bm{\Lambda}_{\mathbf{q}\mathbf{q}} \right) \mathbf{q} \right] = \mathbf{0}_{n_q} \, , 
	\label{eq:deriv_def_conj_q}
\end{equation}
where $\pdv{\conj{\mathbf{q}}}$ denotes the generalized (or Wirtinger) derivative with respect to the complex conjugate of the denominator coefficients $\mathbf{q}$, denoted as the conjugate cogradient.
The definition of the generalized derivatives can be found in \cite{Schreier.2010,kreutz2009complex}.
Since the objective function is real-valued, it is sufficient to consider only the conjugate cogradient, since in this case it holds $\pdv{f( \mathbf{q} ) }{\mathbf{q}} = \conj{\pdv{f( \mathbf{q} ) }{\conj{\mathbf{q}}}}$.
In order to solve Eq.~\eqref{eq:q_MAP2}, we employ a quasi-Newton method and use a limited memory Broyden-Fletcher-Goldfarb-Shanno (L-BFGS) algorithm, provided by \cite{Sorber.2013}.
Details about the algorithm can be found in \cite{Sorber.2012}.
The algorithm uses a quasi-Newton step to update an approximation of the Hessian matrix of the problem in each iteration.
We use the available line-search algorithm in the implementation by \cite{Sorber.2013}.

The quasi-Newton method requires the derivatives of the objective function with respect to the conjugate denominator coefficients.
The derivative in Eq.~\eqref{eq:deriv_def_conj_q} can be found analytically and reads
\begin{alignat}{2}
	\pdv{\conj{\mathbf{q}}} \left[ 
	\ln \det \mathbf{Q} \mathbf{Q}^{\mathrm{H}} - \mathbf{q}^{\mathrm{H}} \left( \bm{\Upsilon}^{\mathrm{H}} \mathbf{C}^{-1} \bm{\Upsilon} + \bm{\Lambda}_{\mathbf{q}\mathbf{q}} \right) \mathbf{q} \right] & = \bm{\Psi}_q^{\mathrm{T}} \left( \bm{\Psi}_q \conj{\mathbf{q}} \right)^{\circ-1} - \left( \bm{\Upsilon}^{\mathrm{H}} \mathbf{C}^{-1} \bm{\Upsilon} + \bm{\Lambda}_{\mathbf{q}\mathbf{q}} \right) \mathbf{q} \label{eq:pdiv_q_1} \, , 
	%
\end{alignat}
where $\left( \cdot \right)^{\circ-1}$ denotes the element-wise inverse.
The next step in the outlined sequential approach requires the computation of the Hessian $\mathbf{H}_{\mathbf{q}\mathbf{q}}$ of the log-posterior.
We find
\begin{equation}
	\mathbf{H}_{\mathbf{q}\mathbf{q}} = \pdv{\mathbf{q}} \left( \pdv{\conj{\mathbf{q}}} \left[ 
	\ln \det \mathbf{Q} \mathbf{Q}^{\mathrm{H}} - \mathbf{q}^{\mathrm{H}} \left( \bm{\Upsilon}^{\mathrm{H}} \mathbf{C}^{-1} \bm{\Upsilon} + \bm{\Lambda}_{\mathbf{q}\mathbf{q}} \right) \mathbf{q} \right] \right)^\mathrm{T} = - \left( \bm{\Upsilon}^{\mathrm{H}} \mathbf{C}^{-1} \bm{\Upsilon} + \bm{\Lambda}_{\mathbf{q}\mathbf{q}} \right) \, . \label{eq:hessian}
\end{equation}
We derive the partial derivatives and Hessian matrices in \ref{app:grad_q}.

After having computed Laplace's approximation for the denominator coefficients, we can express Eq.~\eqref{eq:hyper_MAP} using Eqs.~\eqref{eq:marg_likelihood_y_cond_q}, \eqref{eq:prior_q} and \eqref{eq:hessian} as 
\begin{multline}
	[ \bm{\alpha}_p^\ast, \bm{\alpha}_q^\ast, \beta_{\mathcal{S}}^\ast ] = \argmax_{\underset{\in \mathbb{R}^{n_p \times n_q \times 1}}{[ \bm{\alpha}_p, \bm{\alpha}_q, \beta_{\mathcal{S}} ]}} \left[
	\ntrain \ln \beta_{\mathcal{S}} 
	+ \ln \det \bm{\Sigma}_{\mathbf{p}\mathbf{p}| \mathbf{m}} 
	+ \ln \det \bm{\Lambda}_{\mathbf{p}\mathbf{p}} 
	+ \ln \det \bm{\Lambda}_{\mathbf{q}\mathbf{q}} -
	\right. \\ 
	\left. 
	- \beta_{\mathcal{S}} \mathbf{m}^{\mathrm{H}} \conj{\mathbf{Q}^\ast} \left( \mathbf{Q}^\ast \mathbf{m} - \bm{\Psi}_p \bm{\mu}_{\mathbf{p}|\mathbf{m},, \mathbf{q}^\ast} \right)  
	- \mathbf{q}^{\ast \mathrm{H}} \bm{\Lambda}_{\mathbf{q} \mathbf{q}} \mathbf{q}^{\ast}  
	- \ln \det \left( \bm{\Upsilon}^{\mathrm{H}}  \mathbf{C}^{-1} \bm{\Upsilon} + \bm{\Lambda}_{\mathbf{q} \mathbf{q}} \right)
	\right] \, ,
	\label{eq:hyper_MAP2}
\end{multline}
where $\mathbf{Q}^\ast := \diag \left( \bm{\Psi}_q \mathbf{q}^\ast \right)$ denotes the diagonal matrix of denominator predictions evaluated for the MAP of $\mathbf{q}$. 
Here, we have used the Woodbury identity to express the inverse and determinant of $\mathbf{C}$ as%
\begin{gather}
	\mathbf{C}^{-1} = \beta_{\mathcal{S}} \mathbf{I}_{\ntrain} - \beta_{\mathcal{S}}^2 \bm{\Psi}_p \bm{\Sigma}_{\mathbf{p}\mathbf{p}|\mathbf{m}} \bm{\Psi}_p^{\mathrm{T}} \, , \label{eq:CTilde_inverse_woodb} \\
	\det \mathbf{C} = \det \bm{\Sigma}_{\mathbf{p}\mathbf{p}|\mathbf{m}}^{-1} \det \bm{\Lambda}_{\mathbf{p}\mathbf{p}}^{-1} \det \beta_{\mathcal{S}}^{-1} \mathbf{I}_{\ntrain} \, .
\end{gather}
The derivatives of the objective function in Eq.~\eqref{eq:hyper_MAP2} are derived in \ref{app:grad_alpha_p}, \ref{app:grad_alpha_q}, and \ref{app:grad_beta}.
Based on these, we obtain the following update rules for the hyperparameters:
\begin{gather}
	\alpha_{p,i} = \frac{1}{\left[ \bm{\Sigma}_{\mathbf{p}\mathbf{p}|\mathbf{m}} \right]_{ii} + \abs{\left[ \bm{\mu}_{\mathbf{p}|\mathbf{m}, \mathbf{q}^\ast} \right]_i}^2 + \left[ \bm{\Delta}^\mathrm{H} \left( - \mathbf{H}_{\mathbf{q}\mathbf{q}} \right)^{-1} \bm{\Delta} \right]_{ii}} \, , \label{eq:upd_rule_alpha_p}\\
	%
	\alpha_{q,i} = \frac{1}{\abs{q_i^\ast}^2 + \left[ \left( - \mathbf{H}_{\mathbf{q}\mathbf{q}} \right)^{-1} \right]_{ii}} \, , \label{eq:upd_rule_alpha_q}\\
	\beta_{\mathcal{S}} = \frac{N}{\norm{\mathbf{Q} \mathbf{m} - \bm{\Psi}_p \bm{\mu}_{\mathbf{p}|\mathbf{m}, \mathbf{q}^\ast}}^2 + \tr \left( \bm{\Sigma}_{\mathbf{p}\mathbf{p}|\mathbf{m}} \bm{\Psi}_p^{\mathrm{T}} \bm{\Psi}_p \right) + \tr \left( \left( - \mathbf{H}_{\mathbf{q}\mathbf{q}} \right)^{-1} \mathbf{O}^{\mathrm{H}} \mathbf{O} \right)} \, , \label{eq:upd_rule_beta}
\end{gather}
where $\bm{\Delta} = \beta_{\mathcal{S}} \bm{\Upsilon}^{\mathrm{H}} \bm{\Psi}_p \bm{\Sigma}_{\mathbf{p}\mathbf{p}|\mathbf{m}}$ and $\mathbf{O} = \left( \mathbf{I}_{n_\mathcal{M}} - \beta_{\mathcal{S}} \bm{\Psi}_p \bm{\Sigma}_{\mathbf{p}\mathbf{p}|\mathbf{m}} \bm{\Psi}_p^{\mathrm{T}} \right) \bm{\Upsilon}$.
The model coefficients are sequentially updated in an iterative manner until convergence is reached. 
Convergence is controlled through observation of the change of the hyperparameters.
Whenever their change in log-scale drops below a certain, user-specified threshold, the algorithm is terminated and the current state of coefficients and hyperparameters is utilized further.
The algorithm is summarized in \ref{alg:SBLRA} and straightforwardly follows the one for the standard residual. We refer to \cite{Schneider.2023} for further details.
\begin{algorithm}
	\caption{Sparse Bayesian RPCE training} 
	\label{alg:SBLRA}
	\begin{algorithmic}
		\Require The maximum number of iterations, $i_{\max}$, the tolerances for the convergence criteria $\varepsilon_\alpha$ and $\varepsilon_\beta$, the pruning thresholds $\alpha_{p,\max}$ and $\alpha_{q,\max}$, the initial coefficients $ \mathbf{p}_{\mathrm{init}}$ and $\mathbf{q}_{\mathrm{init}}$, $k$ used in $k$-normalization of $\mathbf{q}$
		\While{$i < i_{\max} \land (\max(\Delta \log \bm{\alpha}) > \varepsilon_\alpha \lor \Delta \log \beta > \varepsilon_\beta)$}
		\State Find useful numerator weights $p_i$ with $\alpha_{p,i}$ lower than threshold $\alpha_{p,\max}$, prune all other basis functions.
		\State Update denominator coefficients $\mathbf{q}$ with normalized MAP-estimate, $\mathbf{q} \leftarrow \mathbf{q}^{\ast}\cdot \norm{\mathbf{q}^{\ast}}_k^{-1}$, by solving Eq.~\eqref{eq:q_MAP}, and compute Hessian matrix $\mathbf{H}_{\mathbf{q}\mathbf{q}}$ using Eq.~\eqref{eq:hessian}.
		\State Find useful denominator coefficients $q_i$ with $\alpha_{q,i}$ lower than threshold $\alpha_{q,\max}$, prune all other basis functions.
		\State Update posterior mean and covariance for numerator coefficients, $\bm{p} \leftarrow \bm{\mu}_{\mathbf{p}|\mathbf{m}, \mathbf{q}^\ast}$,  $\bm{\Sigma} \leftarrow \bm{\Sigma}_{\mathbf{p}\mathbf{p}|\mathbf{m}}$ using Eqs.~\eqref{eq:cov_p_lin} and \eqref{eq:mean_p_lin}.
		\State Update $\bm{\alpha}_p$, $\bm{\alpha}_q$ and $\beta$ using the update rules in Eqs.~\eqref{eq:upd_rule_alpha_p}, \eqref{eq:upd_rule_alpha_q} and \eqref{eq:upd_rule_beta}
		\State Update maximum change in hyperparameters $\Delta \log \alpha_{\max,i} \leftarrow \Delta \log \alpha_{\max,i+1}$ and $\Delta \log \beta_{\mathcal{S},i} \leftarrow \Delta \log \beta_{\mathcal{S},i+1}$
		\EndWhile \\
		\Return Retained coefficients $\mathbf{p}$ and $\mathbf{q}$
	\end{algorithmic} 
\end{algorithm}

\begin{figure}[!hbt]
	\centering
	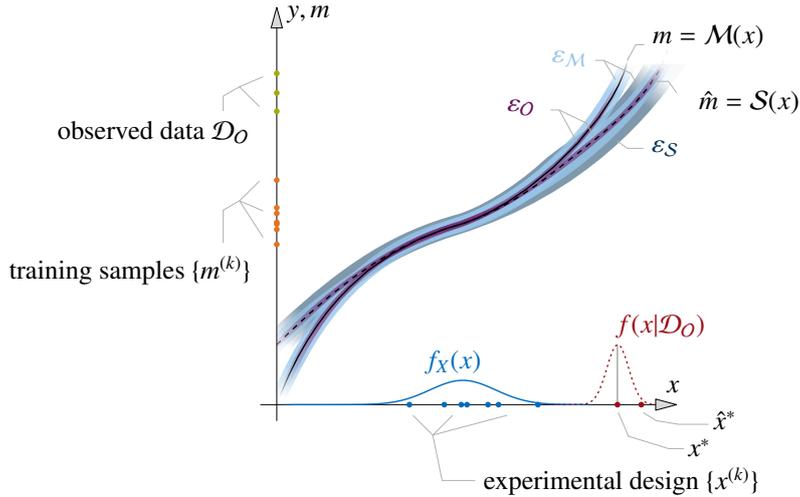
	\caption{Illustration of the inference procedure using a surrogate model. 
		The solid line represents the original model $\mathcal{M}(\mathbf{x})$.
		Based on the prior distribution $f(x)$, a set of input samples $\{x^{(k)}\}$, the experimental design, is generated.
		The corresponding model output samples, the training set, is computed as $\mathcal{M}(x^{(k)})$.
		The trained surrogate model $\hat{\mathcal{S}}(x^{(k)})$ can usually be expected to show good accuracy in the bulk of the prior distribution.
		For the inverse relationship, the model output and the observed data are related via the model error $\varepsilon_{\mathcal{M}}$ and the additive observation noise $\varepsilon_{\mathcal{O}}$.
		If the surrogate model is used in the inverse problem, an additional random error $\varepsilon_{\mathcal{S}}$ can be considered.
		If now data $\mathcal{D}_{\mathcal{O}}$ is observed significantly distant from the training samples, the surrogate model accuracy in the corresponding region in the input space might be poor. 
		Thus, the inferred samples, or the MAP estimate, might be prone to significant error.
		This motivates the development of the active learning procedure to place experimental design samples towards the bulk of the posterior distribution.
		Remark: We disregard the probabilistic nature of the Bayesian regression surrogate model in this illustration for reasons of simplicity.
		The surrogate model response is really a random process over $x$.
	}
	\label{fig:sampling_forw_inv}
\end{figure}
Until now we have assumed that the training procedure is performed using a fixed-size experimental design.
Such design can be obtained through samples following the prior probability distribution of the input parameters.
This approach works well in settings where the rational surrogate model is subsequently used for forward uncertainty propagation applications based on the prior distributions, where one is interested in computing the probability distribution of a model output or function thereof based on a given distribution of the input random variables.
In this case, the training samples will align with the sought-after distribution and the surrogate will possess good predictability in the relevant part of the domain of the output space.
In the inverse setting, introduced in Sec.~\ref{subsec:bu_par_upd}, however, the location of the bulk of the posterior probability mass is typically unknown a priori.
Whenever the generalization capability of the surrogate model is poor or the bulk of the posterior probability mass lies far from the bulk of the prior probability mass, this approach might introduce significant surrogate model errors.
A graphical illustration of this situation is given in Fig.~\ref{fig:sampling_forw_inv}.
This problem can be tackled through active, sequential experimental design strategies, where the experimental design is sequentially enriched based on exploiting information in the already observed sample data.
A powerful framework that builds towards this goal is Bayesian Optimization.
We subsequently propose an active learning approach for  training the introduced Bayesian rational surrogate model in the context of MAP estimation that draws upon Bayesian Optimization.
\subsection{Active, sequential design through Bayesian optimization}
\label{subsec:bayes_opt}
Our goal is the computation of the MAP value as outlined in Section~\ref{subsec:bu_par_upd}.
In order to reduce the computational cost of the evaluation of the objective function, defined in Eq.~\eqref{eq:map_obj}, we replace the set of original models $\{\mathcal{M}_i\}$ with their surrogate model counterparts $\{\mathcal{S}_i\}$.
Due to the fact that we are working with frequency domain models, the RPCE surrogate model, as introduced in Section~\ref{subsec:rpce}, is a specifically well-suited approximation.
To this end, each of the individual models that enter the objective function is approximated through an RPCE, i.e., $\mathcal{S}_i (\mathbf{x}) = R (\mathbf{x}; \mathbf{p}_i, \mathbf{q}_i)$.
For the $n_{\mathcal{O}}$ surrogate models, we collect the coefficients in the matrix $\mathbf{R} = \left[ \mathbf{p}_1, \ldots, \mathbf{p}_{n_{\mathcal{O}}}, \mathbf{q}_1, \ldots, \mathbf{q}_{n_{\mathcal{O}}} \right]$, where $\mathbf{p}_i$ and $\mathbf{q}_i$ are the vectors of numerator and denominator coefficients of the $i$-th surrogate model.
Under consideration of the surrogate model error, as defined in Eq.~\eqref{eq:residual_standard}, we can replace the objective function in Eq.~\eqref{eq:map_obj} by
\begin{multline}
	\hat{\obj} (\mathbf{x}; \mathbf{R}, \bm{\varepsilon}_\mathcal{S}) = \frac{1}{2} \left[
	- \ln \det \bm{\Sigma}_{\mathbf{w}_{\varepsilon}\mathbf{w}_{\varepsilon}} 
	- \ln \det \bm{\Sigma}_{\bm{\varphi}_{\varepsilon}\bm{\varphi}_{\varepsilon}} 
	- \norm{\vphantom{\bm{\varphi}_\mathcal{O}}\mathbf{w}_{\mathcal{O}} - \left( \mathbf{w}_{\mathcal{S}} \left( \mathbf{x}; \mathbf{R} \right) + \ln \abs{ \bm{\varepsilon}_\mathcal{S} }\right)}_{\bm{\Sigma}_{\mathbf{w}_{\varepsilon}\mathbf{w}_{\varepsilon}}^{-1}}^2
	\right. \\ \left.
	- \norm{ \bm{\varphi}_{\mathcal{O}} - \left( \bm{\varphi}_{\mathcal{S}} \left(\mathbf{x}; \mathbf{R} \right) + \angle\bm{\varepsilon}_\mathcal{S} \right)}_{\bm{\Sigma}_{\bm{\varphi}_{\varepsilon}\bm{\varphi}_{\varepsilon}}^{-1}}^2
	\right]
	+ \ln \pdf \left( \mathbf{x} \right) \, ,
	\label{eq:map_obj_surr}
\end{multline}
where 
\begin{gather}
	\mathbf{w}_\mathcal{S} \left( \mathbf{x}; \mathbf{R} \right) = \left[ \ln | y_{\mathcal{S},1} \left( \mathbf{x}; \mathbf{p}_1, \mathbf{q}_1 \right) | ; \ldots;  \ln | y_{\mathcal{S},{n_{\mathcal{O}}}} \left( \mathbf{x}; \mathbf{p}_{n_{\mathcal{O}}}, \mathbf{q}_{n_{\mathcal{O}}} \right) | \right] \, , \\
	\bm{\varphi}_\mathcal{S} \left( \mathbf{x}; \mathbf{R} \right) = \left[ \angle y_{\mathcal{S},1} \left( \mathbf{x}; \mathbf{p}_1, \mathbf{q}_1 \right) ; \ldots;  \angle y_{\mathcal{S},{n_{\mathcal{O}}}} \left( \mathbf{x}; \mathbf{p}_{n_{\mathcal{O}}}, \mathbf{q}_{n_{\mathcal{O}}} \right) \right] \, , \\
	\bm{\varepsilon}_\mathcal{S} = \left[ \varepsilon_{\mathcal{S},1} ; \ldots;  \varepsilon_{\mathcal{S},{n_{\mathcal{O}}}} \right] \, ,
\end{gather}
and $\varepsilon_{\mathcal{S},i}$ is the surrogate model error of the $i$-th RPCE model as defined in Eq.~\eqref{eq:residual_standard} that has covariance matrix as given in Eq.~\eqref{eq:cov_eps_standard} with precision parameter $\beta_{\mathcal{S},i}$.
We note that the objective function in Eq.~\eqref{eq:map_obj_surr}, $\hat{\obj} (\mathbf{x}; \mathbf{R}, \bm{\varepsilon}_\mathcal{S})$, now becomes a random process over the model input parameters that is expressed in terms of the random variables in $\mathbf{R}$ and $\bm{\varepsilon}_\mathcal{S}$.
Therefore, we treat the problem as a Bayesian optimization problem. 
In the context of Bayesian optimization the key idea is now to formulate the optimization problem as a Bayesian decision problem, which, based on a user-specified acquisition function, allows to adaptively sample the experimental design.
The acquisition function can be understood through an underlying utility function in the Bayesian decision making framework \cite{garnett_bayesoptbook_2023, Shahriari.2016}.
Subsequently, we comprehensively summarize the framework.

\begin{figure}[!hbt]
	\centering
	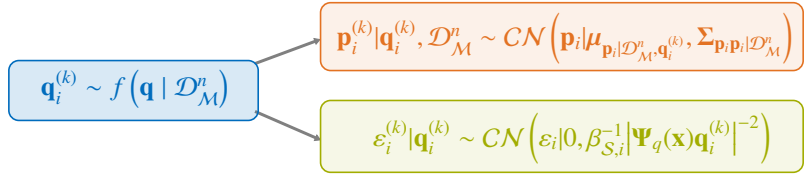
	\caption{
		Illustration of the posterior sampling and the hierarchical dependency. 
	}
	\label{fig:post_samples}
\end{figure}

We denote by $\mathcal{D}_\mathcal{M}^1 = \left\{ \left( \mathbf{x}^{(k)}, \mathbf{y}_\mathcal{M} \left(\mathbf{x}^{(k)}\right) \right) | k = 1, \ldots , n_{\mathrm{init}} \right\}$ the initial training set with $n_{\mathrm{init}}$ number of samples. 
Therein, it holds $\mathbf{x}^{(k)} \in \mathbb{R}^{d}$ and $\mathbf{y}_\mathcal{M} \left(\mathbf{x}^{(k)}\right) \in \mathbb{C}^{n_\mathcal{O}}$.
Throughout the active learning procedure the data set will be enriched.
In iteration step $n$, the observed data set up to this step is denoted $\mathcal{D}_\mathcal{M}^n$.
The expected improvement acquisition is then defined as
\begin{equation}
	\alpha_\mathrm{EI} (\mathbf{x}; \mathcal{D}_\mathcal{M}^n) = \mathbb{E}_{\hat{\obj} | \mathcal{D}_\mathcal{M}^n} \left[ \left( \hat{\obj} (\mathbf{x}; \mathbf{R}, \bm{\varepsilon}_\mathcal{S}) - \obj_{\max} , 0 \right) \right] \, ,
	\label{eq:ei_crit}
\end{equation}
where $\obj_{\max}$ denotes the currently observed maximum objective value in iteration step $n$, i.e., 
\begin{equation}
	\obj_{\max} = \max_{\mathbf{x} \in \left\{ \mathbf{x}^{(k)} | k = 1, \ldots , n_{\mathrm{init}} + n \right\}} \obj \left(\mathbf{x}\right) \, .
\end{equation}
The expectation in Eq.~\eqref{eq:ei_crit} is stated with respect to the distribution of the approximate objective $\hat{\obj}$, which is unknown.
The randomness in $\hat{\obj}$ for a given $\mathbf{x}$ is due to its dependence on random variables $\mathbf{R}$ and $\bm{\varepsilon}_{\mathcal{S}}$, hence, the expectation can be equivalently stated with respect to these random variables. 
Since the expectation in Eq.~\eqref{eq:ei_crit} cannot be solved in closed form, we resort to a sampling-based approximation. 
To this end, we approximate the expected improvement by the arithmetic mean as follows
\begin{equation}
	\alpha_\mathrm{EI} (\mathbf{x}; \mathcal{D}_\mathcal{M}^n) \approx \hat \alpha_\mathrm{EI} (\mathbf{x}; \mathcal{D}_\mathcal{M}^n) = \frac{1}{n_{\alpha}} \sum_{k = 1}^{n_{\alpha}} \max \left( \hat{\obj} (\mathbf{x}; \mathbf{R}^{(k)}, \bm{\varepsilon}_{\mathcal{S}}^{(k)} ) - \obj_{\max} , 0 \right) \, ,
	\label{eq:ei_crit_approx}
\end{equation}
where $\mathbf{R}^{(k)}$ contains samples of the RPCE coefficients and $\bm{\varepsilon}_{\mathcal{S}}^{(k)}$ denotes a sample of the surrogate model error vector. 
All samples are conditional on the available training points in the data set $\mathcal{D}_\mathcal{M}^n$ and the hyperparameters identified in the training process described in Section~\ref{sec:training}.
We generate samples of the denominator coefficients of each surrogate model based on the identified hyperparameters after termination of the training using a transitional Markov Chain Monte Carlo \cite{Ching.2007,Betz.2016} sampler.
Conditional on the denominator coefficients, the numerator coefficients will follow a complex normal distribution with mean vector and covariance matrix as defined in Eqs.~\eqref{eq:mean_p_lin} and \eqref{eq:cov_p_lin}.
The surrogate model error, conditional on the denominator coefficients, also follows a complex normal distribution with covariance matrix as defined in Eq.~\eqref{eq:cov_eps_standard}.
This hierarchical dependency is illustrated in Fig.~\ref{fig:post_samples}.

For numerical purposes, we approximate the maximum function in the expected improvement criterion through the softplus function, i.e., $ \max \left( x, 0 \right) \approx \softplus(x)$.
The softplus function is defined through
\begin{equation}
	\softplus(x) = \gamma^{-1} \log \left( 1 + \exp \left( \gamma x \right) \right) \, .
\end{equation}
This approximation is often encountered in machine learning algorithms to remove the discontinuity in the derivative of the maximum function \cite{NEURIPS2023_419f72cb}.
Since the proposed approach approximates the expected value in Eq.~\eqref{eq:ei_crit} through the sample mean, the acquisition function will yield zeros for great portions of the input space.
Relaxing the maximum function to the softplus function introduces continuous gradients in the acquisition function, which leads to a more robust identification of the optimum of the acquisition function. 

Finally, the data set $\mathcal{D}_\mathcal{M}^{n+1}$ is enriched by the data point that maximizes the acquisition function, i.e.,
\begin{equation}
	\mathbf{x}^+ = \argmax_{\mathbf{x} \in \mathbb{R}^d} \hat \alpha_\mathrm{EI} (\mathbf{x}; \mathcal{D}_\mathcal{M}^n) \, .
\end{equation}
The data set in the next iteration step is then $\mathcal{D}_\mathcal{M}^{n+1} = \mathcal{D}_\mathcal{M}^n \cup \left\{ \left( \mathbf{x}^+ , \mathbf{y}_\mathcal{M} (\mathbf{x}^+) \right) \right\}$.
Subsequently, the set of surrogate models is retrained using the updated data set $\mathcal{D}_\mathcal{M}^{n+1}$.
The algorithm is terminated once a preset number of model function evaluations $n_{\mathrm{budget}}$ is reached.
Subsequently, we consider two possibilities to determine the final optimizer $\mathbf{x}^\ast$ after the algorithm has terminated, following the presentation in \cite{garnett_bayesoptbook_2023}.
The first estimator, termed the \textit{simple reward estimator}, selects as $\mathbf{x}^\ast$ the point $\mathbf{x}$ with highest objective function evaluation among the observed data points, i.e., 
\begin{equation}
	\mathbf{x}_{\mathrm{SR}}^\ast = \argmax_{\mathbf{x} \in \left\{ \mathbf{x}^{(k)} | k = 1, \ldots , n_{\mathrm{budget}} \right\}} \obj \left(\mathbf{x}\right) \, .
	\label{eq:simple_reward}
\end{equation}
This estimator thus only considers points for which the original model has been evaluated.
The second estimator, termed the \textit{global reward estimator}, is based on optimizing the objective function utilizing the random process surrogate model.
To this end, a deterministic estimate of the random process model $\hat{\obj} (\mathbf{x})$ has to be chosen as a proxy in the optimization. 
A common choice is to use the mean of $\hat{\obj} (\mathbf{x})$.
This estimator thereby utilizes the additional information contained in the surrogate model and should be chosen when there is trust in the surrogate model prediction away from the yet observed data points.
Since evaluating the expected value of $\hat{\obj} (\mathbf{x})$ requires sampling in our case, we use the conditional mean of the numerator coefficients and the MAP estimate of the denominator coefficients as plug-in estimates and solve
\begin{equation}
	\mathbf{x}_{\mathrm{GR}}^\ast = \argmax_{\mathbf{x} \in \mathbb{R}^d} \hat{\obj} \left( \mathbf{x}; \mathbf{R}^\ast, \mathbf{0}_{n_\mathcal{O}} \right) \, .
	\label{eq:global_reward}
\end{equation}
where $\mathbf{R}^\ast = \left[ \bm{\mu}_{\mathbf{p}_1| \mathcal{D}_{\mathcal{M}}^{n_{\mathrm{budget}}}, \mathbf{q}_1^\ast}, \ldots, \bm{\mu}_{\mathbf{p}_{n_\mathcal{O}} | \mathcal{D}_{\mathcal{M}}^{n_{\mathrm{budget}}}, \mathbf{q}_{n_\mathcal{O}}^\ast}, \mathbf{q}_1^\ast, \ldots, \mathbf{q}_{n_\mathcal{O}}^\ast \right]$. 
Furthermore, the surrogate model error mean $\E \left[\bm{\varepsilon}_{\mathcal{S}} \right] = \mathbf{0}_{n_\mathcal{O}}$ is chosen in the estimation of the global reward estimator.
The full algorithm is summarized in Alg.~\ref{alg:BO-MAP}.
\begin{algorithm}
	\caption{Active, sequential design through Bayesian optimization} 
	\label{alg:BO-MAP}
	\begin{algorithmic}
		\Require The maximum number of model evaluations, $n_{\mathrm{budget}}$, the initial number of model evaluation $n_{\mathrm{init}}$, the initial training data $\mathcal{D}_{\mathcal{M}}^{1}$, the observation data, 
		\State Set $n=1$
		\While{$n_{\mathrm{init}} + n \leq n_{\mathrm{budget}}$}
		\State Find the next sample point $\mathbf{x}^+$ as the optimizer of the expected improvement acquisition function in Eq.~\eqref{eq:ei_crit_approx} using the surrogate models in $\{\mathcal{S}_i\}$. 
		For this, generate samples of the posterior distribution of the surrogate models as illustrated in Fig.~\ref{fig:post_samples}.
		\State Evaluate the $n_{\mathcal{O}}$ models in $\{\mathcal{M}_i\}$ to yield $\mathbf{y}_\mathcal{M} (\mathbf{x}^+)$.
		\State Enrich the data set: $\mathcal{D}_\mathcal{M}^{n+1} = \mathcal{D}_\mathcal{M}^n \cup \left\{ \left( \mathbf{x}^+ , \mathbf{y}_\mathcal{M} (\mathbf{x}^+) \right) \right\}$.
		\State Retrain the $n_{\mathcal{O}}$ surrogate models in $\{\mathcal{S}_i\}$ according to Algorithm~\ref{alg:SBLRA} using the enriched data set $\mathcal{D}_\mathcal{M}^{n+1}$.
		\EndWhile \\
		\Return MAP estimate $\mathbf{x}^\ast$, according to Eqs.~\eqref{eq:simple_reward} or \eqref{eq:global_reward}.
	\end{algorithmic} 
\end{algorithm}
In the following section, we will apply the proposed Bayesian updating strategy to the problem of Bayesian parameter updating for two structural dynamic models.
\section{Numerical examples}
\label{sec:num_examples}
We apply the methodology outlined in Section~\ref{sec:method} to two examples. 
For illustrative purposes, we investigate updating the parameters of a two degree of freedom system with varying numbers of input parameters in Section~\ref{subsec:2dof}.
The second example, presented in Section \ref{subsec:fem}, includes updating the parameters of the finite element model of a cross-laminated timber plate.

In order to assess the performance of the proposed approach, we compare the results obtained from the active learning procedure with a reference solution obtained using the original model as well as a surrogate model based solution, where a fixed experimental design with size equal to the number of samples in the active learning data set in each iteration step is used. 
Throughout all application examples we use the particle swarm algorithm \cite{Kennedy.1995} for the acquisition function maximization.
\subsection{Algebraic model: two-degree-of-freedom system}
\label{subsec:2dof}
\begin{figure}
	\centering
	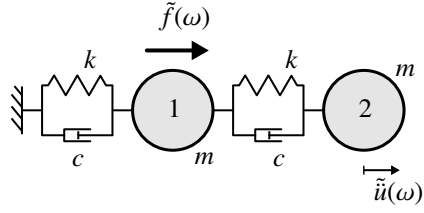
	\caption{
		Illustration of two-degree-of-freedom system with fully dependent spring, mass and damping coefficient.
		The acceleration $\ddot u$ is measured at the second mass, while the frequency response is considered with respect to the first mass with forcing $f$.
		The frequency response is then $h_{21} (\omega) = \frac{\Tilde{\ddot u} (\omega)}{\Tilde{f} (\omega)} = - \omega^2\frac{\Tilde{u} (\omega)}{\Tilde{f} (\omega)}$ (compare with Eq.~\eqref{eq:2dof_appl_frf}). 
	}
	\label{fig:2dof}
\end{figure}

In this example we consider the Bayesian parameter updating problem for a two-degree-of-freedom system. 
It consists of two lumped masses that are interconnected by a spring and damper element. 
We assume that the properties of the system can be modelled through a single set of stiffness, damping and mass values.
This way, the individual spring, mass and damping coefficients become fully dependent.
The system is illustrated in Fig.~\ref{fig:2dof}.
In the following, we investigate the proposed approach for a varying number of random input parameters:
\begin{enumerate}
	\item Stiffness $k$ random, mass $m$ and damping $c$ deterministic and equal to their prior mean value; i.e., $\mathbf{X} = k$,
	\item Stiffness $k$, mass $m$ and damping $c$ random; i.e., $\mathbf{X} = \left[ k, m, c \right]$.
\end{enumerate}
The system parameters' prior distributions are assumed as summarized in Table~\ref{tab:2dof_par_assumpt}.
\begin{table}[!htb]
	\footnotesize
	\centering
	\caption{Prior distribution assumptions for two-degree-of-freedom model. The parameters are assumed to be independent.}
	\begin{tabular}{lcccc}
		\toprule
		Parameter & & Distribution & Mean value $\mu_{(\cdot)}$ & Coefficient of variation $\delta_{(\cdot)}$ \\
		\midrule
		Stiffness & $k$ & Lognormal & $4\cdot 10^6 \, \mathrm{\frac{N}{m}}$ & $0.2$ \\[1mm]
		Mass & $m$ & Lognormal & $3\cdot 10^2 \, \mathrm{kg}$ & $0.2$ \\[1mm]
		Damping coefficient & $c$ & Lognormal & $2\cdot 10^3 \, \mathrm{\frac{N}{sm}}$ & $0.2$ \\
		\bottomrule
	\end{tabular}
	\label{tab:2dof_par_assumpt}
\end{table}
The damping ratio obtained for the mean parameter values follows to $\zeta = \frac{1}{2} \frac{\mu_c}{\sqrt{\mu_k \mu_m}} = 0.0289$. 
The system can thus be considered as lightly damped.
We consider acceleration measurements of the frequency response function at the right mass 2 due to force excitation at the left mass 1 (cf. Fig.~\ref{fig:2dof}).
The data is considered for the frequency set $\{ 10 \, \mathrm{Hz}, 11 \, \mathrm{Hz}, 12 \, \mathrm{Hz}, 28 \, \mathrm{Hz}, 30 \, \mathrm{Hz}, 32 \, \mathrm{Hz} \}$, thus, $n_\mathcal{O} = 6$.
The frequency response function is modeled through
\begin{equation}
	h_{21} (\omega, \mathbf{x}) = - \omega^2 \frac{\Tilde{u} (\omega)}{\Tilde{f} (\omega)} = -\omega^2
	\begin{bmatrix}
		0 & 1 
	\end{bmatrix}
	\left(
	\begin{bmatrix}
		2 k & -k \\ -k & k 
	\end{bmatrix}
	+ \mathrm{i} \omega 
	\begin{bmatrix}
		2 c & -c \\ -c & c 
	\end{bmatrix}
	- \omega^2
	\begin{bmatrix}
		m & 0 \\ 0 & m 
	\end{bmatrix}
	\right)^{-1}
	\begin{bmatrix}
		1 \\ 0 
	\end{bmatrix} \, ,
	\label{eq:2dof_appl_frf}
\end{equation}
where $\Tilde{u} (\omega)$ and $\Tilde{f} (\omega)$ are illustrated in Fig.~\ref{fig:2dof}.
Throughout this section, the polynomial orders are chosen as $m_p = m_q = 2$ and total degree truncation is applied. 
\subsubsection{One Random Parameter}
\label{subsubsec:appl_2dof_1dim}
For illustrative purposes, we first consider a single random parameter.
To this end, we choose the stiffness parameter $k$,  synthetically generate measurement data using Eq.~\eqref{eq:2dof_appl_frf} for the value $k_{\mathrm{true}} = 0.5 \mu_k = 2\cdot 10^6 \, \mathrm{\frac{N}{m}}$ and add samples according to the noise model outlined in Section~\ref{subsec:bu_par_upd} with i.i.d. proper complex errors using $\beta_{\mathcal{O}} = 10^{2}$.

\begin{figure}[!hbt]
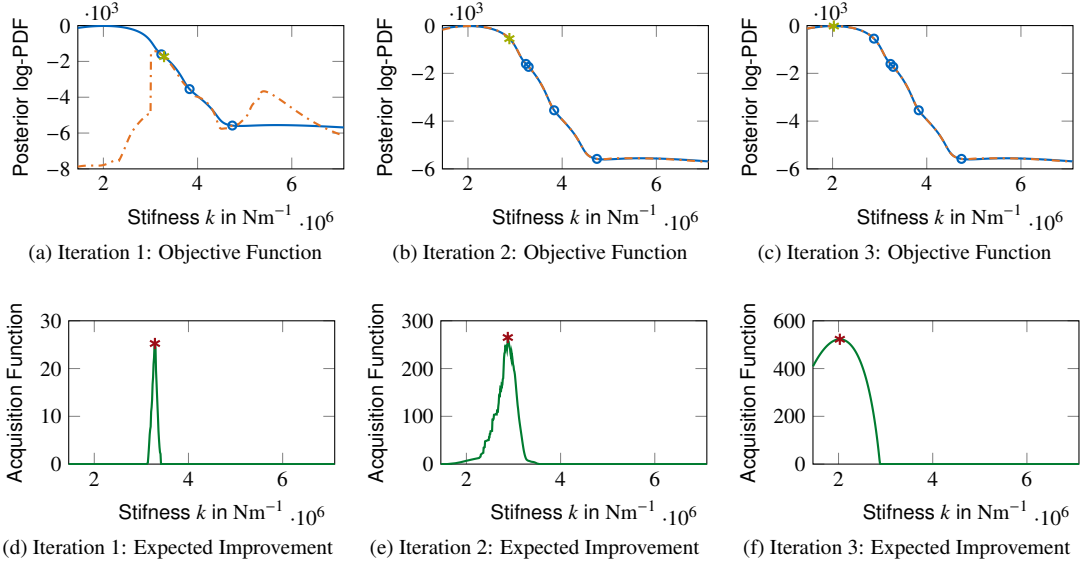

	\centering
	\subfloat[Iteration 1: Objective Function]{
		\input{figures/2dof_1dim/1dim_ego_ei_prMod_lognormal_nRep_1_nEI_100_nObs_6_nSw_100_nStall_20_betaObs_100_nInit_3_nBud_10_fSp_irr_mP_2_mQ_2_ut_softplus_1_wNoise_true_it_1_obj.tikz}
		\label{fig:obj_and_ei_2dof_1dim_obj_1}
	} \enspace
	\subfloat[Iteration 2: Objective Function]{
		\input{figures/2dof_1dim/1dim_ego_ei_prMod_lognormal_nRep_1_nEI_100_nObs_6_nSw_100_nStall_20_betaObs_100_nInit_3_nBud_10_fSp_irr_mP_2_mQ_2_ut_softplus_1_wNoise_true_it_2_obj.tikz}
	} \enspace
	\subfloat[Iteration 3: Objective Function]{
		\input{figures/2dof_1dim/1dim_ego_ei_prMod_lognormal_nRep_1_nEI_100_nObs_6_nSw_100_nStall_20_betaObs_100_nInit_3_nBud_10_fSp_irr_mP_2_mQ_2_ut_softplus_1_wNoise_true_it_3_obj.tikz}
		\label{fig:obj_and_ei_2dof_1dim_obj_3}
	} \\
	\subfloat[Iteration 1: Expected Improvement]{
		\input{figures/2dof_1dim/1dim_ego_ei_prMod_lognormal_nRep_1_nEI_100_nObs_6_nSw_100_nStall_20_betaObs_100_nInit_3_nBud_10_fSp_irr_mP_2_mQ_2_ut_softplus_1_wNoise_true_it_1_ei.tikz}
		\label{fig:obj_and_ei_2dof_1dim_acq_1}
	} \enspace
	\subfloat[Iteration 2: Expected Improvement]{
		\input{figures/2dof_1dim/1dim_ego_ei_prMod_lognormal_nRep_1_nEI_100_nObs_6_nSw_100_nStall_20_betaObs_100_nInit_3_nBud_10_fSp_irr_mP_2_mQ_2_ut_softplus_1_wNoise_true_it_2_ei.tikz}
	} \enspace
	\subfloat[Iteration 3: Expected Improvement]{
		\input{figures/2dof_1dim/1dim_ego_ei_prMod_lognormal_nRep_1_nEI_100_nObs_6_nSw_100_nStall_20_betaObs_100_nInit_3_nBud_10_fSp_irr_mP_2_mQ_2_ut_softplus_1_wNoise_true_it_3_ei.tikz}
		\label{fig:obj_and_ei_2dof_1dim_acq_3}
	} 
	\caption{
		Objective value and expected improvement over three iterations for MAP estimation of two-degree-of-freedom system with random stiffness parameter $k$.
		The left column shows the log-posterior $f(x)$ that is computed using the original model (\linefullblue) and the surrogate model based approximation (\linedashdottedorange). The latter is based on the adaptive experimental design obtained in the corresponding iteration step and the mean model coefficients. 
		In each iteration step, a new sample point (\AsteriskGreen) is added to the experimental design ({\color{TUMBlau}$\circ$}) based on maximizing the expected improvement criterion.
		The right column shows the expected improvement criterion (\linedasheddarkgreen) and its maximum value (\AsteriskRed). The identified parameter point will be added to the experimental design in the subsequent iteration step.}
	\label{fig:obj_and_ei_2dof_1dim}
\end{figure}

We illustrate the proposed procedure in Fig.~\ref{fig:obj_and_ei_2dof_1dim}. 
Therein the figures in the upper row (Figs.~\ref{fig:obj_and_ei_2dof_1dim_obj_1} to~\ref{fig:obj_and_ei_2dof_1dim_obj_3}) depict the objective function (\linefullblue) in the first three iteration steps as well as the approximation using the surrogate models (\linedashdottedorange).
The approximation is computed using the mean value estimates of the surrogate model coefficients conditional on the adaptive experimental design in the corresponding iteration step and can thus be considered a first-order approximation to the mean of the approximation of the objective function $\mu_{\hat{\obj}} (\mathbf{x})$.
The figures in the lower row (Figs.~\ref{fig:obj_and_ei_2dof_1dim_acq_1} to~\ref{fig:obj_and_ei_2dof_1dim_acq_3}) depict the evaluation of the acquisition function, namely the expected improvement, for each iteration step.
We start from an initial experimental design with $n_{\mathrm{init}} = 3$ samples ({\color{TUMBlau}$\circ$}) in Figs.~\ref{fig:obj_and_ei_2dof_1dim_obj_1} and \ref{fig:obj_and_ei_2dof_1dim_acq_1}. 
The maxima of the acquisition function in each iteration step are highlighted ({\color{TUMRed}$\ast$}). 
The corresponding sample point is added to the experimental design in the subsequent iteration step ({\color{TUMGruen}$\ast$}). 
One can observe how the algorithm generates samples that steadily move towards the global maximum of the log-posterior.
The first sample is placed closely to an initial sample point. 
Subsequently, the approximation of the log-posterior is highly accurate and the third added sample point is found in close proximity to the global optimum.
Upon termination, the algorithm recovers the MAP estimate accurately as $k^\ast = 2.8 \cdot 10^6 \, \mathrm{\frac{N}{m}}$, which is identical to the reference solution based on the original model.
\setlength\fwidth{3.5cm}
\setlength\fheight{1.9cm}
\subsubsection{Three Random Parameters}
In this section, we consider the two-degree-of-freedom system of Section~\ref{subsubsec:appl_2dof_1dim} and include all three random parameters in the MAP estimation.
The measurements are synthesized for $k_{\mathrm{true}} = 0.7 \mu_k = 2.8 \cdot 10^6 \, \mathrm{\frac{N}{m}}$, $m_{\mathrm{true}} = 1.5 \mu_m = 450 \, \mathrm{kg}$ and $c_{\mathrm{true}} = 1.5 \mu_c = 3 \cdot 10^3 \, \mathrm{\frac{N}{sm}}$ and samples are added according to the noise model outlined in Section~\ref{subsec:bu_par_upd} with i.i.d. proper complex errors using $\beta_{\mathcal{O}} = 10^{2}$.
\begin{table}[!htb]
	\footnotesize
	\centering
	\caption{Parameter estimates for two-degree-of-freedom model with three random parameters.}
	\begin{tabular}{lcccccccc}
		\toprule
		Parameter & & Units & \multicolumn{2}{c}{Posterior} & \multicolumn{4}{c}{MAP estimate} \\\cmidrule(lr){4-5}\cmidrule(lr){6-9}
		& & & mean & c.o.v. & reference & global reward & simple reward & fixed design \\
		\midrule
		Stiffness & $k$ & $10^6 \, \mathrm{\frac{N}{m}}$ & $2.77$ & $0.083$ & $2.7209$ & $2.7209$ & $2.7205$ & $2.7024$ \\[1mm]
		Mass & $m$ & $10^2 \, \mathrm{kg}$ & $4.45$ & $0.051$ & $4.4066$ & $4.4066$ & $4.4086$ & $4.3827$ \\[1mm]
		Damping & $c$ & $10^3 \, \mathrm{\frac{N}{sm}}$ & $2.61$ & $0.111$ & $2.5922$ & $2.5922$ & $2.6062$ & $2.5835$ \\[1mm]
		\bottomrule
	\end{tabular}
	\label{tab:2dof_ref_estimates_3dim}
\end{table}
Subsequently, we perform MAP-estimation through the proposed active learning approach.
In detail, starting from an initial experimental design generated through LHS, we sequentially add samples that maximize the expected improvement criterion.
Within each iteration step, the simple and global reward estimators of Eqs.~\eqref{eq:simple_reward} and \eqref{eq:global_reward} are computed.
Furthermore, in each iteration step, we generate an experimental design with sample size $\ntrain$ equal to the current number of samples in the active learning data set. 
We then identify the MAP value through maximization of the log-posterior, where we use the surrogate models that are trained on the fixed experimental design sample anew using LHS in each iteration step.
The fixed design solution is computed to investigate whether the improvement in the MAP estimate is due to an increasing number of samples in the training set or the design adaptivity.
To this end, we set $n_{\mathrm{init}} = 15$ and $n_{\mathrm{budget}} = 50$ for this example.
The analysis is repeated $n_{\mathrm{rep}} = 5$ times.
The expected improvement is estimated using $n_{\alpha} = 10^2$ samples.
The optimal expected improvement is found through particle swarm optimization \cite{Kennedy.1995} using the built-in algorithm in \matlab. 
The number of particles is chosen as $n_{\mathrm{sw}} = 500$. 
Smaller numbers of particles have been found to yield reasonable results as well.
A reference value for the MAP estimate is computed using particle swarm optimization.
In addition to the MAP estimation, we generate $n_{\mathrm{post}} = 10^5$ samples from the posterior distribution using aBUS-SuS \cite{Betz.2018}.
In order to quantitatively measure the performance of the proposed method, we compute the relative Euclidean distance of the estimates in the standard normal space $\hat{\mathbf{u}}_{\mathrm{MAP}}$ with respect to the reference MAP value $\mathbf{u}_{\mathrm{MAP}}$ that is obtained through a single optimization run using the original model, i.e., 
\begin{equation}
	\varepsilon_{\mathrm{MAP}} = \frac{\norm{\hat{\mathbf{u}}_{\mathrm{MAP}} - \mathbf{u}_{\mathrm{MAP}}}}{\norm{\mathbf{u}_{\mathrm{MAP}}}} \, .
\end{equation}
We choose to compare the distance in the standard normal space due to the fact that the parameters have significantly different orders of magnitude.
The posterior statistics and the median MAP estimates at $\ntrain = 50$ are summarized in Tab.~\ref{tab:2dof_ref_estimates_3dim}.
It can be observed that the active learning based estimate is accurate to four digits when compared to the reference MAP value. 
The resulting median relative error for $\ntrain = 50$ is $2.8 \cdot 10^{-5}$.
In contrast, the fixed design based median estimate differs slightly, resulting in a median error of $2.2 \cdot 10^{-2}$ for $\ntrain = 50$.
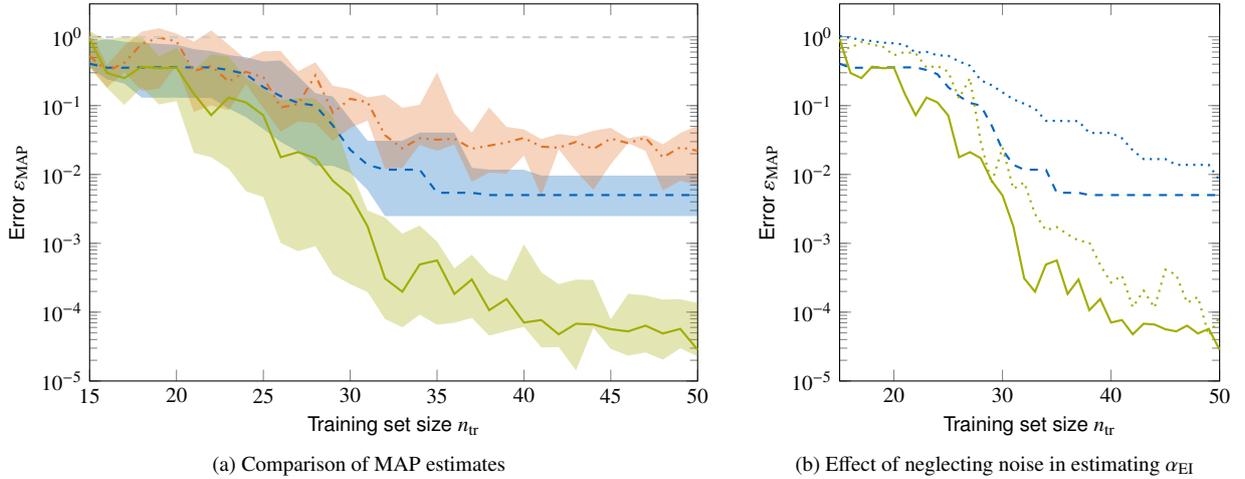
\begin{figure}[!hbt]
	\centering
	\setlength\fwidth{8cm}
	\setlength\fheight{5cm}
	\subfloat[Comparison of MAP estimates]{
%
\definecolor{mycolor1}{rgb}{0.00000,0.39608,0.74118}%
\definecolor{mycolor2}{rgb}{0.89020,0.44706,0.13333}%
\definecolor{mycolor3}{rgb}{0.63529,0.67843,0.00000}%
\begin{tikzpicture}

\begin{axis}[%
width=\fwidth,
height=\fheight,
at={(0cm,0cm)},
scale only axis,
xmin=15,
xmax=50,
xlabel={Training set size $\ntrain$},
ymode=log,
ymin=1e-05,
ymax=3,
yminorticks=true,
ylabel={Error $\varepsilon_{\mathrm{MAP}}$},
]

\addplot[area legend, draw=none, fill=mycolor1, fill opacity=0.3, forget plot]
table[row sep=crcr] {%
x	y\\
15	0.83042327394287\\
16	0.929587687821987\\
17	0.845925836415252\\
18	0.825229866263871\\
19	0.787136915965133\\
20	0.766687868810137\\
21	0.659761135659965\\
22	0.620185245945495\\
23	0.542349446448038\\
24	0.50436988832866\\
25	0.438480468319806\\
26	0.45167159079569\\
27	0.268231231813077\\
28	0.151389397812764\\
29	0.135527735726249\\
30	0.0640844653961889\\
31	0.0308497214010397\\
32	0.0308497214010397\\
33	0.0308497214010397\\
34	0.0405611934873218\\
35	0.0405611934873218\\
36	0.0405611934873218\\
37	0.0124964351914588\\
38	0.0117291675754141\\
39	0.0117291675754141\\
40	0.0117291675754141\\
41	0.00963437978614915\\
42	0.00963437978614915\\
43	0.00963437978614915\\
44	0.00963437978614915\\
45	0.00963437978614915\\
46	0.00963437978614915\\
47	0.00963437978614915\\
48	0.00963437978614915\\
49	0.00963437978614915\\
50	0.00963437978614915\\
50	0.00248739510472555\\
49	0.00248739510472555\\
48	0.00248739510472555\\
47	0.00248739510472555\\
46	0.00248739510472555\\
45	0.00248739510472555\\
44	0.00248739510472555\\
43	0.00248739510472555\\
42	0.00248739510472555\\
41	0.00248739510472555\\
40	0.00248739510472555\\
39	0.00248739510472555\\
38	0.00248739510472555\\
37	0.00248739510472555\\
36	0.00248739510472555\\
35	0.00248739510472555\\
34	0.00248739510472555\\
33	0.00248739510472555\\
32	0.00248739510472555\\
31	0.00596524436102397\\
30	0.0105867531258516\\
29	0.0133806881001726\\
28	0.0133806881001726\\
27	0.0214348318182852\\
26	0.0297186169010286\\
25	0.0455502576926505\\
24	0.067841090400286\\
23	0.0907235656789592\\
22	0.129333630983182\\
21	0.129333630983182\\
20	0.130354976545767\\
19	0.130354976545767\\
18	0.130354976545767\\
17	0.208801034196952\\
16	0.23999922699842\\
15	0.355035208671643\\
}--cycle;

\addplot[area legend, draw=none, fill=mycolor2, fill opacity=0.3, forget plot]
table[row sep=crcr] {%
x	y\\
15	1.25183086698267\\
16	0.389626757614601\\
17	0.741531594960629\\
18	1.18677781039385\\
19	1.3284941386722\\
20	1.11017574134893\\
21	0.851350149369847\\
22	1.2404390984826\\
23	0.767707646420308\\
24	0.468345886242175\\
25	0.614499581018592\\
26	0.621286938627832\\
27	0.314738693427319\\
28	0.427148445221796\\
29	0.194109449639855\\
30	0.174577062539342\\
31	0.131570320743715\\
32	0.144154451649212\\
33	0.0930237879434506\\
34	0.102910164859268\\
35	0.30505579393727\\
36	0.0772886252796985\\
37	0.0404568508280092\\
38	0.0940727450683245\\
39	0.049079415660627\\
40	0.0450878016620892\\
41	0.0326640933478267\\
42	0.0387494641487882\\
43	0.0359822199860223\\
44	0.0432328164584548\\
45	0.0474944037021912\\
46	0.0335663454272186\\
47	0.0370027323119421\\
48	0.0274130967572213\\
49	0.0399700212387887\\
50	0.051037694368178\\
50	0.00695959375843255\\
49	0.00809289059398189\\
48	0.00583926038901069\\
47	0.0219834375430999\\
46	0.0284586567841425\\
45	0.0140778710496747\\
44	0.00597250185092166\\
43	0.0111472654352343\\
42	0.0232376979111905\\
41	0.00487350510348705\\
40	0.0187639129722536\\
39	0.0102890004486392\\
38	0.0106619186984543\\
37	0.00780146247838317\\
36	0.0271848803611022\\
35	0.0219331508835763\\
34	0.0268087382549591\\
33	0.0124754159798067\\
32	0.0120859489050526\\
31	0.0606786530550944\\
30	0.0471015667813805\\
29	0.0308417339646103\\
28	0.0555961712489394\\
27	0.0590296428401548\\
26	0.0419728870941998\\
25	0.0919857702563869\\
24	0.204564015049795\\
23	0.0853903352674469\\
22	0.10030008370184\\
21	0.0808126953885906\\
20	0.38046592309215\\
19	0.176684403257114\\
18	0.164236283736242\\
17	0.302406273063825\\
16	0.182021627378608\\
15	0.369065097485698\\
}--cycle;

\addplot[area legend, draw=none, fill=mycolor3, fill opacity=0.3, forget plot]
table[row sep=crcr] {%
x	y\\
15	1.20664808784079\\
16	0.889732739290712\\
17	1.02790568367016\\
18	0.995765101550714\\
19	0.517266169513653\\
20	0.696886553202776\\
21	0.334622446546533\\
22	0.546246510868879\\
23	0.518138125081416\\
24	0.253067152904467\\
25	0.17514547730617\\
26	0.119212742720161\\
27	0.129280771151461\\
28	0.126956302579642\\
29	0.0697818720305265\\
30	0.0122262707436815\\
31	0.00300184967752052\\
32	0.0023702616662664\\
33	0.00137977542917107\\
34	0.0018115513448536\\
35	0.00105936061186488\\
36	0.000433754289621816\\
37	0.000682549163321688\\
38	0.000362732879396833\\
39	0.000283811131505864\\
40	0.000972076600628691\\
41	0.000378826812962771\\
42	0.000255594563672922\\
43	0.000297794135185282\\
44	0.000292633952787372\\
45	7.87548612059148e-05\\
46	0.000184442159054597\\
47	0.000175044852623811\\
48	0.000152270932959158\\
49	0.000154295476078485\\
50	0.000135910935740639\\
50	2.27992506437639e-05\\
49	2.98681952675509e-05\\
48	2.0426936417098e-05\\
47	2.59541614168662e-05\\
46	2.33691363115561e-05\\
45	2.9135085794801e-05\\
44	5.93001160334962e-05\\
43	1.42269827780601e-05\\
42	3.11788804114691e-05\\
41	3.06878752786963e-05\\
40	6.1103567803197e-05\\
39	5.30626677273898e-05\\
38	4.55500464046737e-05\\
37	8.24480745627281e-05\\
36	7.01090463184919e-05\\
35	6.55798714018689e-05\\
34	9.16719622894435e-05\\
33	6.0228449971655e-05\\
32	7.94549313031876e-05\\
31	0.000196108831114133\\
30	0.000254850111537734\\
29	0.000353403896314758\\
28	0.000905537551075214\\
27	0.000767971618603407\\
26	0.00100817152237784\\
25	0.00469692276045322\\
24	0.00569748763223424\\
23	0.0121009905344215\\
22	0.0177828963312645\\
21	0.0205426738321416\\
20	0.116310518473264\\
19	0.103515945172083\\
18	0.184524794384213\\
17	0.102187411184743\\
16	0.179767554862377\\
15	0.359175672080641\\
}--cycle;

\addplot [color=mycolor1, dashed]
  table[row sep=crcr]{%
15	0.40558565488253\\
16	0.354781561872057\\
17	0.354781561872057\\
18	0.358990728402916\\
19	0.358990728402916\\
20	0.358990728402916\\
21	0.358990728402916\\
22	0.358990728402916\\
23	0.329038218791742\\
24	0.286366864047647\\
25	0.185445729695679\\
26	0.137219289075272\\
27	0.109392158337461\\
28	0.0998242492531021\\
29	0.0514833847087556\\
30	0.0228876764811825\\
31	0.0138062882438082\\
32	0.0117291675754141\\
33	0.0117291675754141\\
34	0.0117291675754141\\
35	0.00541908608146481\\
36	0.00541908608146481\\
37	0.00541908608146481\\
38	0.00502078649224374\\
39	0.00502078649224374\\
40	0.00502078649224374\\
41	0.00502078649224374\\
42	0.00502078649224374\\
43	0.00502078649224374\\
44	0.00502078649224374\\
45	0.00502078649224374\\
46	0.00502078649224374\\
47	0.00502078649224374\\
48	0.00502078649224374\\
49	0.00502078649224374\\
50	0.00502078649224374\\
};

\addplot [color=mycolor2, dashdotdotted]
  table[row sep=crcr]{%
15	0.524618532533899\\
16	0.315226343101636\\
17	0.419202742330332\\
18	0.791969517777272\\
19	0.969709851243788\\
20	0.85782469451224\\
21	0.316841496681905\\
22	0.378353833373313\\
23	0.22355903678528\\
24	0.309689368502476\\
25	0.25753159961656\\
26	0.0940403369712225\\
27	0.104709986912292\\
28	0.275407259620217\\
29	0.0786587995315337\\
30	0.125597090377612\\
31	0.112373633194067\\
32	0.0371067398839537\\
33	0.0239137324695657\\
34	0.0336224843978418\\
35	0.0319308082845484\\
36	0.0330606885312183\\
37	0.023643437555823\\
38	0.0261932524708778\\
39	0.0291565857659495\\
40	0.0339765142561505\\
41	0.0253565121392083\\
42	0.0244578133244982\\
43	0.0297253361000807\\
44	0.0234425791837096\\
45	0.0332211720165483\\
46	0.0285188728753378\\
47	0.034184972802967\\
48	0.0177083526077893\\
49	0.0248149279947758\\
50	0.0220413323198014\\
};

\addplot [color=mycolor3]
  table[row sep=crcr]{%
15	0.926565233852604\\
16	0.298094147741752\\
17	0.250635341910755\\
18	0.36730874811769\\
19	0.349197931923361\\
20	0.355716760559929\\
21	0.148593959244595\\
22	0.0725849322733581\\
23	0.130770790688867\\
24	0.11134048769721\\
25	0.0721394617747438\\
26	0.0177770522862939\\
27	0.0210944543838576\\
28	0.0171863964525275\\
29	0.00810556044308027\\
30	0.00495899187474582\\
31	0.00176026899750175\\
32	0.000309019042235599\\
33	0.000197848455382528\\
34	0.00049236088053726\\
35	0.000564151726222273\\
36	0.000183797371455205\\
37	0.000298552650766384\\
38	0.000106818812793868\\
39	0.000155431638818605\\
40	7.06495996328842e-05\\
41	7.68806024272518e-05\\
42	4.76603978308899e-05\\
43	6.79557281402601e-05\\
44	6.61342314565992e-05\\
45	5.65156167175903e-05\\
46	5.25832628455921e-05\\
47	6.37386923232442e-05\\
48	4.90117637507318e-05\\
49	5.71917500256049e-05\\
50	2.8447944286247e-05\\
};

\addplot [color=TUMDunkelGrau, dashed]
  table[row sep=crcr]{%
15	0.984739566329921\\
16	0.984739566329921\\
17	0.984739566329921\\
18	0.984739566329921\\
19	0.984739566329921\\
20	0.984739566329921\\
21	0.984739566329921\\
22	0.984739566329921\\
23	0.984739566329921\\
24	0.984739566329921\\
25	0.984739566329921\\
26	0.984739566329921\\
27	0.984739566329921\\
28	0.984739566329921\\
29	0.984739566329921\\
30	0.984739566329921\\
31	0.984739566329921\\
32	0.984739566329921\\
33	0.984739566329921\\
34	0.984739566329921\\
35	0.984739566329921\\
36	0.984739566329921\\
37	0.984739566329921\\
38	0.984739566329921\\
39	0.984739566329921\\
40	0.984739566329921\\
41	0.984739566329921\\
42	0.984739566329921\\
43	0.984739566329921\\
44	0.984739566329921\\
45	0.984739566329921\\
46	0.984739566329921\\
47	0.984739566329921\\
48	0.984739566329921\\
49	0.984739566329921\\
50	0.984739566329921\\
};

\end{axis}
\end{tikzpicture}%
		\label{fig:map_conv_2dof_3dim_a}
	} \, 
	\setlength\fwidth{5cm}
	\setlength\fheight{5cm}
	\subfloat[Effect of neglecting noise in estimating $\alpha_\mathrm{EI}$]{
%
\definecolor{mycolor1}{rgb}{0.00000,0.39608,0.74118}%
\definecolor{mycolor2}{rgb}{0.89020,0.44706,0.13333}%
\definecolor{mycolor3}{rgb}{0.63529,0.67843,0.00000}%
\begin{tikzpicture}
	
	\begin{axis}[%
		width=\fwidth,
		height=\fheight,
		at={(0cm,0cm)},
		scale only axis,
		xmin=15,
		xmax=50,
		xlabel={Training set size $\ntrain$},
		ymode=log,
		ymin=1e-05,
		ymax=3,
		yminorticks=true,
		ylabel={Error $\varepsilon_{\mathrm{MAP}}$},
		]
		
		\addplot [color=mycolor1, dashed]
		table[row sep=crcr]{%
			15	0.40558565488253\\
			16	0.354781561872057\\
			17	0.354781561872057\\
			18	0.358990728402916\\
			19	0.358990728402916\\
			20	0.358990728402916\\
			21	0.358990728402916\\
			22	0.358990728402916\\
			23	0.329038218791742\\
			24	0.286366864047647\\
			25	0.185445729695679\\
			26	0.137219289075272\\
			27	0.109392158337461\\
			28	0.0998242492531021\\
			29	0.0514833847087556\\
			30	0.0228876764811825\\
			31	0.0138062882438082\\
			32	0.0117291675754141\\
			33	0.0117291675754141\\
			34	0.0117291675754141\\
			35	0.00541908608146481\\
			36	0.00541908608146481\\
			37	0.00541908608146481\\
			38	0.00502078649224374\\
			39	0.00502078649224374\\
			40	0.00502078649224374\\
			41	0.00502078649224374\\
			42	0.00502078649224374\\
			43	0.00502078649224374\\
			44	0.00502078649224374\\
			45	0.00502078649224374\\
			46	0.00502078649224374\\
			47	0.00502078649224374\\
			48	0.00502078649224374\\
			49	0.00502078649224374\\
			50	0.00502078649224374\\
		};

		\addplot [color=mycolor3]
		table[row sep=crcr]{%
			15	0.926565233852604\\
			16	0.298094147741752\\
			17	0.250635341910755\\
			18	0.36730874811769\\
			19	0.349197931923361\\
			20	0.355716760559929\\
			21	0.148593959244595\\
			22	0.0725849322733581\\
			23	0.130770790688867\\
			24	0.11134048769721\\
			25	0.0721394617747438\\
			26	0.0177770522862939\\
			27	0.0210944543838576\\
			28	0.0171863964525275\\
			29	0.00810556044308027\\
			30	0.00495899187474582\\
			31	0.00176026899750175\\
			32	0.000309019042235599\\
			33	0.000197848455382528\\
			34	0.00049236088053726\\
			35	0.000564151726222273\\
			36	0.000183797371455205\\
			37	0.000298552650766384\\
			38	0.000106818812793868\\
			39	0.000155431638818605\\
			40	7.06495996328842e-05\\
			41	7.68806024272518e-05\\
			42	4.76603978308899e-05\\
			43	6.79557281402601e-05\\
			44	6.61342314565992e-05\\
			45	5.65156167175903e-05\\
			46	5.25832628455921e-05\\
			47	6.37386923232442e-05\\
			48	4.90117637507318e-05\\
			49	5.71917500256049e-05\\
			50	2.8447944286247e-05\\
		};

		\addplot [color=TUMBlau, dotted]
		table[row sep=crcr]{%
				15	1.02721672299525\\
				16	0.960111673130446\\
				17	0.90660461774032\\
				18	0.863338755302896\\
				19	0.821116715543404\\
				20	0.80277031796359\\
				21	0.765999341887874\\
				22	0.587137099539448\\
				23	0.602508301255402\\
				24	0.53808639551431\\
				25	0.530432538927317\\
				26	0.437765839051709\\
				27	0.379330031305426\\
				28	0.233156157311627\\
				29	0.197918662059659\\
				30	0.155915024287078\\
				31	0.12332071457681\\
				32	0.106378934106799\\
				33	0.0899428929676963\\
				34	0.0601160632080085\\
				35	0.0601160632080085\\
				36	0.0595708031613548\\
				37	0.0595708031613548\\
				38	0.0400645583272186\\
				39	0.0400645583272186\\
				40	0.0400645583272186\\
				41	0.0335621358782942\\
				42	0.0223792857801482\\
				43	0.016715104690345\\
				44	0.016715104690345\\
				45	0.016715104690345\\
				46	0.0137896987818444\\
				47	0.0137896987818444\\
				48	0.0137896987818444\\
				49	0.0137896987818444\\
				50	0.00816486602401252\\
			};
		
		\addplot [color=TUMGruen, dotted]
		table[row sep=crcr]{%
				15	0.68398915048354\\
				16	0.686571669973931\\
				17	0.837645408377215\\
				18	0.76708774783803\\
				19	0.714522320180489\\
				20	0.538011553009521\\
				21	0.599806379095421\\
				22	0.575995825561401\\
				23	0.355519532338928\\
				24	0.377133048808176\\
				25	0.309595203727585\\
				26	0.127363780607433\\
				27	0.250593434473032\\
				28	0.0293048356201713\\
				29	0.00740301199586695\\
				30	0.0247217311482205\\
				31	0.00616230767171181\\
				32	0.00766976720552678\\
				33	0.00265589136252313\\
				34	0.00155203082257924\\
				35	0.00172444179129943\\
				36	0.00134774252388921\\
				37	0.00109560611589045\\
				38	0.00102659910694824\\
				39	0.000484607791135567\\
				40	0.000268650501449049\\
				41	0.000340651929834426\\
				42	0.000117771314677419\\
				43	0.000205404160875463\\
				44	0.000116082166407849\\
				45	0.000414124958516011\\
				46	0.000349306308516816\\
				47	0.00013696671976778\\
				48	0.000166093937605059\\
				49	4.58287768599597e-05\\
				50	8.52320636953042e-05\\
			};

	\end{axis}
\end{tikzpicture}%
		\label{fig:map_conv_2dof_3dim_b}
	}
	\caption{
		Convergence of the MAP estimate for the three parameter updating problem of the two-degree-of-freedom system.
		We depict the relative error $\varepsilon_{\mathrm{MAP}}$ in the standard normal space for the simple reward estimator (\linedashedblue), the global reward estimator (\linefullgreen), and a fixed experimental design (\linedashdottedorange) based estimator.
		The relative distance of the prior mean in the standard normal space is also given (\linedasheddarkgrey).
		In Fig.~\ref{fig:map_conv_2dof_3dim_b}, the corresponding simple (\linedottedgreen) and global (\linedottedblue) reward estimator without the inclusion of the noise in estimating $\alpha_\mathrm{EI}$ are compared to their counterparts from Fig.~\ref{fig:map_conv_2dof_3dim_a}.
	}
	\label{fig:map_conv_2dof_3dim}
\end{figure}

In Fig.~\ref{fig:map_conv_2dof_3dim_a}, we depict the median values of $\varepsilon_{\mathrm{MAP}}$ as well as min-to-max range over increasing number of experimental design samples.
We observe that the global reward MAP estimator (\linefullgreen) consistently converges towards the MAP value from $\ntrain \approx 20$ on.
From around $\ntrain \approx 35$, the error of the MAP estimate is in the order of $10^{-4}$ indicating a very good agreement to the MAP value obtained using the original model.
It outperforms the fixed design RPCE based estimator (\linedashdottedorange), highlighting the improvement that is gained through the adaptive sampling.
The simple reward estimator (\linedashedblue) converges to the reference MAP value as well, however, it plateaus at a higher level in comparison the global reward MAP estimator. 

In Fig.~\ref{fig:map_conv_2dof_3dim_b}, we depict both active learning based MAP estimators, namely the global and simple reward estimator, for the case, where the randomness in the surrogate model error $\bm{\varepsilon}_{\mathcal{S}}$ is neglected in the computation of the expected improvement. 
We depict the corresponding results from the previous study in Fig.~\ref{fig:map_conv_2dof_3dim_a} for comparison purposes. 
It can be observed that the estimates also converge to the reference value, however, their accuracy is slightly worse. 
Since the inclusion of the error samples comes at almost no cost, we conclude that it is worthwhile to include the surrogate model error $\bm{\varepsilon}_{\mathcal{S}}$ in the random process surrogate model.

In total, it can be seen that the proposed method is able to effectively guide the sampling process for efficient MAP estimation. 
The resulting estimates are more accurate than the ones based on a fixed experimental design, and thus require less model evaluations for a fixed target estimation error.
Subsequently, we apply the methodology to the parameter updating problem of a more complex finite element model.
\subsection{Finite element model: cross-laminated timber plate}
\label{subsec:fem}
\begin{figure}[!hbt]
	\centering
	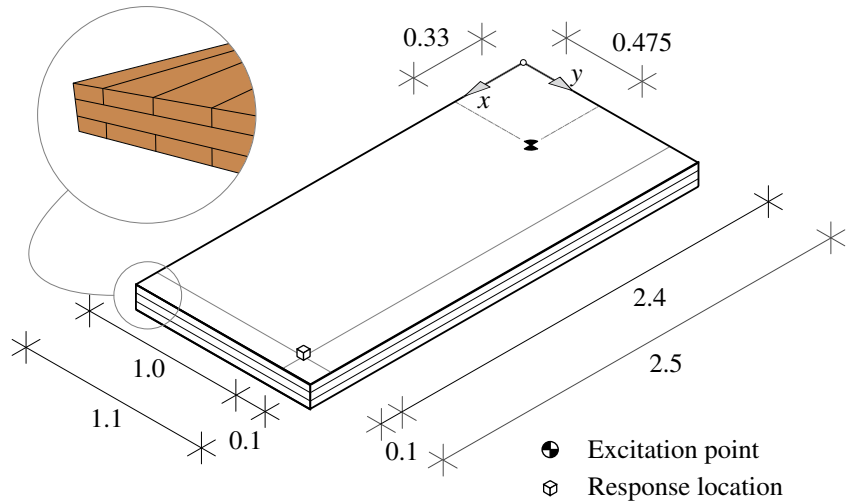
	\caption{Illustration of cross-laminated timber specimen with excitation and response location.}
	\label{fig:clt_setup}
\end{figure}

In the following example, we consider the finite element model parameter updating for a cross-laminated timber plate. 
The plate and the corresponding measurement setup are illustrated in Fig.~\ref{fig:clt_setup}.

The considered system is a plate of dimensions $l \times b \times h = 2.5 \, \mathrm{m} \times 1.1 \, \mathrm{m} \times 0.081 \, \mathrm{m}$ that consists of three timber plank layers that are laminated on top of each other in a cross-wise fashion.
Forced input excitation measurement data is available for a plate suspended freely from the ceiling (see \cite{Mecking.2017}). 
A finite element model of the structure is available in the commercial software package \ansys. 
The model uses solid elements based on three-dimensional linear elasticity for orthotropic materials. 
Details about the model are available in \cite{Schneider.2022}. 

We choose to perform the updating based on the information from the sensor with location $(x_\mathrm{sensor}, y_\mathrm{sensor}) = ( 2.4 \, \mathrm{m}, \, 1.0 \, \mathrm{m} )$.
The frequency transfer function is estimated through the H1-estimator \cite{Pintelon.2012} and sub-sampled within the intervals $[25 \, \mathrm{Hz}, 35 \, \mathrm{Hz}]$ and $[65 \, \mathrm{Hz}, 75 \, \mathrm{Hz}]$ with a frequency step size of $\Delta f = 1 \, \mathrm{Hz}$.
Within both intervals, resonance peaks can be observed.
This sub-selection is made to reduce the overall number of observation points.

We model the error model covariance matrices $\bm{\Sigma}_{\mathbf{w}_{\varepsilon}\mathbf{w}_{\varepsilon}}$ and $\bm{\Sigma}_{\bm{\varphi}_{\varepsilon}\bm{\varphi}_{\varepsilon}}$ under the assumption of a parametric correlation function over the frequency space.
To this end, we utilize a homoscedastic model with variances $\sigma_w^2$ and $\sigma_\varphi^2$ that expresses the errors as random fields over the frequency domain and defines their correlation coefficient function as
\begin{equation}
	\rho \left( f_1, f_2; l_{\mathrm{co},f}, r \right) = r \exp \left(l_{\mathrm{co},f}^{-1} \abs{ f_1 - f_2} \right) + ( 1 - r ) 
\end{equation}
where $f_1$ and $f_2$ denote two frequency points, $l_{\mathrm{co},f}$ is the correlation length of the exponential correlation model and $r$ models the split between the frequency dependent correlation and a baseline model correlation that is constant for all data points. 
$\rho$ is equally applied for both $\bm{\Sigma}_{\mathbf{w}_{\varepsilon}\mathbf{w}_{\varepsilon}}$ and $\bm{\Sigma}_{\bm{\varphi}_{\varepsilon}\bm{\varphi}_{\varepsilon}}$. 
A detailed discussion of the error model can be found in \cite{Schneider.2022}.
In general, the error model hyperparameters $\sigma_w^2$, $\sigma_\varphi^2$, $l_{\mathrm{co},f}$ and $r$ are unknown and need to be chosen based on assumptions or determined from the measurement data.
Here, we use the values that were identified in \cite{Schneider.2022}, which are $r = 0.8$, $l_{\mathrm{co},f} = 5 \, \mathrm{Hz}$ and $\sigma_w = \sigma_\varphi = 0.33$.
The system parameters' prior distributions are assumed as summarized in Table~\ref{tab:clt_par_assumpt}.
\begin{table}[!htb]
	\footnotesize
	\centering
	\caption{Prior distribution assumptions for the cross-laminated timber plate model. The parameters are assumed to be independent.}
	\begin{tabular}{lcccc}
		\toprule
		Parameter & & Distribution & Mean value $\mu_{(\cdot)}$ & Coefficient of variation $\delta_{(\cdot)}$ \\
		\midrule
		Young's modulus & $E_x$ & Lognormal & $1.1 \cdot 10^{10} \, \mathrm{\frac{N}{m^2}}$ & $0.1$ \\[1mm]
		Young's modulus & $E_y$ & Lognormal & $0.85 \cdot 3.667 \cdot 10^8 \, \mathrm{\frac{N}{m^2}}$ & $0.1$ \\[1mm]
		Shear modulus   & $G_{xy}$ & Lognormal & $0.7 \cdot 6.9 \cdot 10^8 \, \mathrm{\frac{N}{m^2}}$ & $0.1$ \\[1mm]
		Damping ratio   & $\zeta$ & Lognormal & $2 \cdot 10^{-2}$ & $0.3$ \\[1mm]
		\bottomrule
	\end{tabular}
	\label{tab:clt_par_assumpt}
\end{table}

The same analyses as for the two degree of freedom model are performed.
The expected improvement is estimated using $n_{\alpha} = 50$ samples.
The optimal expected improvement is found through particle swarm optimization \cite{Kennedy.1995} using the built-in algorithm in \matlab. 
The number of particles is chosen as $n_{\mathrm{sw}} = 40$. 
	Smaller numbers of particles have been found to yield reasonable results as well.
	A reference value for the MAP estimate is computed using particle swarm optimization.
	Here, we generate $n_{\mathrm{post}} = 10^3$ samples from the posterior distribution using aBUS-SuS.
	The posterior statistics and the median MAP estimates at $\ntrain = 50$ are summarized in Tab.~\ref{tab:clt_ref_estimates_4dim}.
	
	In Fig.~\ref{fig:map_conv_clt_4dim}, we depict the median values of $\varepsilon_{\mathrm{MAP}}$ as well as min-to-max range over increasing number of experimental design samples.
	Again, the active learning based MAP estimators exhibit better convergence performance in comparison to the fixed design based estimator. 
	Towards reaching the budget of $n_{\mathrm{budget}} = 35$, the error $\varepsilon_{\mathrm{MAP}}$ lies in the order of $10^{-4}$ in median for the global reward estimator.
	The fixed design based strategy leads to an estimator with $\varepsilon_{\mathrm{MAP}}$ in the order of $10^{-2}$ in median.
	This highlights the improvement gained through the sequential experimental design strategy in estimating the MAP also for this more complex updating problem. 
	\begin{figure}[!hbt]
		\centering
		\setlength\fwidth{8cm}
		\setlength\fheight{5cm}
%
\definecolor{mycolor1}{rgb}{0.00000,0.39608,0.74118}%
\definecolor{mycolor2}{rgb}{0.89020,0.44706,0.13333}%
\definecolor{mycolor3}{rgb}{0.63529,0.67843,0.00000}%
\begin{tikzpicture}

\begin{axis}[%
width=\fwidth,
height=\fheight,
at={(0cm,0cm)},
scale only axis,
xmin=15,
xmax=35,
xlabel={Iteration Number},
ymode=log,
ymin=1e-05,
ymax=10,
yminorticks=true,
ylabel={MAP Estimate Norm Error},
]

\addplot[area legend, draw=none, fill=mycolor1, fill opacity=0.3, forget plot]
table[row sep=crcr] {%
x	y\\
15	0.742751061968845\\
16	0.707581089291495\\
17	0.375734372801662\\
18	0.345142924683492\\
19	0.345142924683492\\
20	0.345142924683492\\
21	0.191812713431458\\
22	0.191812713431458\\
23	0.191812713431458\\
24	0.0642361185095277\\
25	0.0642361185095277\\
26	0.0642361185095277\\
27	0.0396218810917308\\
28	0.0396218810917308\\
29	0.0396218810917308\\
30	0.0396218810917308\\
31	0.0396218810917308\\
32	0.0396218810917308\\
33	0.00830160025822551\\
34	0.00679184761121797\\
35	0.00679184761121797\\
35	0.00102485804532303\\
34	0.00102485804532303\\
33	0.00102485804532303\\
32	0.00102485804532303\\
31	0.00102485804532303\\
30	0.00102485804532303\\
29	0.00102485804532303\\
28	0.00102485804532303\\
27	0.00102485804532303\\
26	0.00102485804532303\\
25	0.00102485804532303\\
24	0.002563996666461\\
23	0.0129638908180611\\
22	0.0129638908180611\\
21	0.0129638908180611\\
20	0.0463232998073212\\
19	0.0463232998073212\\
18	0.0463232998073212\\
17	0.186479003713043\\
16	0.186479003713043\\
15	0.186479003713043\\
}--cycle;

\addplot[area legend, draw=none, fill=mycolor2, fill opacity=0.3, forget plot]
table[row sep=crcr] {%
x	y\\
15	2.9143568823152\\
16	1.3038259572542\\
17	0.349129304666529\\
18	0.466369006450064\\
19	0.246372646161462\\
20	0.176928214081395\\
21	0.1121921030922\\
22	0.074166227007271\\
23	0.144139481262646\\
24	0.0619695922817596\\
25	0.0890177591472396\\
26	0.076992770164834\\
27	0.0339618134322378\\
28	0.0480225975889345\\
29	0.0432166917912558\\
30	0.10001195376227\\
31	0.033582929711064\\
32	0.031557762058744\\
33	0.0293525641802649\\
34	0.0661223145450477\\
35	0.0234671798071118\\
35	0.00559066194469367\\
34	0.00560118331660943\\
33	0.0147204248399897\\
32	0.0142021589248316\\
31	0.0120884599112574\\
30	0.0098965428050328\\
29	0.00847477637827207\\
28	0.0143394199050496\\
27	0.00997001322312787\\
26	0.0111338151410497\\
25	0.0269613738969533\\
24	0.016555940198431\\
23	0.042541853140981\\
22	0.0560485762583438\\
21	0.0222950222705633\\
20	0.0300196079701187\\
19	0.0445931195894209\\
18	0.0521847294752598\\
17	0.0975364909110539\\
16	0.0702581423673351\\
15	0.159985300376837\\
}--cycle;

\addplot[area legend, draw=none, fill=mycolor3, fill opacity=0.3, forget plot]
table[row sep=crcr] {%
x	y\\
15	1.00941766132819\\
16	0.198727069397675\\
17	0.457807885912493\\
18	0.510407800152231\\
19	1.0149414848947\\
20	0.34879560671061\\
21	0.356063538636702\\
22	0.0834069631262204\\
23	0.0773884801574803\\
24	0.00684148499672698\\
25	0.0216487262077377\\
26	0.172312625524545\\
27	0.00441991032646869\\
28	0.00834859290820971\\
29	0.00516011464745645\\
30	0.000698418430009339\\
31	0.00136468847416446\\
32	0.00112951195686717\\
33	0.00182109937021431\\
34	0.000570890835307519\\
35	0.000689790052314369\\
35	0.000192392479250039\\
34	0.000225055984659071\\
33	0.000109615061312757\\
32	0.000176372796982651\\
31	0.000263980341765507\\
30	0.000269362749968387\\
29	0.000321315798968777\\
28	0.000201417362007989\\
27	7.05772704275875e-05\\
26	0.000522417179566433\\
25	0.00072056252632144\\
24	0.00114258150065845\\
23	0.00106335190514951\\
22	0.00364713452570893\\
21	0.00539867120454747\\
20	0.0123238230284572\\
19	0.0592120539497541\\
18	0.0672365738428424\\
17	0.091075756743722\\
16	0.0710779254896616\\
15	0.122468293708454\\
}--cycle;

\addplot [color=mycolor1, dashed]
  table[row sep=crcr]{%
15	0.426673080621642\\
16	0.345142924683492\\
17	0.280770098341782\\
18	0.280770098341782\\
19	0.196214435927684\\
20	0.0841264042007975\\
21	0.0463232998073212\\
22	0.0346606311777129\\
23	0.0223919607699516\\
24	0.0161286109442365\\
25	0.00725162916368548\\
26	0.0112005155386437\\
27	0.00408020610116793\\
28	0.00344896931725891\\
29	0.00344896931725891\\
30	0.00344896931725891\\
31	0.00344896931725891\\
32	0.00344896931725891\\
33	0.00344896931725891\\
34	0.00344896931725891\\
35	0.00344896931725891\\
};

\addplot [color=mycolor2, dashdotdotted]
  table[row sep=crcr]{%
15	0.291354746714696\\
16	0.17059125558682\\
17	0.22050160110018\\
18	0.123746550208373\\
19	0.103843933941835\\
20	0.0742985821403436\\
21	0.081297007626341\\
22	0.0620484047233144\\
23	0.0685653966805432\\
24	0.0529732340992578\\
25	0.041925081849715\\
26	0.029978086233362\\
27	0.0185500768895844\\
28	0.0235974037480076\\
29	0.0266934023705336\\
30	0.0350365230200939\\
31	0.0286683709428944\\
32	0.0157730417641011\\
33	0.0221103648172459\\
34	0.0125681366764966\\
35	0.00732326742553332\\
};

\addplot [color=mycolor3]
  table[row sep=crcr]{%
15	0.270797996398182\\
16	0.139586815550863\\
17	0.124876369313823\\
18	0.147589278010598\\
19	0.0748193404281477\\
20	0.0245406924173749\\
21	0.0169833000808026\\
22	0.00890040098711018\\
23	0.00197401828337477\\
24	0.00246358447254983\\
25	0.000967579563594271\\
26	0.00118951718906682\\
27	0.000397233454091313\\
28	0.000579574102987713\\
29	0.000519713354246074\\
30	0.000373863690593583\\
31	0.000473624285915764\\
32	0.000343273979663946\\
33	0.000349771568861327\\
34	0.00047071232740862\\
35	0.0002777875420867\\
};

\addplot [color=TUMDunkelGrau, dashed]
  table[row sep=crcr]{%
15	0.936895049782939\\
16	0.936895049782939\\
17	0.936895049782939\\
18	0.936895049782939\\
19	0.936895049782939\\
20	0.936895049782939\\
21	0.936895049782939\\
22	0.936895049782939\\
23	0.936895049782939\\
24	0.936895049782939\\
25	0.936895049782939\\
26	0.936895049782939\\
27	0.936895049782939\\
28	0.936895049782939\\
29	0.936895049782939\\
30	0.936895049782939\\
31	0.936895049782939\\
32	0.936895049782939\\
33	0.936895049782939\\
34	0.936895049782939\\
35	0.936895049782939\\
};

\end{axis}
\end{tikzpicture}%
		\caption{
			Convergence of the MAP estimate for the four parameter updating problem of the cross-laminated timber plate model using solid elements.
			We depict the relative error $\varepsilon_{\mathrm{MAP}}$ in the standard normal space for the simple reward estimator (\linedashedblue), the global reward estimator (\linefullgreen), and a fixed experimental design (\linedashdottedorange) based estimator.
			The relative distance of the prior mean in the standard normal space is also given (\linedasheddarkgrey).
		}
		\label{fig:map_conv_clt_4dim}
	\end{figure}
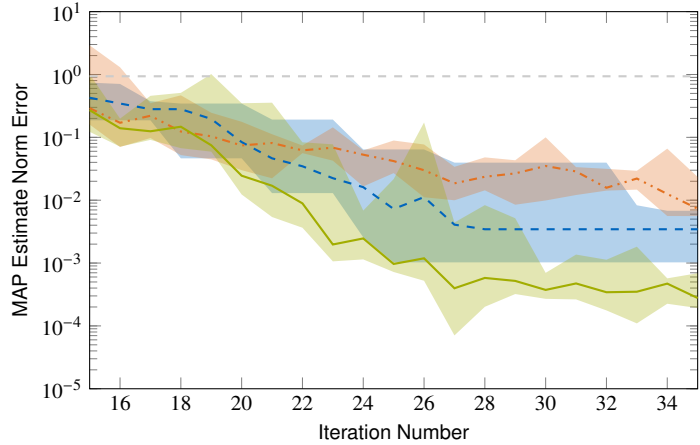
	\begin{table}[!htb]
		\footnotesize
		\centering
		\caption{Reference parameter estimates for the finite element model of the CLT plate.}
		\begin{tabular}{lcccccccc}
			\toprule
			Parameter & & Units & \multicolumn{2}{c}{Posterior} & \multicolumn{4}{c}{MAP estimate} \\\cmidrule(lr){4-5}\cmidrule(lr){6-9}
			& & & mean & c.o.v. & reference & global reward & simple reward & fixed design \\
			\midrule
			Young's modulus & $E_x$ & $10^{10} \, \mathrm{\frac{N}{m^2}}$ & $1.1843$ & $0.023$ & $1.1875$ & $1.1875$ & $1.1874$ & $1.1875$ \\[1mm]
			Young's modulus & $E_y$ & $10^8 \, \mathrm{\frac{N}{m^2}}$ & $3.1339$ & $0.089$ & $3.0375$ & $3.0375$ & $3.0368$ & $3.0372$ \\[1mm]
			Shear modulus & $G_{xy}$ & $10^8 \, \mathrm{\frac{N}{m^2}}$ & $4.5633$ & $0.028$ & $4.5923$ & $4.5924$ & $4.5925$ & $4.5901$ \\[1mm]
			Damping ratio & $\zeta$ & $10^{-2} \, -$& $3.4512$ & $0.126$ & $3.2888$ & $3.2887$ & $3.2907$ & $3.2993$ \\[1mm]
			\bottomrule
		\end{tabular}
		\label{tab:clt_ref_estimates_4dim}
	\end{table}

\section{Discussion}
\label{subsec:disc}
While the presented results indicate that the proposed active learning approach for MAP estimation in conjunction with RPCE models can be applied successfully, there remain a number of open questions that could be addressed in future research.
Currently, the approach requires substituting each scalar model response in spatial and frequency domain that enters the likelihood by a separate surrogate model. 
For a large number of observations this causes a substantial computational effort due to the iterative training for each model as well as the individual sampling of the coefficients.
A possible remedy for this is to formulate an RPCE model that allows one to surrogate the frequency response over the parameter space and the frequency domain simultaneously. 
Another approach is to apply a dimensionality reduction step to the discretized model response. 
Principal component analysis (PCA) has been applied to the frequency response analysis in conjunction with RPCE in \cite{Manfredi.2021}.
It has been found, however, that the application of PCA to the investigated lightly damped systems in this paper is not straightforward since a large number of principal components is necessary and significant errors might be introduced in the procedure. 

Furthermore, the current proposal requires training the Bayesian RPCE model in each active learning iteration step anew, starting from a full set of basis vectors.
Ideally, one would start from the set of basis vectors identified in the previous step and add or delete basis functions as necessary.
While this approach is formulated and well-investigated for linear Bayesian models (cf. \cite{tipping2003fast}, \cite{Babacan.2010}), an analogous variant for RPCE is not yet available.
Without the ability to reactivate basis functions to the active set of basis functions, starting the sequential learning procedure from a truncated set of basis functions might be too restrictive.

The proposed procedure utilizes the well known expected improvement acquisition function.
The design and investigation of alternative acquisition functions is an active research question and different proposals could also improve the approach presented in this work.
It should be noted, however, that due to the iterative training procedure required for Bayesian RPCE, non-myopic acquisition functions, such as the knowledge gradient acquisition function, might be overly computationally expensive, since they require re-training for each sample in the evaluation of the expected value of the acquisition function.
Another interesting extension is motivated by the observation that the acquisition function becomes flat and equal to zero for large parts of the input domain. 
This is due to the approximation of the expected value through a finite number of samples and the limitation of numerical computations to machine precision.
Since a flat objective will exhibit vanishing gradients, this issue complicates the optimization of the expected improvement acquisition function.
Similar issues have been reported and discussed in \cite{NEURIPS2023_419f72cb}.
Therein, the authors propose an alternative formulation of the expected improvement acquisition function, namely the LogEI acquisition, that remedies the identified issue.
The proposed modification could potentially also be applied to the methodology outlined in this paper.

Finally, in all examples, knowledge about the error model hyperparameters has been assumed. 
In common Bayesian updating settings, one often lacks this knowledge. 
For such cases, robust estimation of the hyperparameters would be necessary. 
As an extension to the presented methodology, we propose to sequentially update the model and hyperparameters within each active learning step.
After having enriched the experimental design, an optimal set of hyperparameters can be obtained through solving a separate optimization problem.
The updated set of hyperparameters can then be used in the subsequent iteration step for evaluating the acquisition function.
The applicability of this approach and the influence of an initially possibly poor estimate of the hyperparameters on the procedure could be investigated in future research. 
\section{Conclusion}
\label{subsec:concl}
This work presents a novel maximum a-posteriori estimation technique for Bayesian inverse problems that utilizes frequency response function data to update parameters in linear, time-invariant mechanical models. 
The proposed method integrates a sequential experimental design strategy based on Bayesian optimization into the maximum a-posteriori estimation procedure through utilizing rational polynomial chaos expansion surrogate models.
Therein, the surrogate models alleviate the computational burden of solving the optimization problem and provide a highly accurate representation for the considered parametric frequency response function models.
To train the surrogate model, we extend a previously introduced sparse Bayesian regression approach to determine the surrogate model parameters through introducing an alternative assumption on the distribution of the surrogate modeling error that enables a tractable Laplace approximation of the denominator coefficients.
The training set is sequentially enriched through maximizing the expected improvement acquisition function.
Since the required expectation operation cannot be carried out in closed-form, we resort to a sampling-based approximation based on samples from the posterior distribution of the surrogate model coefficients.
The method is successfully applied to identify the MAP estimates for the parameters of a two-degree-of-freedom system as well as a finite element model of a cross laminated timber plate.
In the former case, synthetic data is generated, while for the latter, actual measurement data is used. 
The active learning procedure is carried out until a prescribed budget of model evaluations is reached.
We compare the proposed methodology to a fixed design based RPCE solution and evaluate the relative difference of the MAP estimates to a reference solution found by global optimization.
For both models, the MAP estimates identified through the active learning based method show improved accuracy compared to a fixed design approach.
This allows to significantly reduce the overall number of model evaluations in the estimation procedure.
This reduction is due to the adaptive sequential experimental design strategy, which places samples in the relevant region of the input space and thus leads to better accuracy of the surrogate models in the region of high posterior probability density.

\section*{Acknowledgement}
The first author thanks Xujia Zhu for his helpful comments and insightful discussions during the first authors stay at the Chair of Risk, Safety and Uncertainty Quantification, ETH Zurich.

\appendix

\section{Partial Derivatives and Hessian Matrix with Respect to the Denominator Coefficients $\mathbf{q}$}
\label{app:grad_q}
In this section we derive the partial derivatives and Hessian matrix of the objective function as defined in the optimization problem in Eq.~\eqref{eq:q_MAP2}.
To this end, we define
\begin{equation}
	\obj_q (\mathbf{q}) = \ln \det \mathbf{Q} \mathbf{Q}^{\mathrm{H}} - \mathbf{q}^{\mathrm{H}} \left( \bm{\Upsilon}^{\mathrm{H}} \Tilde{\mathbf{C}}^{-1} \bm{\Upsilon} + \bm{\Lambda}_{\mathbf{q}\mathbf{q}} \right) \mathbf{q} \, .
	\label{eq:obj_q}
\end{equation}
Our goal is to compute the partial derivatives with respect to the conjugate denominator coefficients, i.e., $\pdv{\obj_q (\mathbf{q})}{\conj{\mathbf{q}}}$.
We consider the first part in Eq.~\eqref{eq:obj_q} and derive the partial derivative with respect to the $i$-th conjugate denominator coefficient $\conj{q}_i$.
To this end, we write $\ln \det \mathbf{Q} \mathbf{Q}^{\mathrm{H}} = \ln \det \mathbf{Q} + \ln \det \mathbf{Q}^{\mathrm{H}}$ and note that $\pdv{\conj{q}_i} \ln \det \mathbf{Q} = \mathbf{0}$.
Then 
\begin{equation}
	\pdv{\conj{q}_i} \ln \det \mathbf{Q} \mathbf{Q}^{\mathrm{H}} = \pdv{\conj{q}_i} \ln \det \mathbf{Q}^{\mathrm{H}} = \tr \left( \mathbf{Q}^{- \mathrm{H}} \pdv{\mathbf{Q}^{\mathrm{H}}}{\conj{q}_i} \right) = \tr \left( \mathbf{Q}^{- \mathrm{H}} \diag \left( \left( \bm{\Psi}_{q} \right)_{:,i} \right) \right) \, ,
\end{equation}
where $(\cdot)_{:,i}$ denotes the $i$-th column of the matrix $(\cdot)$.
The full vector of partial derivatives can then be conveniently computed through a matrix-vector product as
\begin{equation}
	\pdv{\conj{\mathbf{q}}} \ln \det \mathbf{Q} \mathbf{Q}^{\mathrm{H}} = \bm{\Psi}_q^{\mathrm{T}} \left( \bm{\Psi}_q \conj{\mathbf{q}} \right)^{\circ-1}
\end{equation}
Furthermore, it holds
\begin{equation}
	\pdv{\conj{\mathbf{q}}} \left[ \mathbf{q}^{\mathrm{H}} \left( \bm{\Upsilon}^{\mathrm{H}} \Tilde{\mathbf{C}}^{-1} \bm{\Upsilon} + \bm{\Lambda}_{\mathbf{q}\mathbf{q}} \right) \mathbf{q} \right] = \left( \bm{\Upsilon}^{\mathrm{H}} \Tilde{\mathbf{C}}^{-1} \bm{\Upsilon} + \bm{\Lambda}_{\mathbf{q}\mathbf{q}} \right) \mathbf{q} \, .
\end{equation}
Thus,
\begin{equation}
	\pdv{\conj{\mathbf{q}}} \left[ 
	\ln \det \mathbf{Q} \mathbf{Q}^{\mathrm{H}} - \mathbf{q}^{\mathrm{H}} \left( \bm{\Upsilon}^{\mathrm{H}} \Tilde{\mathbf{C}}^{-1} \bm{\Upsilon} + \bm{\Lambda}_{\mathbf{q}\mathbf{q}} \right) \mathbf{q} \right] = \bm{\Psi}_q^{\mathrm{T}} \left( \bm{\Psi}_q \conj{\mathbf{q}} \right)^{\circ-1} - \left( \bm{\Upsilon}^{\mathrm{H}} \Tilde{\mathbf{C}}^{-1} \bm{\Upsilon} + \bm{\Lambda}_{\mathbf{q}\mathbf{q}} \right) \mathbf{q} \, .
	\label{eq:pdiv_q_1_app}
\end{equation}
Next, based on the result in Eq.~\eqref{eq:pdiv_q_1_app}, the Hessian matrix can be expressed as
\begin{equation}
	\mathbf{H}_{\mathbf{q}\mathbf{q}} = \pdv{\mathbf{q}} \left( \pdv{\conj{\mathbf{q}}} \left[ 
	\ln \det \mathbf{Q} \mathbf{Q}^{\mathrm{H}} - \mathbf{q}^{\mathrm{H}} \left( \bm{\Upsilon}^{\mathrm{H}} \mathbf{C}^{-1} \bm{\Upsilon} + \bm{\Lambda}_{\mathbf{q}\mathbf{q}} \right) \mathbf{q} \right] \right)^\mathrm{T} = - \left( \bm{\Upsilon}^{\mathrm{H}} \mathbf{C}^{-1} \bm{\Upsilon} + \bm{\Lambda}_{\mathbf{q}\mathbf{q}} \right) \, . \label{eq:hessian_app}%
\end{equation}
Here, we have made use of the fact that the first summand in Eq.~\eqref{eq:pdiv_q_1_app} does not depend on $\mathbf{q}$, but only on $\conj{\mathbf{q}}$.
\section{Partial derivatives with respect to the hyperparameters $\bm{\alpha}_p$}
\label{app:grad_alpha_p}
In this section we derive the partial derivatives of the objective function as defined in the optimization problem in Eq.~\eqref{eq:hyper_MAP2} with respect to the numerator coefficient precisions.
To this end, we define
\begin{equation}
	\obj_{\bm{\alpha}_p} (\bm{\alpha}_p) = 
	\ln \det \bm{\Sigma}_{\mathbf{p}\mathbf{p}| \mathbf{m}} 
	+ \ln \det \bm{\Lambda}_{\mathbf{p}\mathbf{p}} 
	- \beta_{\mathcal{S}} \mathbf{m}^{\mathrm{H}} \conj{\mathbf{Q}^\ast} \left( \mathbf{Q}^\ast \mathbf{m} - \bm{\Psi}_p \Tilde{\bm{\mu}}_{\mathbf{p}|\mathbf{m}} \right)  
	- \ln \det \left( - \mathbf{H}_{\mathbf{q}\mathbf{q}} \right)
	\, .
	\label{eq:obj_q_alpha_p_app}
\end{equation}
We observe that the objective function in Eq.~\eqref{eq:obj_q_alpha_p_app} possesses a similar structure compared to the one presented in \cite{Schneider.2023} for the standard residual formulation.
While in \cite{Schneider.2023}, the denominator coefficients enter in an augmented design matrix, here, they can be understood as augmenting the data vector.
However, the dependency of the expressions on the precision parameters $\bm{\alpha}_p$ is the same, so the results can be transferred.
In analogy to the expressions in Appendix D in \cite{Schneider.2023}, we obtain
\begin{equation}
	\pdv{\alpha_{p,i}} \left[
	\ln \det \bm{\Sigma}_{\mathbf{p}\mathbf{p}| \mathbf{m}} 
	+ \ln \det \bm{\Lambda}_{\mathbf{p}\mathbf{p}} 
	- \beta_{\mathcal{S}} \mathbf{m}^{\mathrm{H}} \conj{\mathbf{Q}^\ast} \left( \mathbf{Q}^\ast \mathbf{m} - \bm{\Psi}_p \bm{\mu}_{\mathbf{p}|\mathbf{m}, \mathbf{q}^\ast} \right)  
	\right] 
	= 
	- \left[ \bm{\Sigma}_{\mathbf{p}\mathbf{p}| \mathbf{m}, \mathbf{q}^\ast} \right]_{ii} + \alpha_{p,i}^{-1} - \abs{ \left[\mu_{\mathbf{p}|\mathbf{m}, \mathbf{q}^\ast} \right]_i }^2
\end{equation}
Additionally, the derivative with respect to the last term in Eq.~\eqref{eq:obj_q_alpha_p_app} is required.
We find
\begin{equation}
	\pdv{\ln \det \left( - \mathbf{H}_{\mathbf{q} \mathbf{q}} \right)}{\alpha_{p,i}} = \tr \left( \left( - \mathbf{H}_{\mathbf{q}\mathbf{q}} \right)^{-1} \pdv{\left( - \mathbf{H}_{\mathbf{q}\mathbf{q}} \right)}{\alpha_{p,i}} \right) \, .
\end{equation}
It holds
\begin{equation}
	\pdv{\left( - \mathbf{H}_{\mathbf{q}\mathbf{q}} \right)}{\alpha_{p,i}} = \bm{\Upsilon}^{\mathrm{H}} \pdv{\mathbf{C}^{-1}}{\alpha_{p,i}} \bm{\Upsilon} \, ,
\end{equation}
and through application of the following identity, 
\begin{equation}
	\mathbf{C}^{-1} = \beta_{\mathcal{S}} - \beta_{\mathcal{S}}^2 \bm{\Psi}_p \bm{\Sigma}_{\mathbf{p}\mathbf{p}|\mathbf{m}} \bm{\Psi}_p^{\mathrm{T}} \, , \label{eq:CTilde_inverse_woodb}
\end{equation}
we can state
\begin{equation}
	\pdv{\mathbf{C}^{-1}}{\alpha_{p,i}} = - \beta_{\mathcal{S}}^2 \bm{\Psi}_p \pdv{\bm{\Sigma}_{\mathbf{p}\mathbf{p}|\mathbf{m}}}{\alpha_{p,i}} \bm{\Psi}_p^{\mathrm{T}} =  
	\beta_{\mathcal{S}}^2 \bm{\Psi}_p \bm{\Sigma}_{\mathbf{p}\mathbf{p}|\mathbf{m}} \mathbb{I}_{ii} \bm{\Sigma}_{\mathbf{p}\mathbf{p}|\mathbf{m}} \bm{\Psi}_p^{\mathrm{T}} \, ,
\end{equation}
since $\pdv{\bm{\Sigma}_{\mathbf{p}\mathbf{p}|\mathbf{m}}}{\alpha_{p,i}} = - \bm{\Sigma}_{\mathbf{p}\mathbf{p}|\mathbf{m}} \pdv{\bm{\Sigma}_{\mathbf{p}\mathbf{p}|\mathbf{m}}^{-1}}{\alpha_{p,i}}  \bm{\Sigma}_{\mathbf{p}\mathbf{p}|\mathbf{m}}$. 
$\mathbb{I}_{ij}$ denotes the single-entry matrix with unit-entry at position $(i,j)$.
Thereby
\begin{equation}
	\pdv{\ln \det \left( - \mathbf{H}_{\mathbf{q}\mathbf{q}} \right)}{\alpha_{p,i}} = \tr  \left( \beta_{\mathcal{S}}^2 \left( - \mathbf{H}_{\mathbf{q}\mathbf{q}} \right)^{-1} \bm{\Upsilon}^{\mathrm{H}} \bm{\Psi}_p \bm{\Sigma}_{\mathbf{p}\mathbf{p}|\mathbf{m}} \mathbb{I}_{ii} \bm{\Sigma}_{\mathbf{p}\mathbf{p}|\mathbf{m}} \bm{\Psi}_p^{\mathrm{T}} \bm{\Upsilon} \right) \, .
\end{equation}
Through the cyclic permutation property of the trace operator, we find
\begin{align}
	\pdv{\ln \det \left( - \mathbf{H}_{\mathbf{q}\mathbf{q}} \right)}{\alpha_{p,i}} & = \tr \left( \mathbb{I}_{ii} \left( \beta_{\mathcal{S}} \bm{\Sigma}_{\mathbf{p}\mathbf{p}|\mathbf{m}} \bm{\Psi}_p^{\mathrm{T}} \bm{\Upsilon} \right) \left( - \mathbf{H}_{\mathbf{q}\mathbf{q}} \right)^{-1} \left( \beta_{\mathcal{S}} \bm{\Upsilon}^{\mathrm{H}} \bm{\Psi}_p \bm{\Sigma}_{\mathbf{p}\mathbf{p}|\mathbf{m}} \right) \right) \nonumber \\
	& = \left[ \bm{\Delta}^\mathrm{H} \left( - \mathbf{H}_{\mathbf{q}\mathbf{q}} \right)^{-1} \bm{\Delta} \right]_{ii} \, . \label{eq:deriv_log_hessian_alpha_p}
\end{align}
where $\bm{\Delta} \in \mathbb{C}^{n_q \times n_p}$ is defined as
\begin{equation}
	\bm{\Delta} = \beta_{\mathcal{S}} \bm{\Upsilon}^{\mathrm{H}} \bm{\Psi}_p \bm{\Sigma}_{\mathbf{p}\mathbf{p}|\mathbf{m}} = \bm{\Upsilon}^{\mathrm{H}} \mathbf{C}^{-1} \bm{\Psi}_p \bm{\Lambda}_{\mathbf{p}\mathbf{p}}^{-1} \, .
	\label{eq:app_grad_delta}
\end{equation}
Combining the results in Eq.~\eqref{eq:deriv_log_hessian_alpha_q} with Eq.~\eqref{eq:deriv_log_evi_alpha_q}, we obtain
\begin{align}
	\pdv{\obj_{\bm{\alpha}_p} (\bm{\alpha}_p)}{\alpha_{p,i}} = - \left[ \bm{\Sigma}_{\mathbf{p}\mathbf{p}| \mathbf{m}, \mathbf{q}^\ast} \right]_{ii} + \alpha_{p,i}^{-1} - \abs{ \left[\mu_{\mathbf{p}|\mathbf{m}, \mathbf{q}^\ast} \right]_i }^2 - \left[ \bm{\Delta}^\mathrm{H} \left( - \mathbf{H}_{\mathbf{q}\mathbf{q}} \right)^{-1} \bm{\Delta} \right]_{ii} \, .
\end{align}
Setting the derivative to zero, i.e., $\pdv{\obj_{\bm{\alpha}_p} (\bm{\alpha}_p)}{\alpha_{p,i}} = 0$, we can find the optimal value in closed form:
\begin{equation}
	\alpha_{p,i} = \frac{1}{\left[ \bm{\Sigma}_{\mathbf{p}\mathbf{p}|\mathbf{m}, \mathbf{q}} \right]_{ii} + \abs{\left[ \bm{\mu}_{\mathbf{p}|\mathbf{m}, \mathbf{q}^\ast} \right]_i}^2 + \left[ \bm{\Delta}^\mathrm{H} \left( - \mathbf{H}_{\mathbf{q}\mathbf{q}} \right)^{-1} \bm{\Delta} \right]_{ii}} \, .
\end{equation}
Alternatively, we can define a fixed-point update as
\begin{gather}
	\underbrace{1 - \alpha_{p,i} \left[ \bm{\Sigma}_{\mathbf{p}\mathbf{p}| \mathbf{m}, \mathbf{q}^\ast} \right]_{ii}}_{=:\gamma_{p,i}} - \alpha_{p,i} \abs{ \left[\mu_{\mathbf{p}|\mathbf{m}, \mathbf{q}^\ast} \right]_i }^2 - \alpha_{p,i} \left[ \bm{\Delta}^\mathrm{H} \left( - \mathbf{H}_{\mathbf{q}\mathbf{q}} \right)^{-1} \bm{\Delta} \right]_{ii} = 0 \\
	\to \alpha_{p,i} = \frac{\gamma_{p,i}}{\abs{\left[ \bm{\mu}_{\mathbf{p}|\mathbf{m}, \mathbf{q}^\ast} \right]_i}^2 + \left[ \bm{\Delta}^\mathrm{H} \left( - \mathbf{H}_{\mathbf{q}\mathbf{q}} \right)^{-1} \bm{\Delta} \right]_{ii}} \, .
\end{gather}
\section{Partial derivatives with respect to the hyperparameters $\bm{\alpha}_q$}
\label{app:grad_alpha_q}
In this section we derive the partial derivatives of the objective function as defined in the optimization problem in Eq.~\eqref{eq:hyper_MAP2} with respect to the denominator coefficient precisions.
To this end, we define
\begin{equation}
	\obj_{\bm{\alpha}_q} (\bm{\alpha}_q) = 
	\ln \det \bm{\Lambda}_{\mathbf{q}\mathbf{q}} -
	- \mathbf{q}^{\ast \mathrm{H}} \bm{\Lambda}_{\mathbf{q} \mathbf{q}} \mathbf{q}^{\ast}  
	- \ln \det \left( - \mathbf{H}_{\mathbf{q}\mathbf{q}} \right)
	\, .
	\label{eq:obj_q_alpha_q_app}
\end{equation}
We start by computing 
\begin{equation}
	\pdv{\alpha_{q,i}} \left[\ln \det \bm{\Lambda}_{\mathbf{q}\mathbf{q}} - \mathbf{q}^{\ast \mathrm{H}} \bm{\Lambda}_{\mathbf{q} \mathbf{q}} \mathbf{q}^{\ast} \right] =   
	\pdv{\alpha_{q,i}} \left[ \sum_{j=0}^{n_q - 1} \ln \alpha_{q,j} - \alpha_{q,j} \abs{q_j^\ast}^2 \right] = \alpha_{q,i}^{-1} - \abs{q_i^\ast}^2 \, .
	\label{eq:deriv_log_hessian_alpha_q}
\end{equation}
Furthermore,
\begin{align}
	\pdv{\ln \det \left( - \mathbf{H}_{\mathbf{q}\mathbf{q}} \right)}{\alpha_{q,i}} = \tr  \left( \left( - \mathbf{H}_{\mathbf{q}\mathbf{q}} \right)^{-1} \pdv{\left( - \mathbf{H}_{\mathbf{q}\mathbf{q}} \right)}{\alpha_{q,i}} \right) = \tr \left( \left( - \mathbf{H}_{\mathbf{q}\mathbf{q}} \right)^{-1} \mathbb{I}_{ii} \right) = \left[ \left( - \mathbf{H}_{\mathbf{q}\mathbf{q}} \right)^{-1} \right]_{ii} \, ,
	\label{eq:deriv_log_evi_alpha_q}
\end{align}
since
\begin{equation}
	\pdv{\left(- \mathbf{H}_{\mathbf{q}\mathbf{q}} \right)}{\alpha_{q,i}} = \pdv{\bm{\Lambda}_{\mathbf{q} \mathbf{q}}}{\alpha_{q,i}} = \mathbb{I}_{ii} \, .
\end{equation}
Combining the results in Eqs.~\eqref{eq:deriv_log_hessian_alpha_q} and \eqref{eq:deriv_log_evi_alpha_q}, we obtain
\begin{align}
	\pdv{\obj_{\bm{\alpha}_q} (\bm{\alpha}_q)}{\alpha_{q,i}} = \alpha_{q,i}^{-1} - \abs{q_i^\ast}^2 - \left[ \left( - \mathbf{H}_{\mathbf{q}\mathbf{q}} \right)^{-1} \right]_{ii} \, .
\end{align}
Setting the derivative to zero, i.e., $\pdv{\obj_{\bm{\alpha}_q} (\bm{\alpha}_q)}{\alpha_{q,i}} = 0$, we can find the optimal value in closed form:
\begin{equation}
	\alpha_{q,i} = \frac{1}{\abs{q_i^\ast}^2 + \left[ \left( - \mathbf{H}_{\mathbf{q}\mathbf{q}} \right)^{-1} \right]_{ii}} \, .
\end{equation}
Alternatively, we can define a fix-point update as
\begin{gather}
	\underbrace{1 - \alpha_{q,i} \left[ \left( - \mathbf{H}_{\mathbf{q}\mathbf{q}} \right)^{-1} \right]_{ii}}_{=:\gamma_{q,i}} - \alpha_{q,i} \abs{q_i^\ast}^2 = 0 \\
	\to \alpha_{q,i} = \frac{\gamma_{q,i}}{\abs{q_i^\ast}^2} \, .
\end{gather}
\section{Partial derivatives with respect to the hyperparameter $\beta_{\mathcal{S}}$}
\label{app:grad_beta}
In this section we derive the partial derivatives of the objective function as defined in the optimization problem in Eq.~\eqref{eq:hyper_MAP2} with respect to the error model precision $\beta_{\mathcal{S}}$.
To this end, we define
\begin{equation}
	\obj_{\beta_{\mathcal{S}}} (\beta_{\mathcal{S}}) = 
	\ntrain \ln \beta_{\mathcal{S}} +
	\ln \det \bm{\Sigma}_{\mathbf{p}\mathbf{p}| \mathbf{m}} 
	%
	- \beta_{\mathcal{S}} \mathbf{m}^{\mathrm{H}} \conj{\mathbf{Q}^\ast} \left( \mathbf{Q}^\ast \mathbf{m} - \bm{\Psi}_p \bm{\mu}_{\mathbf{p}|\mathbf{m}} \right)  
	- \ln \det \left( - \mathbf{H}_{\mathbf{q}\mathbf{q}} \right)
	\, .
	\label{eq:obj_q_beta_app}
\end{equation}
Analogously to the argument made in~\ref{app:grad_alpha_p}, we observe that we can transfer results from Appendix F in \cite{Schneider.2023} and find 
\begin{equation}
	\pdv{\beta_{\mathcal{S}}} \left[ \ntrain \ln \beta_{\mathcal{S}} + \ln \det \bm{\Sigma}_{\mathbf{p}\mathbf{p}| \mathbf{m}}
	- \beta_{\mathcal{S}} \mathbf{m}^{\mathrm{H}} \conj{\mathbf{Q}^\ast} \left( \mathbf{Q}^\ast \mathbf{m} - \bm{\Psi}_p \bm{\mu}_{\mathbf{p}|\mathbf{m}} \right) \right] 
	= 
	\frac{\ntrain}{\beta_{\mathcal{S}}} 
	- \tr \left( \bm{\Sigma}_{\mathbf{p}\mathbf{p}| \mathbf{m}} \bm{\Psi}_p^{\mathrm{T}} \bm{\Psi}_p \right) 
	- \norm{\mathbf{Q} \mathbf{m} - \bm{\Psi}_p \bm{\mu}_{\mathbf{p}|\mathbf{m}, \mathbf{q}^\ast}}^2 \, . \label{eq:deriv_log_evi_beta}
\end{equation}
The remaining element is the computation of the derivative of the log-determinant of the Hessian matrix in Eq.~\eqref{eq:obj_q_beta_app}.
It holds
\begin{equation}
	\pdv{\ln \det \left( - \mathbf{H}_{\mathbf{q}\mathbf{q}} \right)}{\beta_\mathcal{S}} = \tr  \left(  \left( - \mathbf{H}_{\mathbf{q}\mathbf{q}} \right)^{-1} \pdv{ \left( - \mathbf{H}_{\mathbf{q}\mathbf{q}} \right)}{\beta_\mathcal{S}} \right) = \tr  \left(  \left( - \mathbf{H}_{\mathbf{q}\mathbf{q}} \right)^{-1} \bm{\Upsilon}^{\mathrm{H}} \pdv{\mathbf{C}^{-1}}{\beta_\mathcal{S}} \bm{\Upsilon} \right) \, ,
\end{equation}
Under consideration of Eq.~\eqref{eq:evi_C_matrix}, we find
\begin{equation}
	\pdv{\mathbf{C}^{-1}}{\beta_\mathcal{S}} 
	= - \mathbf{C}^{-1} \pdv{\mathbf{C}}{\beta_\mathcal{S}} \mathbf{C}^{-1} 
	= - \beta_\mathcal{S}^{-2} \mathbf{C}^{-1} \mathbf{C}^{-1} \, .
	\label{eq:deriv_C_tilde_beta1}
\end{equation}
Thus,
\begin{equation}
	\pdv{\ln \det \left( - \mathbf{H}_{\mathbf{q}\mathbf{q}} \right)}{\beta_\mathcal{S}} = \tr  \left( \left( - \mathbf{H}_{\mathbf{q}\mathbf{q}} \right)^{-1} \bm{\Upsilon}^{\mathrm{H}} \mathbf{C}^{-1} \pdv{\bm{\Sigma}_{\Tilde{\bm{\varepsilon}}\Tilde{\bm{\varepsilon}}}}{\bm{\theta}} \mathbf{C}^{-1} \bm{\Upsilon} \right) 
	= \tr \left( \left( - \mathbf{H}_{\mathbf{q}\mathbf{q}} \right)^{-1} \mathbf{O}^{\mathrm{H}} \mathbf{O} \right) \, , \label{eq:deriv_log_hessian_beta}
\end{equation}
where 
\begin{equation}
	\mathbf{O} = \beta_{\mathcal{S}}^{-1} \mathbf{C}^{-1} \bm{\Upsilon} \, .
	\label{eq:omikron1}
\end{equation}
For numerical purposes it can be beneficial to rewrite the inverse $\mathbf{C}^{-1}$ using the Woodbury inversion theorem.
Then, Eq.~\eqref{eq:omikron1} reads
\begin{equation}
	\mathbf{O} 
	= \beta_{\mathcal{S}}^{-1} \left( \beta_{\mathcal{S}} \mathbf{I}_{\ntrain} - \beta_{\mathcal{S}}^2 \bm{\Psi}_p \bm{\Sigma}_{\mathbf{p}\mathbf{p}|\mathbf{m}} \bm{\Psi}_p^{\mathrm{T}} \right) \bm{\Upsilon} 
	= \left( \mathbf{I}_{\ntrain} - \beta_{\mathcal{S}} \bm{\Psi}_p \bm{\Sigma}_{\mathbf{p}\mathbf{p}|\mathbf{m}} \bm{\Psi}_p^{\mathrm{T}} \right) \bm{\Upsilon} \, .
	\label{eq:omikron2}
\end{equation}
Combining the results in Eqs.~\eqref{eq:deriv_log_hessian_beta}, \eqref{eq:deriv_log_evi_beta} and \eqref{eq:omikron2}, we obtain
\begin{align}
	\pdv{\obj_{\beta_{\mathcal{S}}}}{\beta_{\mathcal{S}}} = 
	\frac{\ntrain}{\beta_{\mathcal{S}}} 
	- \tr \left( \bm{\Sigma}_{\mathbf{p}\mathbf{p}|\mathbf{m}} \bm{\Psi}_p^{\mathrm{T}} \bm{\Psi}_p \right) 
	- \norm{\mathbf{Q} \mathbf{m} - \bm{\Psi}_p \bm{\mu}_{\mathbf{p}|\mathbf{m}}}^2 
	- \tr \left( \left( - \mathbf{H}_{\mathbf{q}\mathbf{q}} \right)^{-1} \mathbf{O}^{\mathrm{H}} \mathbf{O} \right)   \, .
\end{align}
Setting the derivative to zero, we can find the optimal value analytically,
\begin{equation}
	\beta_{\mathcal{S}} = \frac{\ntrain}{\norm{\mathbf{Q} \mathbf{m} - \bm{\Psi}_p \bm{\mu}_{\mathbf{p}|\mathbf{m}, \mathbf{q}^\ast}}^2 + \tr \left( \bm{\Sigma}_{\mathbf{p}\mathbf{p}|\mathbf{m}} \bm{\Psi}_p^{\mathrm{T}} \bm{\Psi}_p \right) + \tr \left( \left( - \mathbf{H}_{\mathbf{q}\mathbf{q}} \right)^{-1} \mathbf{O}^{\mathrm{H}} \mathbf{O} \right)} \, .
\end{equation}
Alternatively, we can define the following fixed-point update as follows
\begin{equation}
	\beta_{\mathcal{S}} = \frac{n_{\mathcal{M}} - \sum_{i=0}^{n_p-1} \gamma_{p,i}}{\norm{\mathbf{Q} \mathbf{m} - \bm{\Psi}_p \bm{\mu}_{\mathbf{p}|\mathbf{m}, \mathbf{q}^\ast}}^2 + \tr \left( \left( - \mathbf{H}_{\mathbf{q}\mathbf{q}} \right)^{-1} \mathbf{O}^{\mathrm{H}} \mathbf{O} \right)} \, .
\end{equation}

\clearpage
\bibliographystyle{elsarticle-num}
\bibliography{dokument_red}

\begin{thebibliography}{10}
\expandafter\ifx\csname url\endcsname\relax
  \def\url#1{\texttt{#1}}\fi
\expandafter\ifx\csname urlprefix\endcsname\relax\def\urlprefix{URL }\fi
\expandafter\ifx\csname href\endcsname\relax
  \def\href#1#2{#2} \def\path#1{#1}\fi

\bibitem{Astrom.1971}
K.~J. {\AA}str{\"o}m, P.~Eykhoff, System identification---{A} survey,
  Automatica 7~(2) (1971) 123--162.
\newblock \href {http://dx.doi.org/10.1016/0005-1098(71)90059-8}
  {\path{doi:10.1016/0005-1098(71)90059-8}}.

\bibitem{Soederstroem.1989}
T.~Soederstroem, P.~Stoica, System {I}dentification, Prentice Hall
  International {S}eries in {S}ystems and {C}ontrol {E}ngineering, {Prentice
  Hall}, New York and London, 1989.

\bibitem{Mottershead.1993}
J.~E. Mottershead, M.~I. Friswell, Model updating in structural dynamics: A
  survey, Journal of Sound and Vibration 167~(2) (1993) 347--375.
\newblock \href {http://dx.doi.org/10.1006/jsvi.1993.1340}
  {\path{doi:10.1006/jsvi.1993.1340}}.

\bibitem{Pintelon.2012}
R.~Pintelon, J.~Schoukens, System identification: a frequency domain approach,
  John Wiley \& Sons, 2012.

\bibitem{friswell2013finite}
M.~Friswell, J.~E. Mottershead, Finite element model updating in structural
  dynamics, Vol.~38, Springer Science \& Business Media, 2013.

\bibitem{gelman2013bayesian}
A.~Gelman, H.~S. Stern, J.~B. Carlin, D.~B. Dunson, A.~Vehtari, D.~B. Rubin,
  Bayesian data analysis, Chapman and Hall/CRC, 2013.

\bibitem{Beck.2010}
J.~L. Beck, Bayesian system identification based on probability logic,
  Structural Control and Health Monitoring 17~(7) (2010) 825--847.
\newblock \href {http://dx.doi.org/10.1002/stc.424}
  {\path{doi:10.1002/stc.424}}.

\bibitem{mackay2003information}
D.~J. MacKay, D.~J. Mac~Kay, Information theory, inference and learning
  algorithms, Cambridge {U}niversity {P}ress, 2003.

\bibitem{gilks1995markov}
W.~R. Gilks, S.~Richardson, D.~Spiegelhalter, Markov chain Monte Carlo in
  practice, Chapman and Hall/CRC, 1995.

\bibitem{cheung2017new}
S.~H. Cheung, S.~Bansal, A new gibbs sampling based algorithm for bayesian
  model updating with incomplete complex modal data, Mechanical Systems and
  Signal Processing 92 (2017) 156--172.

\bibitem{muto.2008}
M.~Muto, J.~L. Beck, Bayesian updating and model class selection for hysteretic
  structural models using stochastic simulation, Journal of {V}ibration and
  {C}ontrol 14~(1-2) (2008) 7--34.
\newblock \href {http://dx.doi.org/10.1177/1077546307079400}
  {\path{doi:10.1177/1077546307079400}}.

\bibitem{Zhang.2010}
E.~Zhang, P.~Feissel, J.~Antoni, Bayesian model updating with consideration of
  modeling error, European Journal of Computational Mechanics 19~(1-3) (2010)
  255--266.
\newblock \href {http://dx.doi.org/10.3166/EJCM.19.255-266}
  {\path{doi:10.3166/EJCM.19.255-266}}.

\bibitem{Zhang.2011}
E.~L. Zhang, P.~Feissel, J.~Antoni, A comprehensive bayesian approach for model
  updating and quantification of modeling errors, Probabilistic {E}ngineering
  {M}echanics 26~(4) (2011) 550--560.
\newblock \href {http://dx.doi.org/10.1016/j.probengmech.2011.07.001}
  {\path{doi:10.1016/j.probengmech.2011.07.001}}.

\bibitem{Zhang.2013}
E.~Zhang, J.~D. Chazot, J.~Antoni, M.~Hamdi, Bayesian characterization of
  young's modulus of viscoelastic materials in laminated structures, Journal of
  Sound and Vibration 332~(16) (2013) 3654--3666.
\newblock \href {http://dx.doi.org/10.1016/j.jsv.2013.02.032}
  {\path{doi:10.1016/j.jsv.2013.02.032}}.

\bibitem{Straub.2015}
D.~Straub, I.~Papaioannou, Bayesian updating with structural reliability
  methods, Journal of {E}ngineering {M}echanics 141~(3) (2015) 04014134.
\newblock \href {http://dx.doi.org/10.1061/(ASCE)EM.1943-7889.0000839}
  {\path{doi:10.1061/(ASCE)EM.1943-7889.0000839}}.

\bibitem{diazdelao2017bayesian}
F.~DiazDelaO, A.~Garbuno-Inigo, S.~Au, I.~Yoshida, Bayesian updating and model
  class selection with subset simulation, Computer {M}ethods in {A}pplied
  {M}echanics and {E}ngineering 317 (2017) 1102--1121.

\bibitem{Betz.2018}
W.~Betz, I.~Papaioannou, J.~L. Beck, D.~Straub, Bayesian inference with subset
  simulation: Strategies and improvements, Computer Methods in Applied
  Mechanics and Engineering 331 (2018) 72--93.
\newblock \href {http://dx.doi.org/10.1016/j.cma.2017.11.021}
  {\path{doi:10.1016/j.cma.2017.11.021}}.

\bibitem{ghanem1991stochastic}
R.~G. Ghanem, P.~D. Spanos, Stochastic finite element method: Response
  statistics, in: Stochastic Finite Elements: A Spectral Approach, Springer,
  1991, pp. 101--119.

\bibitem{xiu2002wiener}
D.~Xiu, G.~E. Karniadakis, The {W}iener--{A}skey polynomial chaos for
  stochastic differential equations, SIAM Journal on Scientific Computing
  24~(2) (2002) 619--644.

\bibitem{yamazaki1988neumann}
F.~Yamazaki, M.~Shinozuka, G.~Dasgupta, Neumann expansion for stochastic finite
  element analysis, Journal of {E}ngineering {M}echanics 114~(8) (1988)
  1335--1354.

\bibitem{papadrakakis1996structural}
M.~Papadrakakis, V.~Papadopoulos, N.~D. Lagaros, Structural reliability analyis
  of elastic-plastic structures using neural networks and {M}onte {C}arlo
  simulation, Computer Methods in Applied Mechanics and Engineering 136~(1-2)
  (1996) 145--163.

\bibitem{Rasmussen.2008}
C.~E. Rasmussen, C.~K.~I. Williams, Gaussian processes for machine learning,
  3rd Edition, Adaptive computation and machine learning, {MIT Press},
  Cambridge, Mass., 2008.

\bibitem{mckay1979comparison}
M.~D. McKay, R.~J. Beckman, W.~J. Conover, Comparison of three methods for
  selecting values of input variables in the analysis of output from a computer
  code, Technometrics 21~(2) (1979) 239--245.

\bibitem{owen2003quasi}
A.~B. Owen, Quasi-{M}onte {C}arlo sampling, Monte Carlo Ray Tracing: Siggraph 1
  (2003) 69--88.

\bibitem{garnett_bayesoptbook_2023}
R.~Garnett, {Bayesian Optimization}, Cambridge University Press, 2023.

\bibitem{Frazier.2018}
P.~I. Frazier, A tutorial on bayesian optimization.
\newblock \href {http://dx.doi.org/10.48550/arXiv.1807.02811}
  {\path{doi:10.48550/arXiv.1807.02811}}.

\bibitem{Shahriari.2016}
B.~Shahriari, K.~Swersky, Z.~Wang, R.~P. Adams, N.~de~Freitas, Taking the human
  out of the loop: A review of bayesian optimization, Proceedings of the IEEE
  104~(1) (2016) 148--175.
\newblock \href {http://dx.doi.org/10.1109/jproc.2015.2494218}
  {\path{doi:10.1109/jproc.2015.2494218}}.

\bibitem{Greenhill.2020}
S.~Greenhill, S.~Rana, S.~Gupta, P.~Vellanki, S.~Venkatesh, Bayesian
  optimization for adaptive experimental design: A review, IEEE Access 8 (2020)
  13937--13948.
\newblock \href {http://dx.doi.org/10.1109/ACCESS.2020.2966228}
  {\path{doi:10.1109/ACCESS.2020.2966228}}.

\bibitem{Wang.2023}
X.~Wang, Y.~Jin, S.~Schmitt, M.~Olhofer, Recent advances in bayesian
  optimization, ACM Computing Surveys 55~(13s) (2023) 1--36.
\newblock \href {http://dx.doi.org/10.1145/3582078}
  {\path{doi:10.1145/3582078}}.

\bibitem{Jones.1998}
D.~R. Jones, M.~Schonlau, W.~J. Welch,
  \href{https://link.springer.com/article/10.1023/a:1008306431147}{Efficient
  global optimization of expensive black-box functions}, Journal of Global
  Optimization 13~(4) (1998) 455--492.
\newblock \href {http://dx.doi.org/10.1023/A:1008306431147}
  {\path{doi:10.1023/A:1008306431147}}.
\newline\urlprefix\url{https://link.springer.com/article/10.1023/a:1008306431147}

\bibitem{Jin.2016}
S.-S. Jin, H.-J. Jung, Sequential surrogate modeling for efficient finite
  element model updating, Computers {\&} Structures 168 (2016) 30--45.
\newblock \href {http://dx.doi.org/10.1016/j.compstruc.2016.02.005}
  {\path{doi:10.1016/j.compstruc.2016.02.005}}.

\bibitem{Pandita.2021}
P.~Pandita, P.~Tsilifis, N.~M. Awalgaonkar, I.~Bilionis, J.~Panchal,
  Surrogate-based sequential bayesian experimental design using non-stationary
  gaussian processes, Computer Methods in Applied Mechanics and Engineering 385
  (2021) 114007.
\newblock \href {http://dx.doi.org/10.1016/j.cma.2021.114007}
  {\path{doi:10.1016/j.cma.2021.114007}}.

\bibitem{Kapadia.2024}
H.~Kapadia, L.~Feng, P.~Benner, Active-learning-driven surrogate modeling for
  efficient simulation of parametric nonlinear systems, Computer Methods in
  Applied Mechanics and Engineering 419 (2024) 116657.
\newblock \href {http://dx.doi.org/10.1016/j.cma.2023.116657}
  {\path{doi:10.1016/j.cma.2023.116657}}.

\bibitem{Schneider.2022}
F.~Schneider, I.~Papaioannou, D.~Straub, C.~Winter, G.~M{\"u}ller,
  \href{https://www.sciencedirect.com/science/article/pii/S0888327021007573}{Bayesian
  parameter updating in linear structural dynamics with frequency transformed
  data using rational surrogate models}, Mechanical Systems and Signal
  Processing 166 (2022) 108407.
\newblock \href {http://dx.doi.org/10.1016/j.ymssp.2021.108407}
  {\path{doi:10.1016/j.ymssp.2021.108407}}.
\newline\urlprefix\url{https://www.sciencedirect.com/science/article/pii/S0888327021007573}

\bibitem{Mares.2006}
C.~Mares, B.~Dratz, J.~Mottershead, M.~Friswell, Model updating using
  {B}ayesian estimation, in: International Conference on Noise and Vibration
  Engineering, ISMA2006, Katholieke Universiteit Leuven, 2006, pp. 18--20.

\bibitem{Bect.25.07.2023}
J.~Bect, N.~Georg, U.~R{\"o}mer, S.~Sch{\"o}ps,
  \href{http://arxiv.org/pdf/2307.13484v2}{Rational kernel-based interpolation
  for complex-valued frequency response functions}.
\newline\urlprefix\url{http://arxiv.org/pdf/2307.13484v2}

\bibitem{wagner2021bayesian}
P.-R. Wagner, S.~Marelli, B.~Sudret, Bayesian model inversion using stochastic
  spectral embedding, Journal of Computational Physics 436 (2021) 110141.

\bibitem{Schneider.2023}
F.~Schneider, I.~Papaioannou, G.~M{\"u}ller, Sparse bayesian learning for
  complex--valued rational approximations, International Journal for Numerical
  Methods in Engineering 124~(8) (2023) 1721--1747.
\newblock \href {http://dx.doi.org/10.1002/nme.7182}
  {\path{doi:10.1002/nme.7182}}.

\bibitem{Jacquelin.2015}
E.~Jacquelin, S.~Adhikari, J.-J. Sinou, M.~I. Friswell, Polynomial chaos
  expansion and steady-state response of a class of random dynamical systems,
  Journal of Engineering Mechanics 141~(4) (2015) 04014145.
\newblock \href {http://dx.doi.org/10.1061/(ASCE)EM.1943-7889.0000856}
  {\path{doi:10.1061/(ASCE)EM.1943-7889.0000856}}.

\bibitem{Jacquelin.2015b}
E.~Jacquelin, S.~Adhikari, J.-J. Sinou, M.~I. Friswell, Polynomial chaos
  expansion in structural dynamics: Accelerating the convergence of the first
  two statistical moment sequences, Journal of Sound and Vibration 356 (2015)
  144--154.
\newblock \href {http://dx.doi.org/10.1016/j.jsv.2015.06.039}
  {\path{doi:10.1016/j.jsv.2015.06.039}}.

\bibitem{Jacquelin.2016}
E.~Jacquelin, S.~Adhikari, M.~I. Friswell, J.-J. Sinou, Role of roots of
  orthogonal polynomials in the dynamic response of stochastic systems, Journal
  of Engineering Mechanics 142~(8) (2016) 06016004.
\newblock \href {http://dx.doi.org/10.1061/(ASCE)EM.1943-7889.0001102}
  {\path{doi:10.1061/(ASCE)EM.1943-7889.0001102}}.

\bibitem{JacquelinE.2016}
{Jacquelin E.}, {Dessombz O.}, {Sinou J.--J.}, {Adhikari S.}, {Friswell M. I.},
  Polynomial chaos--based extended pad{\'e} expansion in structural dynamics,
  International Journal for Numerical Methods in Engineering 111~(12) (2016)
  1170--1191.
\newblock \href {http://dx.doi.org/10.1002/nme.5497}
  {\path{doi:10.1002/nme.5497}}.

\bibitem{Schneider.2019}
F.~Schneider, I.~Papaioannou, M.~Ehre, D.~Straub,
  \href{http://www.sciencedirect.com/science/article/pii/S0045794920300262}{Polynomial
  chaos based rational approximation in linear structural dynamics with
  parameter uncertainties}, Computers \& Structures 233 (2020) 106223.
\newblock \href
  {http://dx.doi.org/https://doi.org/10.1016/j.compstruc.2020.106223}
  {\path{doi:https://doi.org/10.1016/j.compstruc.2020.106223}}.
\newline\urlprefix\url{http://www.sciencedirect.com/science/article/pii/S0045794920300262}

\bibitem{Tipping.2001}
M.~E. Tipping, Sparse bayesian learning and the relevance vector machine,
  Journal of machine learning research 1~(Jun) (2001) 211--244.

\bibitem{Geradin.1997}
M.~G{\'e}radin, D.~Rixen, Mechanical vibrations: Theory and application to
  structural dynamics, 3rd Edition, Wiley, Hoboken, New Jersey, 1997.

\bibitem{rosenblatt1952remarks}
M.~Rosenblatt, Remarks on a multivariate transformation, The Annals of
  Mathematical Statistics 23~(3) (1952) 470--472.

\bibitem{Blatman.2011}
G.~Blatman, B.~Sudret, Adaptive sparse polynomial chaos expansion based on
  least angle regression, Journal of Computational Physics 230~(6) (2011)
  2345--2367.
\newblock \href {http://dx.doi.org/10.1016/j.jcp.2010.12.021}
  {\path{doi:10.1016/j.jcp.2010.12.021}}.

\bibitem{cox2007ideals}
D.~Cox, J.~Little, D.~O'shea, Ideals, varieties, and algorithms, Vol.~3,
  Springer, 2007.

\bibitem{Chantrasmi.2009}
T.~Chantrasmi, A.~Doostan, G.~Iaccarino, Pad{\'e}--legendre approximants for
  uncertainty analysis with discontinuous response surfaces, Journal of
  Computational Physics 228~(19) (2009) 7159--7180.
\newblock \href {http://dx.doi.org/10.1016/j.jcp.2009.06.024}
  {\path{doi:10.1016/j.jcp.2009.06.024}}.

\bibitem{Schreier.2010}
P.~J. Schreier, L.~L. Scharf,
  \href{https://doi.org/10.1017/CBO9780511815911}{Statistical signal processing
  of complex-valued data: The theory of improper and noncircular signals},
  {Cambridge University Press}, Cambridge, 2010.
\newblock \href {http://dx.doi.org/10.1017/CBO9780511815911}
  {\path{doi:10.1017/CBO9780511815911}}.
\newline\urlprefix\url{https://doi.org/10.1017/CBO9780511815911}

\bibitem{kreutz2009complex}
K.~Kreutz-Delgado, The complex gradient operator and the cr-calculus, arXiv
  preprint arXiv:0906.4835.

\bibitem{Sorber.2013}
L.~Sorber, M.~{van Barel}, L.~D. Lathauwer,
  \href{http://esat.kuleuven.be/stadius/cot/}{Complex optimization toolbox
  v1.03} (2013).
\newline\urlprefix\url{http://esat.kuleuven.be/stadius/cot/}

\bibitem{Sorber.2012}
L.~Sorber, M.~{van Barel}, L.~D. Lathauwer, Unconstrained optimization of real
  functions in complex variables, SIAM Journal on Optimization 22~(3) (2012)
  879--898.
\newblock \href {http://dx.doi.org/10.1137/110832124}
  {\path{doi:10.1137/110832124}}.

\bibitem{Ching.2007}
J.~Ching, Y.-C. Chen, Transitional markov chain monte carlo method for bayesian
  model updating, model class selection, and model averaging, Journal of
  Engineering Mechanics 133~(7) (2007) 816--832.
\newblock \href {http://dx.doi.org/10.1061/(ASCE)0733-9399(2007)133:7(816)}
  {\path{doi:10.1061/(ASCE)0733-9399(2007)133:7(816)}}.

\bibitem{Betz.2016}
W.~Betz, I.~Papaioannou, D.~Straub, Transitional markov chain monte carlo:
  Observations and improvements, Journal of Engineering Mechanics 142~(5).
\newblock \href {http://dx.doi.org/10.1061/(ASCE)EM.1943-7889.0001066}
  {\path{doi:10.1061/(ASCE)EM.1943-7889.0001066}}.

\bibitem{NEURIPS2023_419f72cb}
S.~Ament, S.~Daulton, D.~Eriksson, M.~Balandat, E.~Bakshy,
  \href{https://proceedings.neurips.cc/paper{\_}files/paper/2023/file/419f72cbd568ad62183f8132a3605a2a-Paper-Conference.pdf}{Unexpected
  improvements to expected improvement for bayesian optimization}, in: A.~Oh,
  T.~Neumann, A.~Globerson, K.~Saenko, M.~Hardt, S.~Levine (Eds.), Advances in
  Neural Information Processing Systems, Vol.~36, Curran Associates, Inc.,
  2023, pp. 20577--20612.
\newline\urlprefix\url{https://proceedings.neurips.cc/paper{\_}files/paper/2023/file/419f72cbd568ad62183f8132a3605a2a-Paper-Conference.pdf}

\bibitem{Kennedy.1995}
J.~Kennedy, R.~Eberhart, Particle swarm optimization, in: IEEE International
  Conference on Neural Networks, 1995. Proceedings, {IEEE / Institute of
  Electrical and Electronics Engineers Incorporated}, 1995, pp. 1942--1948.
\newblock \href {http://dx.doi.org/10.1109/ICNN.1995.488968}
  {\path{doi:10.1109/ICNN.1995.488968}}.

\bibitem{Mecking.2017}
S.~Mecking, T.~Kruse, C.~Winter, U.~Schanda, Schlussbericht: Vibroakustik im
  Planungsprozess f{\"u}r Holzbauten: Teilprojekt 3: Parameterentwicklung und
  SEA-Modellierung: Research Report, 2017.

\bibitem{Manfredi.2021}
P.~Manfredi, S.~Grivet-Talocia, Fast stochastic surrogate modeling via rational
  polynomial chaos expansions and principal component analysis, IEEE Access 9
  (2021) 102732--102745.
\newblock \href {http://dx.doi.org/10.1109/ACCESS.2021.3097543}
  {\path{doi:10.1109/ACCESS.2021.3097543}}.

\bibitem{tipping2003fast}
M.~E. Tipping, A.~C. Faul, Fast marginal likelihood maximisation for sparse
  bayesian models, in: International workshop on artificial intelligence and
  statistics, PMLR, 2003, pp. 276--283.

\bibitem{Babacan.2010}
S.~D. Babacan, R.~Molina, A.~K. Katsaggelos, Bayesian compressive sensing using
  laplace priors, IEEE transactions on image processing : a publication of the
  IEEE Signal Processing Society 19~(1) (2010) 53--63.
\newblock \href {http://dx.doi.org/10.1109/TIP.2009.2032894}
  {\path{doi:10.1109/TIP.2009.2032894}}.

\end{thebibliography}

\end{document}